%% file: ms.tex
\pdfoutput=1 
\documentclass[letterpaper]{article} 
\usepackage{times}  
\usepackage{helvet} 
\usepackage{courier}  
\usepackage[hyphens]{url}  
\usepackage{graphicx} 
\usepackage{natbib}  
\usepackage{caption} 
\usepackage[switch]{lineno}  %

\usepackage{graphicx}
\usepackage{subfigure}
\usepackage{amsmath,amsgen,amstext,amssymb}
\usepackage{amsthm}
\usepackage{etoolbox}
\usepackage{algorithm}
\usepackage{algorithmic}
\usepackage{enumitem}
\usepackage{hyperref}

\newcommand{\citeN}[1]{{\citet{#1}}}

\newcommand{\cD}{{\mathcal{D}}}
\newcommand{\cL}{{\mathcal{L}}}
\newcommand{\cM}{{\mathcal{M}}}
\newcommand{\cN}{{\mathcal{N}}}
\newcommand{\cP}{{\mathcal{P}}}

\newcommand{\cT}{{\mathcal{T}}}
\newcommand{\cG}{{\mathcal{G}}}

\newcommand{\be}{{\bf e}}
\newcommand{\blm}{{\bf \mu}}
\newcommand{\bzr}{{\mathbf{0}}}
\newtheorem{theorem}{Theorem}
\newtheorem{lemma}[theorem]{Lemma}
\newtheorem{remark}{Remark}

\newtheorem{assumption}{Assumption}
\newtheorem{corollary}{Corollary}
\newtheorem{proposition}[theorem]{Proposition}

\newcommand{\ProofEnd}{\mbox{$\Box$} \vspace{ 8pt}}
\newcommand{\ProofOf}[1] {{\noindent \bf Proof of~#1: }}

\newcommand{\refeq}[1] {(\ref{#1})}
\newcommand{\refcor}[1]   {Corollary~\ref{#1}}
\newcommand{\refsec}[1]   {{Section~\ref{#1}}}
\newcommand{\refthm}[1]   {{Theorem~\ref{#1}}}

\newcommand{\refprop}[1]  {{Proposition~\ref{#1}}}
\newcommand{\reflem}[1]  {{Lemma~\ref{#1}}}
\newcommand{\refrem}[1]   {{Remark~\ref{#1}}}
\newcommand{\refasm}[1]   {{Assumption~\ref{#1}}}
\newcommand{\refalg}[1]  {{Algorithm~\ref{#1}}}
\newcommand{\reffig}[1]  {{Figure~\ref{#1}}}

\newcommand{\E}{\mathbb{E}}
\newcommand{\Prb}{\mathbb{P}}
\newcommand{\tndi}{\rightarrow \infty}
\newcommand{\tndni}{\rightarrow -\infty}
\newcommand{\tndo}{\rightarrow 0}

\def\D{\mathrm{d}}
\newcommand{\inP}{\mbox{$\,\stackrel{\scriptsize\mbox{p}}{\rightarrow}\,$}}

\newcommand{\DefAs}{:=}
\newcommand{\real}{\mathbb{R}}

\newcommand{\rob}{{\text{rob}}}

\newcommand{\setX}{\mathbb{X}}
\newcommand{\Reals}{\mathbb{R}}

\title{Efficient Stochastic Gradient Descent for Learning \\with Distributionally Robust Optimization}
\author{
    Soumyadip Ghosh\textsuperscript{\rm 1},~Mark S.~Squillante\textsuperscript{\rm 1}~and~
    Ebisa D.~Wollega\textsuperscript{\rm 2} \\
\\
    \textsuperscript{\rm 1}Mathematical Sciences, IBM Research AI, 
    Thomas J.~Watson Research Center, \\
    Yorktown Heights, NY 20198, USA\\
    \\
    \textsuperscript{\rm 2}Department of Engineering, 
    Colorado State University-Pueblo, \\
    Pueblo, CO 81001, USA\\
}
\begin{document}

\maketitle

\begin{abstract}
Distributionally robust optimization (DRO) problems are increasingly seen as a viable method to train machine learning models for improved model generalization.
These min-max formulations, however, are more difficult to solve.
We therefore provide a new stochastic gradient descent algorithm to efficiently solve this DRO formulation.
Our approach applies gradient descent to the outer minimization formulation and estimates the gradient of the inner maximization based on a sample average approximation.
The latter uses a subset of the data in each iteration, progressively increasing the subset size to ensure convergence.
Theoretical results include establishing the optimal manner for growing the support size to balance a fundamental tradeoff between stochastic error and computational effort.
Empirical results demonstrate the significant benefits of our approach over previous work, and also illustrate how learning with DRO can improve generalization.
\end{abstract}

\input{intro}

\input{algo}

\input{expt}

\bibliography{robust_sgd}

\appendix

\clearpage
\newpage
\newpage
\newpage
\input{appendix}

\end{document}

%% file: intro.tex
\section{Introduction}
Consider a general formulation of the distributionally robust optimization (DRO) problem.
Let $\setX$ denote a sample space, $P$ a probability distribution on $\setX$, and $\Theta \subseteq \Reals^d$ a parameter space.
Define $L_P(\theta) \DefAs \E_P[l(\theta,\xi)]$ to be the expectation with respect to (w.r.t.) $P$ of a loss function
$l : \Theta \times \setX \rightarrow \Reals$ representing the estimation error for a learning model with parameters
$\theta \in \Theta$ over data $\xi \in \setX$.
Define the worst-case expected loss
$R(\theta) \DefAs \E_{P^*(\theta)}[ l(\theta,\xi) ] = \sup_{P\in\cP} \{ L_P(\theta) \}$,
which maximizes the loss $L_P$ over a well-defined set of measures $\cP$.
This set typically takes the form
$\cP =\{ P \, | \, D(P, P_{b}) \le \rho , \int dP(\xi) = 1,  P(\xi) \ge 0 \}$, 
where $D(\cdot,\cdot)$ is a metric on the space of probability distributions on $\setX$ and
where the constraints limit the feasible candidates to be within a distance $\rho$ of a base distribution, denoted by $P_{b}$. 
We then seek to find parameters $\theta \in \Theta$ that, for a given $\setX$ and $\cP$, solves the
DRO problem formulated as
\begin{equation}\label{absfmln}
 R(\theta^*_{rob}) \, = \, \min_{\theta\in\Theta} \, \Big\{ R(\theta) \Big\} \, = \,
    \min_{\theta\in\Theta} \, \Big\{ \sup_{P\in\cP} \{ L_P(\theta) \} \Big\}.
\end{equation}

\textbf{DRO and Model Generalization:}
In machine learning, optimal values for model parameters $\theta$ are calculated from a finite training dataset (having size $N$) with a view towards using this model for inference over other test datasets,
all of which are typically assumed to be identically distributed~\cite{vapnik98,wahba90}.
The equal-weight {\em empirical} distribution $U_N = \{1/N\}$ over the finite training dataset is the non-parametric maximum likelihood estimator~\cite{owen01} of the (unknown) distribution underlying the datasets.
Let $\theta^*_{erm}$ denote the minimizer of the empirical loss $L_{U_N}(\cdot)$ over $\Theta$.
In real-world settings, any two finite datasets sampled from the same underlying distribution can violate the identical distribution assumption,
leading to poor {\em generalization} when using $\theta^*_{erm}$ over other datasets~\cite{perlich03}.
Popular model selection techniques, such as cross-validation \cite{stone74},
seek to improve the estimation error between finite training and testing datasets,
but they are often computationally prohibitive and
make it hard to provide rigorous guarantees. 

The DRO formulation~\eqref{absfmln} with $U_N$ over the finite training dataset as the base distribution $P_b$ has been proposed as an alternative approach by \citeN{bwm16}, \citeN{nd17}, \citeN{lm17}, \citeN{gk17},
with roots in non-parametric statistics~\cite{owen01} and optimization \cite{shap06,gk17}.
The approach explicitly treats the ambiguity in the identity of the true (unknown) distribution, denoted by $P_0$.
\citeN{bwm16} show, for Wasserstein distance metrics and an appropriately chosen value of $\rho$ in the set of measures $\cP$,
that $P_0\in\cP$ with high probability. 
The $U_N\neq P_0$ in general and it is highly likely that, at $\theta^*_{rob}$, the 
worst-case distribution $P^*(\theta^*_{rob})\neq P_0$. 
\citeN{nd17} show however, for specific instances of $L_p$ and appropriate setting of the parameter $\rho$
in the set of measures $\cP$,
that this performance gap at
$\theta^*_{rob}$ is $O(1/N)$ whereas this performance gap at the solution $\theta^*_{erm}$ is a larger $O(1/\sqrt{N})$.
The approach therefore holds great promise and a general theory is actively being pursued.

\textbf{Goal:} Our primary focus is on efficiently obtaining solutions of~\eqref{absfmln} so that it is a viable
alternative for model generalization.
The key obstacle is the min-max form, and specifically the inner maximization problem.
In some cases, its solution may be explicitly available;
e.g., instances of Wasserstein distance constrained $\cP$ studied by~\citeN{bwm16} and \citeN{snd18} admit an explicit characterization of the objective
function value $\E_{P^*(\theta)} [l(\theta,\xi)]$;
see also~\cite{gao2018,chen-paschalidis2018,esfahani2018}.
However,
such reductions do
not hold in general for all interesting Wasserstein distance metrics, and require solving a convex non-linear program.~\citeN{nd17} show that the inner maximization with $\chi^2$-divergence constraints can be efficiently solved (see~\refsec{ssec:robopt}). 
We therefore
focus
on
the general $\phi$-divergence 
distance metric:
$D_{\phi}(P,P_b) = \E_{P_b} [\phi( \frac{dP}{dP_b} )]$,
where $\phi(s)$ is a non-negative convex function taking a value of $0$ only at $s=1$.  
The modified $\chi^2$ and Kuhlback-Leibler (KL) divergences are given by $\phi(s) = (s-1)^2$ and $\phi(s) = s\log s - s +1$, respectively.
Define the vector $P := (p_n)$ of dimension $N$.
We
set the base $P_b$ to the uniform empirical distribution, $U_N$,
and thus
the loss function and constraint set $\cP$ are given by $L_{P}(\theta) = \sum_{n=1}^N p_n l(\theta, \xi_n)$ and
$\cP = \{ P \, | \, \sum_{n=1}^N p_n = 1, p_n\ge 0,\forall n, D_{\phi}(P,U_{N}) = \frac{1}{N} \sum_{n=1}^N \phi(Np_n) \le \rho \}.$ 

\textbf{Related Work:}
Assuming
the loss functions
$l(\cdot,\xi_n)$ are convex, this DRO problem is convex in $\theta$.
\citeN{bhwmg13} describe the typical Lagrangian dual algorithm used
for the convex-concave case,
given by \eqref{dualfmln} in the supplement,
where the convex conjugate of $\phi$,
$\phi^*(s) = \max_{u\ge 0} \{su - \phi(u)\}$, is known in closed form
for various $\phi$ such as modified $\chi^2$- and $KL$-divergence.
(Note that this strong duality does not hold when $l$ is non-convex.)
Since the reformulation 
is a standard stochastic minimization problem, \citeN{bhwmg13} apply classical
stochastic gradient descent (SGD) methods to solve it. 
However,~\citeN{nd16} observe that the characteristics of certain dual variables can cause instability in SGD (see discussion following ~(\ref{dualfmln}) in the supplement),
which our experiments show is likely.

\citeN{nd16} therefore propose an alternative approach (for convex losses $l(\theta,\xi)$) that interleaves an SGD step in each of the $\theta$ and $P$ variables,
where the 
steps apply stochastic mirror-descent to each variable.
In effect, the algorithm does a randomized coordinate descent in $P$ along coordinates sampled from the current (non-uniform) probability mass function (pmf) $P$.
This sampling step is computationally demanding, where the fastest methods to sample an $N$-sized non-uniform pmf require $O(N)$
computational effort, which is only slightly smaller than the effort needed to solve the inner maximization problem \emph{exactly} (see below).
Coupled with the coordinate-descent 
iterations 
in a composite dimension $d+N$,
this
results in slow convergence; refer to~\reffig{fig:rho0.1:HIV+nd16} and its discussion in the supplement.

To rectify this, \citeN{nd17} 
determine the
optimal $P^*(\theta)$ that defines $R(\theta)$ directly, namely solving
\refeq{absfmln} as a large deterministic 
problem that is well-defined for finite $N$.
This full-gradient approach is feasible for specific choices of $\phi$-divergences.
For the modified $\chi^2$ case,
they show that the inner maximization can be reduced to two one-dimensional
root-finding problems, which can be solved via bisection search, requiring an $O(N\log N)$
effort (see~\refprop{prop:im_cmp_bnd}) at each iteration, which
can be expensive.

\textbf{Our Contributions: }
We propose and investigate a new SGD algorithm to efficiently solve large-scale DRO problems.
Our \refalg{alg:overall} significantly reduces the expense of computing the exact solution of the inner maximization over the full training dataset by
\emph{subsampling} the support of the variable $P$ from the (finite) training dataset in the iterates of the algorithm and estimating the robust loss via a sample average approximation. 
While subsampling (mini-batching) typically works well with SGD because the gradient estimates are unbiased,
\refthm{thm:bias} shows that, in the DRO context, subsampling induces a bias in the estimation of
the robust loss gradient,  and this can only be reduced if the subsample size grows in~\refalg{alg:overall}. The impact of this bias is illustrated in~\reffig{fig:rcv1} (center). In~\refthm{thm:main-gen}, we establish that convergence is assured, even for non-convex losses $l$,
as long as the subsamples are grown at a certain minimal rate.
Our analysis is extended further for the specific case of $\phi$-divergence constraints
with strongly convex losses $l$ in
\refthm{thm:main}.
This provides a key contribution on understanding how to optimally set the parameters of our algorithm so that,
in a strong statistical sense, the required computational effort is balanced with the required level of accuracy, thus providing the fastest rate of convergence.
%
Our approach is also applied to~\eqref{absfmln} formulated with general 
convex Wasserstein distance constraints in related work that we will not discuss further here.
Additional theoretical results, additional technical details, and all proofs 
can be found in the supplement.

Our empirical results in \refsec{sec:expt} consider 
convex DRO formulations of binary classification problems,
comparing the performance of our algorithm against those proposed by~\citeN{nd16} and~\citeN{nd17}.
Additionally, we study how key parameters of~\refalg{alg:overall} affect performance (\refthm{thm:main}). 
We further show that the formulation~\eqref{absfmln} can
attain the same generalization performance over a model trained with a regularized ERM formulation tuned via $k$-fold cross validation,
where our solution is obtained at a very small fraction of the computational expense.
Additional empirical results and related technical details
can be found in the supplement.


\vskip -1em

%% file: algo.tex
\section{Algorithm and Analysis} \label{sec:algo}
%
\refalg{alg:overall} presents our dynamically sampled subgradient descent algorithm.
It follows SGD-like iterations for the outer minimization problem in~\eqref{absfmln}:
\begin{equation}\label{ssgd}
\theta_{t+1} \; = \;
\theta_t - \gamma_t \nabla_{\theta} \hat{R}_{t}(\theta_t) \; = \;
\theta_t - \gamma_t G_t ,
\end{equation}
where $\gamma_t$ is
the step size (or gain sequence or learning rate),
$\hat{R}_{t}(\cdot)$ is a sample-average approximation of the robust loss $R(\cdot)$ from the inner maximization formulation over $D_{\phi}$-constrained $\cP$,
and $G_t := \nabla_{\theta} \hat{R}_{t}(\theta_t)$.
This view of~\eqref{absfmln} lets us depart from the convex-concave formulations of~\citeN{bhwmg13} and consider non-convex losses $l$,
as long as the subgradient $\nabla_{\theta} \hat{R}_{t}(\cdot)$ approximates the gradient $\nabla_{\theta} {R}(\cdot)$ sufficiently well.
First we establish that the gradient $\nabla_{\theta} {R}(\cdot)$ exists.
To this end, define the set $\Theta_{\varnothing} \DefAs  \{\theta : l(\theta, \xi_{n_1}) = l(\theta, \xi_{n_2}),\,\,\forall\,n_1,n_2 \}$, and for a small $\varsigma> 0$ let 
the set $\Theta_{\varnothing,\varsigma} \DefAs \cup_{\theta_o \in \Theta_{\varnothing} } \{\theta : |\theta - \theta_{o}| < \varsigma \}$ define the $\varsigma$-neighborhood of $\Theta_{\varnothing}$.
\refprop{prop:rgrad} assumes that the learning model precludes this neighborhood set in order to
avoid a degenerate inner maximization objective function that does not depend on the decision variables, in which case the entire feasible set is optimal.
The existence and form of $\nabla_{\theta} R(\theta)$ is then derived in part by exploiting Danskin's Theorem~\citep[Theorem~7.21]{shap09}.

\begin{proposition}\label{prop:rgrad}
	Let the feasible region $\Theta$ be compact and assume $\Theta\subseteq \Theta_{\varnothing,\varsigma}^c$, for a small $\varsigma>0$.
Further suppose $\phi$ in the $D_{\phi}$-constraint has strictly convex level sets, and let $\rho < \bar {\rho}(N,\phi)$ with
$\bar{\rho}(N,\phi)$ defined in~\eqref{rhomax}.
	Then, (i) the optimal solution $P^*$ of $R(\theta)= \sup_{P\in\cP} \{L_P(\theta)\}$ is unique, and (ii) the gradient is given by
$\nabla_{\theta} {R}(\theta) = \sum_{n\in\cN}{p}^*_{n}(\theta) \,\nabla_{\theta} l(\theta,\xi_n)$.
\end{proposition}
%

We depart from~\citeN{nd16, nd17} by constructing the estimate $\hat{R}_t$ from inner maximization problems restricted only to a relatively small subset $\cM_t$ of the full dataset with size $|\cM_t| = M_t$.
In particular, the subset ${\cM_t}$ is constructed by {\em uniformly} sampling {\em without replacement} $M_t$ values from the complete training
dataset (of size $N$).
Sampling without replacement differs from the standard with-replacement approach in the stochastic optimization literature, though it is preferred
by practitioners in machine/deep learning.
\refrem{rmk:smpl} below describes why this strategy is needed here.
%
Defining ${P} = ({p}_m)$
of dimension $M_t$ and the objective coefficients $z_m = l(\theta,\xi_m)$, we have 

\begin{align} \label{restrob}
&\hat{R}_{t}(\theta) \;=\; \max_{{P}=({p}_m)} \sum_{m\in\cM_t}{p}_m z_m
\\
&
\;\; \mbox{s.t. }  \sum_{m\in\cM_t} \phi(M_t {p}_m) \le M_t\rho_{t},\, \, \sum_{m\in\cM_t} {p}_m = 1 ,\, {p}_m \ge 0.
\nonumber
\end{align}
The uncertainty radius $\rho_t$ now changes with $t$; see~\refthm{thm:bias} (\refsec{ssec:ssapx}). 
In~\eqref{restrob}, we subsample from the discrete training dataset, and thus our analysis will primarily use related tools of probability; see the start of Appendix~\ref{apdx:bias}. 


Now suppose ${P}_t^*(\theta) = ({p}^*_{t,m}(\theta))$ is an optimal solution to~\eqref{restrob}.
Then a valid subgradient for $\hat{R}_t(\theta_t)$ is obtained as an expression analogous to
that in \refprop{prop:rgrad}(ii)
under appropriate substitutions related to $\theta_t$, $P_t^*$ and $\cM_t$.
\refthm{thm:bias} provides a bound on the error in using
the gradient expression in \refprop{prop:rgrad}(ii)
to approximate $\nabla_{\theta}{R}$ (the gradient of the true full-sample robust loss) as a function of the sample size $M_t$.
This bound on the bias vanishes only as $M_t\nearrow N$ as $t\tndi$.
Since fixed bias violates a basic requirement for SGD~\citep[Section~4.3]{bcn16}
that the gradient estimator
$\E[\nabla_{\theta}\hat{R}_t(\theta)] = \Theta(\nabla_{\theta} R(\theta))$
(using standard time complexity notation; see~\cite{sipser06}),
then the convergence of \eqref{ssgd} cannot be guaranteed when $M_t=M,\,\forall t$ where $M<N$.

The standard SGD ($M_t=M$) 
method uses diminishing step-sizes
to eliminate the impact of noise when gradient estimators are unbiased.
\citet[Section 5.1]{bcn16} show that for this same case keeping step-sizes fixed and increasing the mini-batch size directly is more advantageous because (informally) the larger
$M_t$ makes the gradient estimates more accurate.
In our case, the SGD algorithm~\eqref{ssgd} requires sample size growth even to eliminate bias, and this also provides a decrease in noise as a consequence.
Thus, when increasing $M_t$ it is no longer necessary to diminish the step size $\gamma_t$;
indeed, doing so negates the benefits of the extra work in computing gradients using
larger $M_t$. 
%
%
\refalg{alg:overall} therefore takes 
fixed-length steps $\gamma$. The maximum size $N$ is hit after a (large) finite number of iterations $\cT$, at which point we switch to a deterministic optimization algorithm,
e.g.,~\citeN{nd17}. 
\refthm{thm:main-gen} shows that our algorithm comes substantially close to a local minimizer 
in $\cT$ iterations when the sample set sizes $M_t$ are chosen to satisfy a
minimum-growth condition.
The analysis applies to the case where $l(\theta,\xi)$ possess Lipschitz gradients, but may otherwise be non-convex.
This allows our algorithm to be used in important cases when $l(\cdot,\xi_n)$ are non-convex, such as training deep learning models. 
\vskip -.05in
\begin{algorithm}[!htp]
	\caption{Dynamically Sampled Subgradient Descent}\label{alg:overall}
	{\bf Given}: Step size $\gamma$;~
	Sample size sequence $\{M_t\}_{t=1}^\cT$, $M_\cT=N$;~
Initial iterate $\theta_0$.
	\begin{algorithmic}[1]
		\FOR { $t=1,2,\ldots,\cT$}
		\STATE\label{step:enstart} 
		Sample $M_t$ indices {\em without replacement} uniformly from
		$\{1,\ldots,N\}$, and gather them in $\cM_t$ 
		\STATE Solve inner maximization problem to obtain optimal solution ${P}^*_{t}\qquad\qquad\quad$
		\COMMENT{see \refsec{ssec:robopt}}
		\STATE Set $G_t \gets \sum_{m\in\cM_t} {p}^*_{t,m} \nabla_{\theta} l(\theta_t,\xi_m)$
		\STATE Set $\theta_{t+1} \gets \theta_t - \gamma G_t \quad$
		\COMMENT{subgradient descent step}
		\STATE Increment $t\gets t+1$
		\ENDFOR
	\end{algorithmic}
\end{algorithm}
\vskip -.05in
Further addressed below is the question of choosing a {\em good} value for the step-length $\gamma$ and the sample growth sequence that obtains an optimal balance
between the added computational burden of each iteration and the expected reduction in the optimality gap.
We show for the case of strong-convex loss functions $l(\theta,\xi)$ that too slow a growth sequence, namely diminishing-factor growth (defined below), is inefficient;
while constant-factor growth sequences are efficient in the sense that the expected optimality gap drops at a rate proportional to the increase in the computational effort.
Extending these results to the convex and non-convex cases are subjects
of our ongoing research.

Our detailed analysis starts with an estimation of the effort needed to obtain an
exact solution to the inner maximization problem,
followed by two subsections establishing
various mathematical properties for our approach w.r.t.\ bias and convergence,
respectively, thus providing theoretical justification for the settings in our algorithm.

\input{inner_max}

\input{bias}

\input{convergence}

%% file: inner_max.tex
\subsection{Solving for $P^*(\theta)$ and $R(\theta)$}
\label{ssec:robopt}
Recall that the inner problem in~\eqref{restrob} for a \emph{random} subsample of size $M_t$, with a target $D_{\phi}$-divergence of $\rho_{t}$, uses a decision variable ${P} = ({p}_m)$ of dimension $M_t$. 
Writing the Lagrangian objective of~\refeq{restrob} as
\begin{align}
  &\cL(\alpha,\lambda, {P})
  = \sum_{m\in\cM_t} z_m {p}_m + \lambda \bigg(1 - \sum_{m\in\cM_t}{p}_m \bigg) \;\; 
  + \;\;\frac{\alpha}{M_t} \bigg(M_t  \rho_{t} - \sum_{m\in\cM_t}\Phi(M_t{p}_m) \bigg) ,
\label{roblag}
\end{align}
we then have the optimal objective value 
$\hat{R}^*_{t}(\theta) = \min_{\alpha\ge 0,\lambda} \max_{\hat{p}_m\ge 0} \cL(\alpha, \lambda, {P})$;
see~\cite{lbgr69}.
%
%
%
The optimal primal and dual variables can be obtained for various $\phi$ functions by a general procedure to solve Lagrangian formulations: see \textbf{Procedure 1} in the supplement.
%
%
The next result provides a worst-case bound on the computational effort required to obtain an $\epsilon$-optimal solution to~\eqref{restrob}.  
\begin{proposition} \label{prop:im_cmp_bnd}
For any $\phi$-divergence, 
\textbf{Procedure 1} finds a feasible primal-dual solution
$(\tilde{P}^*_t,\tilde{\alpha}^*,\tilde{\lambda}^*)$ to
problem~\refeq{restrob}
with an objective value $\tilde{R}^*_t$
such that $|\hat{R}^*_t(\theta) - \tilde{R}^*_t|<\epsilon$ with a worst-case computational effort bounded by
$O (M_t \log M_t + (\log \frac 1 {\epsilon} )^2)$,
where $\epsilon$ is a small precision parameter.
\end{proposition}
The machine-precision $\epsilon$ does not relate to any other parameter of the formulation or algorithm (e.g., $M_t, N,\rho$),
and it is required because \textbf{Procedure 1} solves two one-dimension bisection searches in sequence.
In the sequel we assume that $\epsilon$ is a fixed small value and \textbf{Procedure 1} returns the exact unique solution $(P_t^*,\alpha^*,\lambda^*)$ to~\eqref{restrob}, and that the computational effort is bounded by $O(M_t\log M_t)$.
%

\vskip -.1in

%% file: bias.tex
\subsection{Small-Sample Approximation of $\nabla_{\theta} R(\theta)$}\label{ssec:ssapx}
Let the mass vector $P^*=(p^*_1,\ldots,p^*_N)$ be the optimal solution to the full-data version of~\eqref{restrob},
i.e., with $M_t=N$, and
let
$P^*_t = (p^*_1,\ldots,p^*_{M_t})$ denote the optimal solution when restricted to any subset $\cM_t$.
\begin{assumption} \label{asm:phicont}
The 
$\phi$-divergence 
satisfies 
uniformly for all $s$ and $\zeta<\zeta_0$
the continuity condition (for constants $\zeta_0, \kappa_1, \kappa_2 >0$):
$|\phi(s(1+\zeta))-\phi(s)| \le \kappa_1 \zeta \phi(s) + \kappa_2 \zeta$.
\end{assumption}
This condition,
as described in \cite{shap09},
only allows for (local) linear growth in $\phi$, and it can be verified for many common $\phi$-divergences of interest including the modified $\chi^2$ metric and the KL-divergence metric.
Let $\E_{t}$ and $\Prb_{t}$ respectively denote expectation and probability w.r.t.\ the random set $\cM_t$.

\begin{theorem}\label{thm:bias}
  Suppose~\refasm{asm:phicont} and the assumptions of~\refprop{prop:rgrad} hold.
 Define $\eta_{t} = c (\frac{1}{M_t} -\frac{1}{N})^{(1-\delta)/2}$ for small constants $c, \delta >0$, and set the $D_{\phi}$-target
  in~\eqref{restrob} to be $\rho_{t} = \rho + \eta_{t}$.
Then, there exists a small positive $M'$ defined in \eqref{eq:Mprime}
and of order $o(N)$
such that, for all $M_t \ge M'$, the subgradient $\nabla_{\theta}\hat{R}_{t}(\theta)$ and
  full-gradient $\nabla_{\theta} R(\theta)$ satisfy for any $C<\infty$ and $1-\bar{\tau}_t= O(\eta_{t}^{2\delta/(1-\delta)})$: 
 $$\quad \Prb_{t} (\eta_{t}^{-2} \|  \nabla_{\theta}\hat{R}_{t}(\theta) - \nabla_{\theta}R(\theta) \|^2_2  \le  C ) \ge \bar{\tau}_t.$$ 
\end{theorem}

In the sequel, we use the following corollary of~\refthm{thm:bias} that
follows from Theorem 17.4 in~\cite{jacod2004probability}.
\begin{corollary}\label{cor:bias}
       If the conditions for~\refthm{thm:bias} are satisfied, then
       $       \| \E_{t} [\nabla_{\theta}\hat{R}_{t}(\theta)]  -
       \nabla_{\theta}{R}(\theta)\|^2_2 = O(\eta_{t}^2).$
\end{corollary}
The proof of~\refthm{thm:bias} (in supplement) includes
exact expressions for the constants such as $\bar{\tau}_t$, whose dependence on the
constant $C$ and on the magnitude of $\nabla_\theta R(\theta)$ is made explicit
in Lemma~\ref{lem:objbias}.
Motivating the form of $\eta_t$, note that sampling a subset of size $M$ from a larger finite set of size $N$ induces variance terms
of the form $(\frac 1 M-\frac 1 N)$ in place of the standard $\frac 1 M$ (see Appendix~\ref{apdx:bias}).
An outline of the proof starts by constructing $\tilde{P}^*$, 
a restriction of the (unique) 
$P^*$ 
onto the (random) subset $\cM_t$ 
in the restricted problem~\eqref{restrob}, where
$\tilde{p}^*_m \propto {p^*_m},\;\forall m\in\cM_t$. 
The assumptions 
ensure with probability at least $\bar{\tau}_t$ that $\tilde{p}^*_m \neq {\mathbf{0}}$.
With the same high probability, $\tilde{P}^*$ is also a feasible solution to~\eqref{restrob} when $\rho_{t}$ is inflated as assumed.
Next, we establish that its objective value is within $\eta_{t}$ of the optimal with high probability, which yields the desired result.
%

\begin{remark}\label{rmk:smpl}
The squared bias in~\refthm{thm:bias} is more generally of the order of 
$\eta^2$ with $M_t$ replaced by $|\cM_t|$ in its expression, and $|\cM_t|$ is the number of
support points. If 
$M_t$ support points are sampled {\em
  with replacement}, the set $\cM_t$ will have an expected number of unique values $\E|\cM_t| = 1 + 1/2+1/3+\ldots=O(\log M_t)$, obtained by adding the expected number of additional samples needed to see a new support sample. Hence, sampling with replacement is inefficient since it results in a slow reduction in bias.
\end{remark}

%% file: convergence.tex
\subsection{Convergence of \refeq{ssgd}}
\label{ssec:cvg}
We now analyze the convergence of \refalg{alg:overall} under the following additional assumptions.
\begin{assumption}\label{asm:props}
\noindent(i) A lower bound $R_{\inf}$ exists for the robust loss function $R(\theta)\ge R_{\inf}$, $\,\,\forall \theta\in\Theta$.\\
\noindent(ii) The variance of
$\nabla \hat{R}_t(\theta)$ with subsample size $M$ obeys 
$\E [ \|\nabla\hat{R}(\theta) - \E [\nabla\hat{R}(\theta)]\|^2_2 ] \le   C(\frac 1 {M} - \frac 1 N)$.\\
\noindent  (iii) The loss functions $l(\theta,\xi_n)$ are $c$-strongly convex.
\end{assumption}
Assumption~\ref{asm:props}(ii) ensures that the variance of $\nabla \hat{R}_t(\theta)$ follows what can be expected for sampling without replacement from any finite set~\cite{wilks}. 
Combining this with the order of the bias (\refcor{cor:bias}) allows us to progressively decrease the mean squared error. 
%
%
%
Since $N$ is finite, any scheme to strictly increase $M_t$ as
$t\tndi$ will eventually end at an iterate $\cT <\infty$ where
$M_{\cT}=N$, at which point we switch to the deterministic full-support optimization~\eqref{absfmln}.
Seen in this light, \refalg{alg:overall} is not guaranteed to converge
by the $\cT$-th iteration. We can nevertheless provide a 
guarantee on the performance of the method over any loss function with Lipschitz gradients.
%
\begin{theorem}\label{thm:main-gen}
  Suppose the constant step size $\gamma_t=\gamma$ satisfies
  $\gamma \le \frac{1}{2L}$, the loss
  functions satisfy~\refasm{asm:props}(i) and (ii), and the
  conditions of~\refthm{thm:bias} hold.
Further assume
the gradient $\nabla_{\theta} R(\theta)$ is $L$-Lipschitz.
Then, at termination,
  \begin{equation}\label{eq:maingen}
\sum_{t=1}^{\cT} \left\|\nabla_{\theta}R(\theta_t)\right\|^2_2 \le \frac {R(\theta_0) -
  R_{\inf}}{\frac {\gamma} 2 (2-L \gamma)} + C \frac {L\gamma +1 }
{2-L \gamma} \sum_{t=1}^\cT \eta_{t}^2. 
  \end{equation}
\end{theorem}
%
%
\refthm{thm:main-gen} establishes that the sum of the gradients of $R(\theta_t)$
at iterates visited by the algorithm is bounded above by, in particular,
$\sum_{t=1}^\cT (\frac 1 {M_t} - \frac 1 N)^{(1-\delta)}$.
If this summation remains finite as $\cT\tndi$, then the
upper bound of~\eqref{eq:maingen} remains finite,
and thus the gradients $\|\nabla_{\theta} R(\theta_t)\|^2$
at the iterates converge to $0$; in other words, the algorithm converges to a
local optimal solution.
The summation can converge for $M_t$ increasing moderately, such as at a
polynomial rate.

\refthm{thm:main-gen} assumes that the gradient $\nabla_{\theta}R(\theta)$ of the robust loss 
is Lipschitz continuous. The gradients of such extreme value functions are 
in general not Lipschitz if the objective function is Lipschitz. For example, a linear objective $l(\theta,\xi_n) = \theta^t\xi_i$ leads to $R(\theta) = \max_p \sum_i p_i \theta^t\xi_i$ and when maximized over a {\em polyhedral}
constraint set (e.g., the probability simplex constraints of~\eqref{absfmln}) it will not preserve the $0$-Lipschitzness of the objective functions, because in this case the optimal solutions $P^*$ are picked from the discrete set of vertices of the polyhedron and thus $\nabla_{\theta}R(\theta)$ is piecewise discontinuous.
Our assumptions from~\refprop{prop:rgrad} yield an inner maximization with a non-zero linear objective over a \textit{strictly convex} feasible set. The desired smoothness can then be obtained with some additional conditions on the loss functions $l(\theta,\xi_i)$.
~\refprop{prop:rgradlip} provides one such condition where the Lipschitzness of $\nabla_{\theta} R(\theta)$ follows from the Hessian of $R(\theta)$ being bounded in norm, which is often satisfied by common statistical learning losses such as log-logistic and squared losses of linear models over compact spaces.

\begin{proposition}\label{prop:rgradlip}
	Assume the conditions in~\refprop{prop:rgrad} hold.
	Further suppose that the Hessians $\nabla^2_{\theta} l(\theta,\xi_n)$ exist $\forall \theta$ and each $\xi_n$, and are bounded in Frobenious norm $\|\nabla^2_{\theta} l(\theta,\xi_n)\|_F\le L$, $\forall \theta, n$.
	Then, the robust loss also follows $\|\nabla^2_{\theta} R(\theta)\|_F \le M $ for some positive $M< \infty$.
\end{proposition}

A key consideration then is to obtain $\theta_\cT$ as close as possible to the minimizer $\theta_{\rob}$ that attains $R_{\inf}$.
The tradeoff in \eqref{eq:maingen} suggests that increasing $M_t$ aggressively will
lead to smaller gradients at termination, but this will also increase the computational effort in each iteration.
In the remainder of the section, we study this tradeoff under the \refasm{asm:props}(iii), which \refthm{thm:main} will show yields the strong convexity of $R(\theta)$.
Consequently, it also provides a unique minimizer $\theta_{\rob}$ for~\eqref{absfmln} that satisfies $R_{\inf} \DefAs R(\theta_{\rob})$.
We therefore seek to ``optimally'' increase the value of $M_t$ to strike a balance between the increased computational effort
and the value of $R(\theta_\cT)$ and its gradient at termination.

Our notion of efficiency will be developed w.r.t.\ the total computational effort $W_t$
that is expended up until iterate $t$, which is the sum of the amount of individual work
$w_s$ in each iterate $s\leq t$.
From the discussion following~\refprop{prop:im_cmp_bnd}, we have that $w_t=O(M_t\log M_t)$. 
Defining the ratio $\nu_t \DefAs M_{t+1}/M_{t}$ as the {\em growth factor} of the sequence \(\{M_t\}\),
we consider two important cases, namely~~(i) {\em Diminishing}-factor growth: if $\nu_t\searrow 1$ as $t\uparrow$, e.g., the
polynomial growth of $\nu_t=1+\frac{1}{t}$;~~and~~(ii) {\em Constant}-factor growth: if
$\nu_t =\nu > 1,\,\,\forall t$.
The proposed algorithm is employed to a maximum of $\cT$ iterations, where $\cT = \inf\{t: N = M_0 \prod_{s=1}^t \nu_s\}$.
For constant growth sequences, we have $\cT = \log (N/M_0) / \log \nu$, while the slower diminishing growth sequences have a much larger $\cT$.  
Our final result characterizes the
rate at which the expected optimality gap
$E_{t+1} \DefAs \E_{t} [ R(\theta_{t+1})] - R(\theta_{\rob})$
decreases as $M_t\nearrow N$.
Recall that
$\delta$ defines the parameter $\eta_t$ in~\refthm{thm:bias}.

\begin{theorem}\label{thm:main}
  Suppose all the conditions of~\refthm{thm:main-gen} are satisfied and \refasm{asm:props}(iii) holds.  Then the function $R(\theta)$ is $c$-strongly convex.
Further suppose that $\gamma \le \min\{\frac{1}{4L} , 4c\}$ and let $r = 1- \frac{\gamma}{4c}$.
  We have:~~(i) If $M_t = M_0 \nu^t $ with parameter $1< \nu < r^{-1/(1-\delta)}$, then for $t\le \cT$, $W_t E_{t+1} \le K_1 t \nu^{t\delta}$ for a constant $K_1$;~~(ii) If $M_t = M_0\nu^t$ with parameter $\nu \geq r^{-1/(1-\delta)}$, then for $t\le \cT$, $W_t E_{t+1} \le K_2 t (r\nu)^{t}$
  for a constant $K_2$;~~(iii) If $M_t$ is a diminishing growth sequence, then $E_{t+1} = o(W^{-1}_t)$.
\end{theorem}

The proof of Theorem~\ref{thm:main} (in the supplement) includes exact expressions for the constants such as $K_1$ and $K_2$.
To understand this result, first note from~(\ref{onestep})
that if the full batch-gradient method is applied in each iteration ($M_t=N$),
a strongly-convex objective $R$ would enjoy a linear (i.e., constant factor)
reduction of size $r$ in the error $R(\theta_t)-R(\theta_{\rob})$ for a step-size $\gamma$ chosen to satisfy the conditions of~Theorem~\ref{thm:main}.
The average optimality gap in our algorithm can be written as a sum of this deterministic error and an additional term representing the stochastic error induced by the subsampling of the support.
\refthm{thm:main}(iii) establishes that any general diminishing-factor growth of $M_t$ will lead to the stochastic error decreasing to zero much slower than the geometric drop in the deterministic error, and thus the stochastic error dominates. Consequently, there is a suboptimal reduction in the optimality gap w.r.t.\ the total computational effort $W_t$, which grows in proportion to the sample size, as opposed to constant-factor growth sequences.

Constant factor sequences can, however, trade off the rate of reduction in stochastic error
against the drop in deterministic error in strongly convex functions.
Specifically,
parameter values $\nu \in (1, r^{-1/(1-\delta)})$ produce the best balance between the optimality gap $E_{t+1}$ and computational effort $W_t$,
with the upper bound on their product growing at the slowest of the three cases at a near-linear rate w.r.t.\ iteration count $t$ (recall $\delta$ is arbitrarily small).
In contrast,
if $\nu\geq r^{-1/(1-\delta)}$, the deterministic error now drops slower than the stochastic error, and in a manner that the product $E_{t+1}W_t$ escapes to infinity
at a geometric rate w.r.t.\ $t$ which speeds up the farther $\nu$ is from this critical value. 
Since $r$ depends on $c$ and $L$ through $\gamma$, it
can be difficult
to identify in practice.
Meanwhile, these results can be used to determine an (optimal) sequence of $M_t$ and minimal total effort $W_t$ needed to achieve a desired level of accuracy (see the supplement).

\citeN{rag18,hpt17} analyze increasing-batch SGD algorithms (with unbiased gradient estimators) for the case of convex and non-convex Lipschitz-gradient functions,
but the results are less striking.
They conclude that, since here the full-batch deterministic iterates exhibit sub-linear convergence (i.e., diminishing factor reduction),
increasing the batch size $M_t$ with a constant factor results in the stochastic error decreasing too quickly.
We expect a result similar to~\refthm{thm:main}(iii) to hold in our formulation for such $l(\theta,\xi_n)$, 
in that both constant and diminishing growth sequences lead to the net error $E_t$ decreasing too slowly compared to the effort $W_t$. 


%% file: expt.tex
\section{Experimental Results}
\label{sec:expt}
\begin{figure*}[tbp]
	\vskip -0.2in
	\begin{center}
		\includegraphics[width=0.40\textwidth]
		{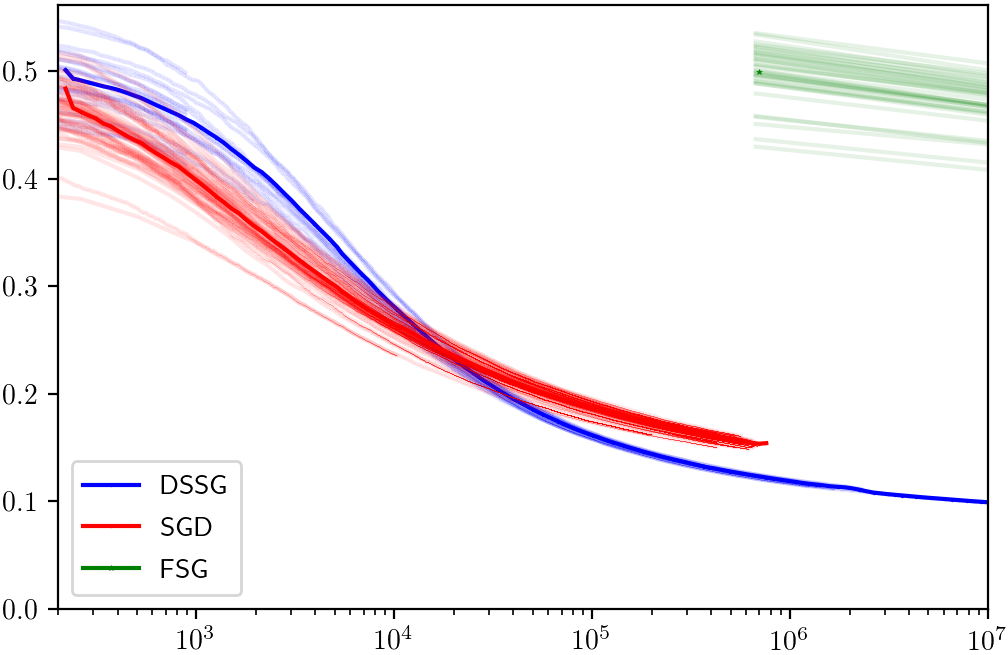}
		\hskip 0.2in
		\includegraphics[width=0.26\textwidth]{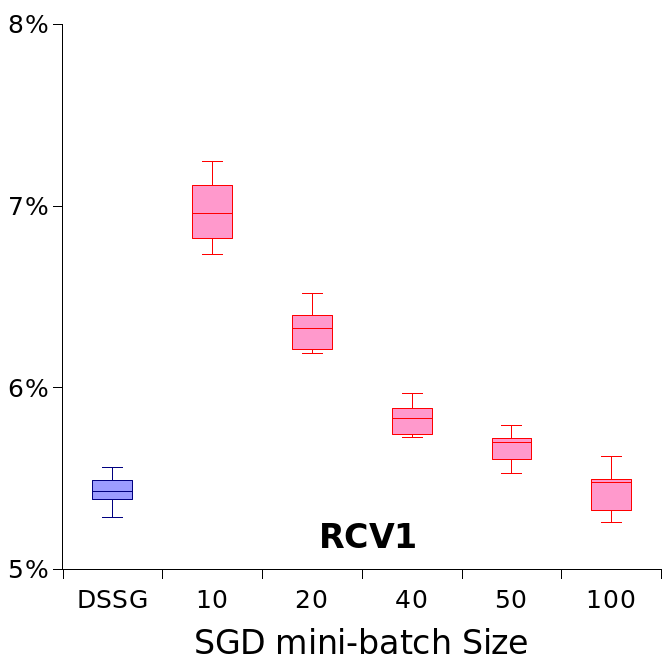}
		\hskip 0.2in
		\includegraphics[width=0.26\textwidth]{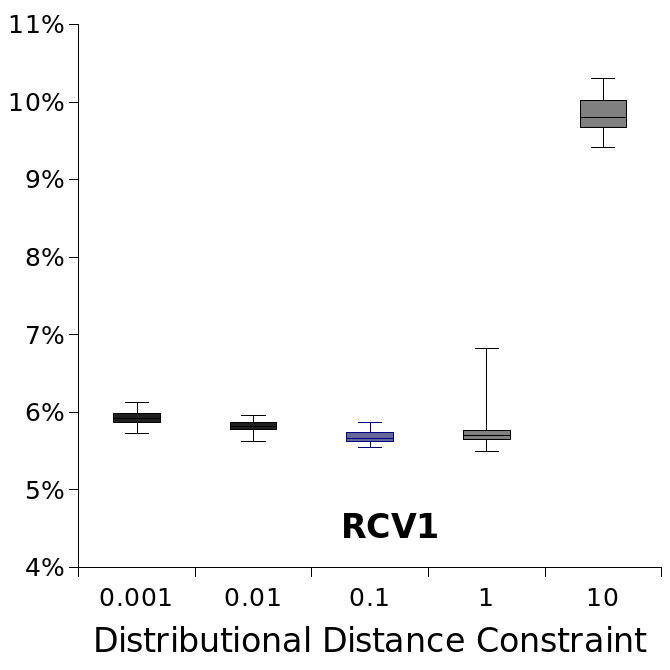}
		\caption{On the left, a comparison of DSSG (blue), standard SGD (red) and FSG (green) on the fraction of misclassification in testing ($y$-axis)
			versus cumulative samples ($x$-axis) over the RCV1 dataset with $\rho=0.1$ and log-scale $x$-axis.
			In the center, an evaluation of the bias suffered by SGD for different batch sizes compared with that of our DSSG-solved formulation.
			On the right, a study of the performance of the DRO formulation for increasing values of $\rho$.
		}
		\label{fig:rcv1}
	\end{center}
	\vskip -0.3in
\end{figure*}

Numerous experiments were conducted to empirically evaluate our
dynamically sampled subgradient descent (DSSG) algorithm.
We report on two sets of experiments,
the first of which compares DSSG with competing algorithms to solve the DRO formulation~(\ref{absfmln})
while the second evaluates the efficacy of the DRO approach as a viable alternative to regularized ERM in producing models with good generalization. 
Following~\citeN{nd17}, all examples use a logistic regression loss function $l(\theta, (x,y)) = \log (1 + \exp(-y \theta^t x))$, 
where $x$ represents samples with $d$ features, and $y$ represents the binary class labels $\pm 1$.
Additional results and details can be found in the supplement.

For the base parameters of our new DSSG algorithm in both sets of experiments,
we set the initial sample size $M_0=1$, \emph{constant-growth factor} $\nu = 1.001$, and fixed step lengths $\gamma=0.5$.
\citeN{bwm16} and \citeN{nd17} consider at length the question of setting the parameter $\rho$ for the set of measures $\cP$,
providing as a broad guideline for binary classification with logistic models that $\rho = O(\sqrt{d/N})$.
Given the values of $d$ and $N$ in our experimental datasets, we set $\rho=0.1$.
Though parameter $\delta$ appears prominently in the inflation of $\rho$ to individual $\rho_t$ as defined  in~\refthm{thm:bias}, the result requires only that $\delta$ be a small positive constant.
Since the experimental results are not sensitive to $\delta$, we set $\delta=0.01$.

\noindent\textbf{Solving DRO Formulation.}
We now compare our DSSG algorithm (Algorithm ~\ref{alg:overall}) with the full-support gradient (FSG) algorithm of~\citeN{nd17}.
We also include the standard SGD (fixed minibatch $M_t=M$) method to gauge the impact of the bias shown in~\refthm{thm:bias}.
In~\reffig{fig:rho0.1:HIV+nd16} of the supplement, we further include the SGD method of~\citeN{nd16} for a small dataset to demonstrate that the method was not competitive. 
Each of these methods was re-implemented since full source code was not made available with the corresponding publications.

The main body of the paper presents a comparison of the various solution methods over the Reuters Corpus Volume 1 (RCV1) dataset~\cite{RCV1},
with the supplement containing detailed comparisons over three additional datasets.
The RCV1 dataset comprises
$804414$
samples each with
$47236$
features, thus representing a substantially large and challenging dataset,
the purpose of which is to classify each sample article as either belonging to a corporate/industrial category or not based on its content.


All algorithms sampled the initial $\theta_0$ uniformly from the hypercube $[-1,1]^d$,
and the inner-maximization formulation is solved to within $\epsilon$-accuracy where $\epsilon = 10^{-7}$.
The SGD method is run with mini-batch size $M=10$ and a stepsize sequence of $\gamma_t = 0.5 * (5000/(5000+t))$ for all experiments.
The step lengths of the FSG algorithm are determined by the LBFGS-B algorithm with a maximum step length of $0.5$. 
Beyond the base parameter settings for DSSG, the supplement contains comparisons corresponding to those in \reffig{fig:rcv1} for DSSG with $\rho=\{0.01, 0.5, 1.0\}$. 

Our 
results are based on comparisons over
$10$ experimental runs,
where each run uses a different random partition of the original dataset 
into training (80\%) and testing (20\%) datasets.
All experiments were implemented in Python 3.7 and run on a 16-core 2.6GHz Intel Xeon processor with 32GB memory.
Each method monitors the performance of the current model in correctly classifying the set-aside test set and stops if the average of the last $20$ misclassification fraction values
does not improve more than $1\%$ when compared with the average of the previous $80$ evaluations.

\reffig{fig:rcv1} (left) presents our empirical results from $10$ experimental runs for DSSG, FSG and SGD comparing the fractional misclassification performance
over the testing dataset as a function of the cumulative sample size used by each algorithm (log scale).
(We note there are no significant differences between these performance results as functions of cumulative samples and CPU time,
consistent with Proposition~\ref{prop:im_cmp_bnd};
see the supplement.)
As expected, FSG is simply not competitive with DSSG, significantly so.
Further,
the impact of the bias induced by the fixed $M=10$ is noticeable
here
on SGD; \refthm{thm:bias} gives the relationship between the bias suffered by the standard SGD method and the batch size $M$.
In contrast, when our DSSG algorithm is run with such a small growth factor of $\nu=1.001$, the iterations run with $M_t=1$ until $t\approx 400$ when it then rises to $M_t=2$.
This allows DSSG to initially enjoy the benefits of fast objective value reduction similar to a standard SGD run with fixed steps, but then eventually eliminate the bias introduced 
for
standard SGD and thus also enjoy convergence.

The center plot in~\reffig{fig:rcv1} provides a comparison of the performance of SGD over various mini-batch sizes against our DSSG algorithm and shows that the bias drops to insignificance only as $M\rightarrow 100$.
Therefore, our DSSG algorithm avoids the expense of the hyper-parameter tuning of the batch size of standard SGD for bias reduction.
This is further supported by the results in Figure~\ref{fig:rcv1} (right) that illustrate the relative insensitivity of the output of our DSSG algorithm to values of $\rho$ from $0.001$ to $1$.
The main advantages of our DRO algorithm include that it does not need any such tuning and it efficiently provides a solution to~\eqref{absfmln}. 

\begin{figure}[htbp]
	\vskip -0.1in
	\begin{center}
		\includegraphics[width=0.435\columnwidth]{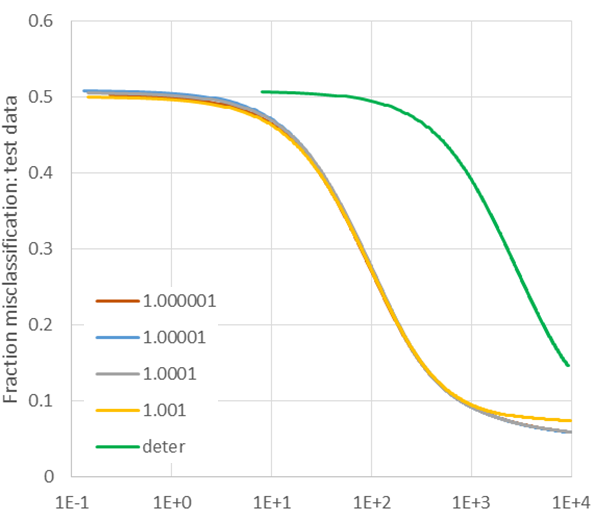}\hskip .2in
		\includegraphics[width=0.465\columnwidth]{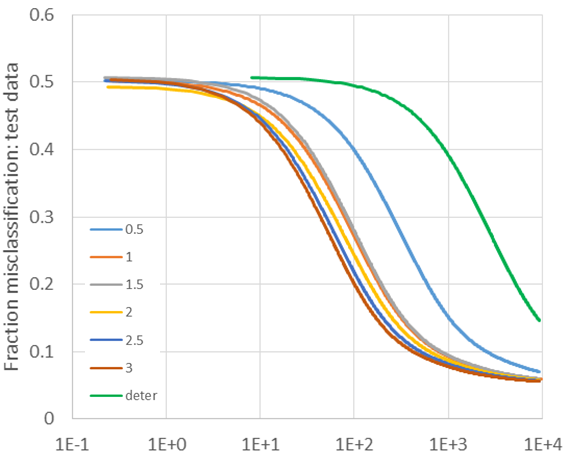}
		\caption{The {\em average} misclassification performance of
			the FSG (green lines, `deter' in legend) and DSSG (other colored lines)
			algorithms over the RCV1 dataset: (left) keeping $\gamma=1.0$ fixed and varying
			$\nu$ from $1.000001$ to $1.001$; and (right) keeping $\nu=1.001$ fixed and varying $\gamma$ from $0.5$ to $3$.
			Log-scale computation time (in seconds) on $x$-axis.}
		\label{fig:rcv1.dssg.params}
	\end{center}
	\vskip -0.2in
\end{figure}

We
next
consider
empirical results to study the impact of two key
parameters of~\refalg{alg:overall}, the sample growth $\nu$ and step length $\gamma$.
Broadly, we expect these settings to impact the computation times and solution quality. 
%
\refthm{thm:main} establishes a bound on the sample size growth factor $\nu$ and step length $\gamma$ for the iterates
to converge geometrically with total computational effort/time, where the geometric factor depends on $\gamma$.
Figure~\ref{fig:rcv1.dssg.params} (left) contrasts the performance of our DSSG algorithm keeping $\gamma=1.0$ fixed while varying $\nu$.
Our results show that the performance is insensitive to $\nu$ for values smaller than $1.001$. 
While DSSG takes a large number of SGD-like iterations initially, 
the additional noise inherent in these iterations prevent any faster convergence in terms of total computational effort.
In each case, DSSG converges much faster, taking up to two orders of magnitude less effort than FSG.
Although no finite-time performance penalty is evident for small $\nu$ in the results provided here, we anticipate somewhat of a degradation in larger and non-convex models.
Figure~\ref{fig:rcv1.dssg.params} (right) fixes $\nu=1.001$ and varies $\gamma$, showing that $\gamma$ does have more effect, but DSSG is relatively insensitive to the chosen step length beyond $\gamma = 0.5$.

\noindent\textbf{Generalization.}
We now 
investigate whether 
the DRO formulation~\eqref{absfmln} can improve generalization of learning models,
building on the above experimental framework.
We compare the results of the DRO formulation, solved via our DSSG, against an ERM-trained model that is regularized via $10$-fold cross validation (CV).
The \emph{$k$-fold CV} procedure partitions the full training dataset into $k$ equal parts and trains a {\em regularized} model over each dataset formed by holding out one of the $k$ parts as the \emph{validation} dataset.
The term $\lambda\|\theta\|_2^2$ regularizes the ERM loss objective, and
a fixed-batch SGD (size $10$) is used for each combination of partition and $\lambda$ values.
Enumeration is used to find the optimal $\lambda$ 
from a grid of $20$ points in the range $[10^{-6}, 10^6]$,
starting with $10^6$ and backtracking till the \emph{average} performance over the $10$ validation datasets does not improve for three $\lambda$ enumerations.

Experiments were conducted over $14$ public-domain datasets
from
UCI~\cite{UCI}, OpenML~\cite{openml} and SKLearn~\cite{RCV1}, with
sizes ranging from $O(10^2)$ to $O(10^6)$. 
Table~\ref{table:gen} presents a comparison of the test misclassification produced by our DSSG algorithm and the regularized ERM algorithm at termination for the $14$ datasets.
The 95\% confidence intervals are calculated over $10$ permutations of the datasets into training (80\%) and testing (20\%) sets.
For each dataset, the method that produces the best generalization error
(within the specified confidence intervals)
is highlighted in bold.
As evident from the table, the DRO method produces models of equal or better quality as produced by the regularized ERM formulation in most datasets, and tends to work better with the larger datasets.
%
\input{results_table}

Recall that DSSG provides this level of performance by solving a single instance of the DRO formulation~\eqref{absfmln}, thus avoiding the burdensome $10$-fold CV enumeration.
Table~\ref{table:gen} also provides the average CPU time in seconds recorded over the $10$ permutations.
The average time taken by the ERM $10$-fold regularization in solving its formulation multiple times in a serial computing mode to identify the best $\lambda$
exceeds that taken on average to solve the DRO formulation by two to three orders of magnitude.
(We note that the computational benefits of DRO would be even larger if all $\lambda$ values over all grid points were enumerated; see the supplement.)
The computation time of a single DRO run is of the same order as that of a single ERM run.
This indicates a significant computational savings in using DRO because of the elimination of the expensive hyper-parameter tuning step.
%
%
%

We also find a significant amount of variability in the optimal $\lambda$ over the $14$ datasets, with no evident pattern relating to dataset characteristics, 
which highlights the need for ERM computations over a wide range of $\lambda$ values for each dataset.
The performance of the DRO formulation requires an adequate choice of $\rho$ (set to $0.1$ here),
and the results for the smaller datasets might be further improved by a very small number of additional runs with different $\rho$ values while still maintaining significant computational benefits over ERM.


\noindent\textbf{In Summary.}
Our empirical results support our theoretical results and show that DSSG provides the same quality of performance as FSG but with orders of magnitude less computational effort,
while also outperforming SGD and not requiring its hyper-parameter tuning.
Our results further show that DSSG renders models of comparable or better quality as those from regularized ERM but with orders of magnitude less computational effort,
and thus provides a strong alternative ML approach to improve model generalization.

%% file: results_table.tex
\begin{table}[thbp]
	\vskip -0.1in
	\centering
	\begin{tabular}{l||r|r||r|r}
		\textbf{Dataset}&\multicolumn{2}{|c||}{\textbf{Test Misclassified (\%)}}	
		&\multicolumn{2}{|c}{\textbf{CPUTime (sec)}}\\
		\hline
		&\multicolumn{1}{c|}{DRO} &\multicolumn{1}{c||}{ERM}	&DRO	& ERM\\
		\hline\hline
		tr31.wc$^{\dag}$		&$\mathbf{2.6 \pm 0.3}$	&$\mathbf{2.7 \pm 0.1}$		&6 &987 \\ 
		hiv1$^{\ast}$			&$5.9 \pm 0.1$	&$\mathbf{5.6 \pm 0.0}$		&10	&1012 \\ 
		gina\_prior$^{\dag}$	&$13.0 \pm 0.5$	&$\mathbf{11.8 \pm 0.1}$	&12	& 1147 \\ 
		la1s.wc$^{\dag}$		&$\mathbf{8.3 \pm 0.2}$	&$\mathbf{8.5 \pm 0.0}$		&12	&2456 \\ 
		gina\_agnostic$^{\dag}$&$13.9 \pm 0.3$	&$\mathbf{12.6 \pm 0.1}$	&13	& 1765\\
		bioresponse$^{\dag}$	&$24.2 \pm 0.4$	&$\mathbf{21.6 \pm 0.2}$	&16	& 2791\\ 
		ova\_breast$^{\dag}$	&$3.0 \pm 0.1$	&$\mathbf{1.8 \pm 0.1}$		&17	& 4310\\ 
		fabert$^{\dag}$			&$\mathbf{9.8 \pm 0.0}$	&$10.1 \pm 0.0$		&20	& 4128\\ 
		dilbert$^{\dag}$		&$\mathbf{1.2 \pm 0.1}$	&$\mathbf{1.1 \pm 0.0}$	&33	& 8543\\ 
		adult$^{\ast}$			&$\mathbf{16.6 \pm 0.1}$	&$\mathbf{16.7 \pm 0.0}$	&36	& 2542\\ 
		imdb.drama$^{\dag}$		&$\mathbf{36.2 \pm 0.1}$	& $37.1 \pm 0.0$	&89	& 19436\\
		guillermo$^{\dag}$		&$\mathbf{30.2 \pm 0.5}$	& $\mathbf{30.7 \pm 0.1}$ &116 & 31547\\ 
		riccardo$^{\dag}$		&$1.6 \pm 0.0$	&$\mathbf{1.5 \pm 0.0}$		&120 & 86575\\ 
		rcv1$^{\ddag}$			&$\mathbf{5.4 \pm 0.0}$	&$5.6 \pm 0.0$		&543 & 701843\\ 
		\hline
	\end{tabular}
\caption{Comparison of the DRO and regularized ERM formulations over $14$ publicly available machine learning (ML) datasets,
from UCI$^{\ast}$ \cite{UCI}, OpenML$^{\dag}$ \cite{openml} and SKLearn$^{\ddag}$ \cite{RCV1},
arranged in increasing training set size.
The first pair of columns provides a 95\% confidence interval of the percentage misclassified over withheld test datasets, with the best-performing method highlighted in bold.
The second set of columns provides the average CPU time taken to solve each formulation.}
\label{table:gen}
\vskip -0.2in
\end{table}

%% file: appendix.tex
\newpage

\appendix
\section*{Supplement: \textit{Efficient Stochastic Gradient Descent for Learning with Distributionally Robust Optimization}}
This supplement contains additional results and technical details in support of the main body of the paper.
\refsec{apdx:proofs} covers further theoretical results and all the proofs of our theoretical results, together with associated technical details.
\refsec{apdx:expt} covers further empirical results from our numerical experiments and all related technical details.

\section{Proofs of Results} \label{apdx:proofs}
\refsec{apdx:phibasics} provides the proofs of~\refprop{prop:rgrad} and \refprop{prop:rgradlip},
whereas \refsec{apdx:getPstar} describes \textbf{Procedure 1} and presents the proof of~\refprop{prop:im_cmp_bnd}.
The proof of~\refthm{thm:bias} is considered in~\refsec{apdx:bias}, while
\refsec{apdx:cvg} provides the proofs of \refthm{thm:main-gen} and~\refthm{thm:main}.
The following additional notation is used extensively: Throughout, we write $a^T$
where $T$ denotes the transpose operator.
For a sequence of positive-valued
random variables (r.v.s) $\{A_n\}$, we write $A_n=o_p(1)$ if $A_n \inP 0$ as $n \to \infty$;
We write $A_n=\mathcal{O}_p(1)$ if $\{A_n\}$ is stochastically bounded, that is, for a given $\epsilon>0$
there exists $c(\epsilon) \in (0,\infty)$ with $\mathbb{P}(A_n < c(\epsilon)) > 1 - \epsilon$
for sufficiently large $n$;
If $\{B_n\}$ is another sequence of positive-valued r.v.s, we write $A_n = \mathcal{O}_p(B_n)$ if $A_n/B_n= \mathcal{O}_p(1)$.

\input{apdx_im}

\input{apdx_bias}

\input{apdx_cvg}

\section{Experimental Results} \label{apdx:expt}
\input{apdx_expt}

%% file: apdx_im.tex
\subsection{Uniqueness of $P^*(\theta)$, Form of $\nabla_{\theta} R(\theta)$, and Boundedness of $\|\nabla^2_{\theta} R(\theta)\|_F$}\label{apdx:phibasics}
We start by establishing some basic properties of $\phi$-divergences.

\begin{lemma} \label{lem:unif_small_phi}
	Consider mass functions over a support of size $M_1$.
	Suppose 
	\begin{align*}
	{\cal P}(M_2) =\bigg\{P &= (p_m) \in \real^{M_1} \,\,\Big|\,\,\sum_m p_m = 1, \,\, p_m\ge 0,\,\,
\quad p_{m'} = 0, \,\, \forall m'=M_2+1,\ldots,M_1\bigg\}
	\end{align*}
	to be the subset of all mass functions that place positive mass in only a subset of size $M_2<M_1$ of the full support size $M_1$.
	Define $U_{M_2} \DefAs  \left(\underbrace{\frac 1 {M_2},\ldots,\frac 1 {M_2}}_{M_2},\underbrace{0,\ldots,0}_{M_1-M_2}\right)$.
	Then 
	\[
	U_{M_2} =\arg\min_{P\in{\cal P}(M_2)} D_{\phi}(P, U_{M_1}).
	\]
\end{lemma}
\begin{proof}
	For any $P\in{\cal P}(M_2)$, we have
	\begin{align*}
	D_{\phi}(P, U_{M_1}) &= \sum_{m\le M_2}\frac{1}{M_1} \phi(M_1{p}_m) + \sum_{m>M_2} \frac{1}{M_1} \phi(0) \\
	&\hspace*{-0.4in} 
= \frac{M_2}{M_1} \sum_{m\le M_2} \frac{1/M_1}{M_2/M_1} \phi(M_1 p_m) + \sum_{m>M_2} \frac{1}{M_1} \phi(0) \\
	&\hspace*{-0.4in} 
\ge \frac{M_2}{M_1} \phi\left( \sum_{m\le M_2} \frac{1/M_1}{M_2/M_1} M_1 p_m \right) +  \sum_{m > M_2} \frac{1}{M_1} \phi(0)
\\
	&\hspace*{-0.4in} 
= D_{\phi}( U_{M_2},U_{M_1}) ,
	\end{align*}
	where in the last step Jensen's inequality is applied to the convex function $\phi$.
\end{proof}

\reflem{lem:unif_small_phi} shows that the $D_{\phi}$ distance between
the uniform distribution $U_N$ which assigns equal mass to all
$N$ support points and any probability mass function (pmf)
on an $M$-subset is minimized by the probability distribution
$U_M$, and the minimal distance is given by
$$D_{\phi}(U_{M}, U_{N}) = \frac M N \,\,\phi\left(\frac
N M\right) + \frac{(N-M)} N \,\,\phi(0).$$
%
If $D_{\phi}(U_M,U_N) > \rho$, then the
feasibility set admitted by the $D_{\phi}$-constraint
in
$$\cP =\Big\{ P \, | \, D(P, P_{b}) \le \rho , \int dP(\xi) = 1,  P(\xi) \ge 0 \Big\}$$
does not admit any pmf with mass only on $M$ support points.
Note that $D_{\phi}(U_{M}, U_{N})$ is also decreasing in $M$, because $D_{\phi}(U_M,U_N)$ is a convex mixture of $\phi(0)$ and $\phi(N/M)$.
Since $\phi(s)$ is strictly increasing for $s>1$, we have that $\phi(N/M_2) > \phi(N/M_1)>1$ for $M_1>M_2$.
Hence, any convex combination of the form above will satisfy $D_{\phi}(U_{M_2},U_N) \ge D_{\phi}(U_{M_1},U_N)$, and thus, 
for a given $\rho$, there exists an $M'(\rho)$ such that the constraint in~\eqref{restrob} only admits pmfs with mass on $M\ge M'(\rho)$ support points.
As a consequence, the optimal solution of the problem~\eqref{restrob} may lie on the intersection of up
to $N-M'(\rho)$ hyperplane constraints of the form $p_i=0,\,\,\forall i=M'(\rho)+1,\ldots,N$.
This may in turn lead to degenerate optimal solutions for some objective coefficients $l(\theta,\xi_i)$ for the linear program~\eqref{restrob}.
We preclude this possibility
by assuming
that the parameter $\rho$ in~\eqref{restrob} satisfies
\begin{equation}\label{rhomax}
\rho < \bar{\rho}(N,\phi) = \left(1-\frac 1 N\right) \,\phi\left(\frac  N {N-1}\right) + \frac 1 N \phi(0).
\end{equation}
This assumption is included in~\refprop{prop:rgrad}.

Another cause for degeneracy in an optimization solution is if the objective function does not depend on the decision variables, in which case the entire feasible set is optimal.
In the problem~\eqref{restrob}, this could happen if $l(\theta,\xi_n) = \ell, \forall n$, since the objective would be $\sum_n l(\theta,\xi_n) p_n = \ell \sum_n p_n = \ell$.
Define the set $\Theta_{\varnothing} \DefAs  \{\theta : l(\theta, \xi_{n_1}) = l(\theta, \xi_{n_2}),\,\,\forall\,n_1,n_2 \}$,
and for a small $\varsigma> 0$ let the set $\Theta_{\varnothing,\varsigma} \DefAs \cup_{\theta_o \in \Theta_{\varnothing} } \{\theta : |\theta - \theta_{o}| < \varsigma \}$
define the $\varsigma$-neighborhood of $\Theta_{\varnothing}$.
Then, we assume in \refprop{prop:rgrad} that $\Theta \subseteq \Theta_{\varnothing,\varsigma}^c$ for some small $\varsigma > 0$.
In other words, for each $\theta \in \Theta$, there exists two sample points $\xi_{n_1}$ and $\xi_{n_2}$ such that $l(\theta, \xi_{n_1})\neq l(\theta,\xi_{n_2}) $.

\ProofOf{\refprop{prop:rgrad}}
From the preceding discussion, our assumption that $\rho < \bar{\rho}(N,\phi)$ only admits feasible pmfs that assign non-zero mass to all support points.
For strictly convex functions $\phi(\cdot)$, this then ensures that the problem~\eqref{restrob} has a unique optimal solution $P^*$ when combined with the assumption that the objective coefficients $l(\theta,\xi_n) \neq \ell$ for all $n$ and some $\ell$.
Moreover, $D_{\phi}(P^*, U_N)=\rho$.

For part (ii), first recall from the discussion in the introduction that
\citeN{bhwmg13} describe the typical Lagrangian dual algorithm used
for the convex-concave case by
\begin{align}
R(\theta^*_{rob}) &= \min_{\theta\in\Theta} \max_{p_n\ge 0}\min_{\alpha\ge 0,\lambda} \Big\{ L_{P}(\theta) + \alpha( \rho - D_{\phi}(P, U_{N}) )
+ \lambda\Big( 1 - \sum_{n=1}^N p_n \Big) \Big\} \nonumber\\
&\hspace*{-0.3in} 
=\min_{\theta\in\Theta, \alpha\ge 0 , \lambda} \alpha \rho + \lambda + \frac{\alpha}{N} \sum_{n=1}^N\phi^*\Big(\frac{l(\theta,\xi_n)-\lambda}{\alpha}\Big)  ,
\label{dualfmln}
\end{align}
where the convex conjugate of $\phi$,
$\phi^*(s) = \max_{u\ge 0} \{su - \phi(u)\}$, is known in closed form
for various $\phi$ such as modified $\chi^2$- and $KL$-divergence.
Now
define $\cL(\theta,\alpha,\lambda,P)$ as the Lagrangian in~\eqref{dualfmln}:
\begin{equation}\label{lagr}
\cL(\theta,\alpha,\lambda,P) = L_{P}(\theta) + \alpha( \rho - D_{\phi}(P, U_{N})) + \lambda(1-\sum_n p_n).
\end{equation}
By (i), there exists a unique solution $P^*(\theta)$, and by Lagrangian duality principles ~\citep[Lemma 2.1]{shap85}, a corresponding unique pair $(\alpha^*,\lambda^*)$ exists.
Collectively call the primal and dual variables $v^*(\theta) = (\alpha^*(\theta),\lambda^*(\theta), P^*(\theta))$,
and thus $R(\theta) = \cL(\theta, v^*(\theta))$, where the first term $L_{P^*}(\theta) = \sum_n p^*_n(\theta) l(\theta,\xi_n).$  
Differentiating using the chain rule, we have
\begin{align}
\nabla_{\theta}R(\theta) &= \nabla_{\theta} L_{P^*(\theta)}(\theta) + \nabla_{\theta} v^*(\theta) \,\, \nabla_{v} \cL(\theta, v^*(\theta)) \label{firstder}\\
& = \sum_{n\in \cN}p^*_n(\theta)\nabla_{\theta} l(\theta,\xi_n), \nonumber
\end{align}  
where the second term on the right in \eqref{firstder} vanishes because $\nabla_{v} \cL(\theta, v^*(\theta)) = 0$ by the first order optimality conditions of $v^*$.
The same result is obtained in a more general setting that allows for multiple solutions to the maximization problem~\citep[Theorem~7.21, p.~352]{shap09}. 
\ProofEnd

\ProofOf{\refprop{prop:rgradlip}}
Further differentiating~\eqref{firstder} using the chain rule and again applying the first order optimality conditions for $v^*$, we obtain
\begin{equation}\label{hessdef}
\nabla_{\theta}^2 R(\theta) = \sum_{n\in \cN} p^*_n \nabla^2_{\theta} l(\theta,\xi_n) + \nabla_{\theta} v^*(\theta) \,\,[\nabla^2_{\theta v}\cL(\theta,v^*(\theta)) ]^T .
\end{equation}
Since the $P^*$ are bounded, we have our desired result from the triangle inequality if we can show that the second term has bounded components. 

Following \citep[Lemma 2.2]{shap85}, the gradient $\nabla_{\theta} v^*(\theta)$ can be expressed as
\begin{align*}
\nabla_{\theta} v^*(\theta) = -\nabla^2_{\theta v}\cL(\theta,v^*(\theta)) [\nabla^2_{vv}\cL(\theta,v^*(\theta))]^{-1}.
\end{align*}
Let $\blm = (\alpha, \lambda)$.
Further let $\phi'(s) = \D \phi(s)/\,\D s$ and $\phi''(s) = \D^2 \phi(s)/\,\D s^2$ be the first and second derivative of $\phi(s)$ w.r.t.\ $s$, respectively.
For~\eqref{dualfmln}, we then obtain the following components:
\begin{align*}
\nabla^2_{\blm\blm}\cL = 0 , & & \nabla^2_{pp}\cL = -N\alpha \mbox{Diag}(\phi''(NP)) , \\
\nabla^2_{p\blm}\cL = -\left[\begin{array}{c}\be \\  \phi'(NP)\end{array}\right]  ,
& &\nabla^2_{\theta v} \cL = \left[\begin{array}{c}
\nabla_{\theta} l_1(\theta,\xi_1) \\
\ldots \\
\nabla_{\theta} l_N(\theta,\xi_N)\\
 \bzr \\
  \bzr
  \end{array}
  \right] ,
\end{align*}
where $\phi'(NP)$ and $\phi''(NP)$ represent the vectors of first and second derivatives of $\phi$ at the components of the vector $NP$, and $\be$ represents the vector of all ones. 

We calculate the inverse of $\nabla^2_{vv} \cL$ using the Schur complement of $\nabla^2_{pp}\cL$ in $\nabla^2_{vv} \cL$, along with the components $\nabla^2_{\blm\blm}\cL$ and $\nabla^2_{p\blm}\cL$.
Note that, for a square matrix $M$ which can be partitioned into submatrices $A,B,C,D$ (as shown below) and where $A$ is invertible as $A^{-1}$, the Schur complement of $A$ is $M/A=D - CA^{-1}B$.
We then have
\begin{align*}
M &= \left[\begin{array}{cc}A & B\\C &D\end{array}\right] ,\\
M^{-1} &= \left[\begin{array}{cc}
A^{-1} + A^{-1}B(M/A)^{-1}CA^{-1} & -A^{-1}B(M/A)^{-1}\\
-(M/A)^{-1}CA^{-1} & (M/A)^{-1}
\end{array}\right].
\end{align*}
Applying this to $M=\nabla^2_{vv}\cL$ with $A=\nabla^2_{pp}\cL$, we obtain $D=\nabla^2_{\blm\blm}$ and $C=B^T= \nabla^2_{p\blm}\cL$.
Given the form of $\nabla^2_{\theta v} \cL$, we need only compute the top-left element of $M^{-1} = [\nabla^2_{vv}\cL]^{-1}$ in order to compute the second term in~\eqref{hessdef}.
The Schur complement is given by
\begin{align*}
& M/A = \frac 1 {N\alpha} 
\left[
\begin{array}{c}
\be \\  
\phi'(NP)
\end{array}
\right] 
\mbox{Diag} \left(\frac 1 {\phi''(NP)}\right)
\left[\be^T \,\,\,\,\phi'(NP)^T\right] \\
&= \frac 1 {N\alpha} 
\left[
\begin{array}{cc}
\sum_n(1/\phi''(Np_n)) & \sum_n(\phi'(Np_n)/\phi''(Np_n))\\  
\sum_n(\phi'(Np_n)/\phi''(Np_n)) & \sum_n (\phi'(Np_n))^2/\phi''(Np_n)
\end{array}
\right] .
\end{align*}
Note that $M/A$ is a $2\times 2$ matrix,
and the inverse of a $2\times 2$ matrix $[a\,,\, b ; c \,,\, d]$ is $[d \,,\, -b ; -c \,,\, a]/ (ad-bc)$.
Since the $\phi$ are strictly convex, $\phi''(Np_n) \ge  \delta > 0$ for some $\delta$ and any $p_n$.
Then the term $(ad-bc)$ is the variance of a random variable taking values $\phi'(Np_n)$ with probability $1/\phi''(Np_n)$, and thus it is strictly positive, again because of the strict convexity of $\phi$.
Finally, the terms $a,b,c,d$ are all finite because their denominators are strictly away from zero by $\delta$. 

The term $A^{-1} + A^{-1}B(M/A)^{-1}CA^{-1}$ can similarly be found to have elements that are all finite (the optimal $\alpha^*<\infty$ as seen in the proof of~\refprop{prop:im_cmp_bnd} below),
and hence the second term in~\eqref{hessdef} also has finite elements, rendering the desired result.
\ProofEnd

\subsection{{Solving for $P^*(\theta)$ in $D_{\phi}$-Constrained Inner-Maximization}}\label{apdx:getPstar}

Let $\phi'(s) = \D \phi(s)/\,\D s$ be the derivative of $\phi(s)$ w.r.t.\ $s$, with $(\phi')^{-1}$ denoting its inverse.
By assumption, $\phi$ is strictly convex, and thus $\phi'(s)$ is strictly increasing in $s$, which provides us with the existence of its inverse.
The derivative $\phi'(s)$ plays a key role in the proof of~\refprop{prop:im_cmp_bnd} below.
See~\reffig{fig:phi_conv} for an illustration of plots of
$\phi$ as a function of $s$,
$\phi'$ as a function of $s$,
its inverse $(\phi')^{-1}(y)$ as a function of $y$,
and finally $(\phi')^{-1}((z_i-\lambda)/\alpha)$ as a function of $\lambda$.
The plots illustrate both cases where $\phi'(s)\tndni$  (e.g., KL-divergence $\phi(s) = s\log s -s +1$) on the top row
and $\lim_{s\tndo+} \phi'(s) > -\infty$ (e.g., modified $\chi^2$-divergence $\phi(s) = (s-1)^2$) on the bottom row.

We use the following general procedure to solve the Lagrangian formulations~\eqref{roblag} 
for a given iteration $t$.
This basic approach has been pursued, either explicitly or in a similar spirit, in previous work such as~\citeN{bhwmg13,gl18,nd16,nd17}.\\

\textbf{Procedure 1.}
\begin{enumerate} 
	\item 
	{\em Case:} 
	$\alpha^*=0$ along with constraint $D_{\phi}({P}^*_{t},P_b) \le \rho_t$.
	\begin{enumerate}
		\item Let $\cM'_t = \{m\in\cM_t : z_m = \max_{u\in\cM_t} z_u \}$ and
		$M'_t = |\cM'_t|$. Set $\alpha^*=0$ in \refeq{roblag}, and then an optimal solution is ${P}^*$ where ${p}^*_m = \frac{1}{M'_t},\;\forall m\in\cM'_t$,
		and ${p}^*_m =0,\; \forall m\notin\cM'_t$; see~\refprop{prop:im_cmp_bnd}.
		\item If $D_{\phi}({P}^*, P_b) \le \rho_M$, then
		{\bf stop} and return ${P}^*$.
	\end{enumerate}
	\item {\em Case:} constraint $D_{\phi}({P}^*_{t},P_b) = \rho_t$ with
	$\alpha^*\ge 0$.
	\begin{enumerate}
		\item Keeping $\lambda, \alpha$ fixed, solve for the optimal
		${P}^*_t$ (as a function of $\lambda,\alpha$) that maximizes
		$\cL(\alpha, \lambda, {P})$, applying the constraint
		${p}_m\ge 0$.
		\item Keeping $\alpha$ fixed, solve for the optimal
		$\lambda^*$ using the first order optimality condition on
		$\cL(\alpha,\lambda,{P}^*_t)$.  Note that this is equivalent to
		satisfying the equation $\sum_{m\in\cM_t}{p}^*_m = 1$.
\refprop{prop:im_cmp_bnd} shows that this step is at worst a bisection search in one dimension, but in some cases (e.g., KL-divergence) a solution $\lambda^*$ is available in closed form.
		The proof of~\refprop{prop:im_cmp_bnd} also provides finite bounds $[\underline{\lambda}, \bar{\lambda}]$ on the range over which we need to search for $\lambda^*$.
		\item Apply the first order optimality condition to the
		one-dimensional function $\cL(\alpha, \lambda^*(\alpha), {P}_t^*)$ to
		obtain the optimal $\alpha^* \ge 0$. This is equivalent to
		requiring that $\alpha^*$ satisfies the equation
		$\sum_{m\in\cM}\phi({p}^*_{t,m}) = \rho_t$.
\refprop{prop:im_cmp_bnd} shows that this is at worst a one-dimensional bisection search which embeds the previous step in each function call of the search. 
		\item Define the index set $\cN=\{m \in \cM_t\,\,\mid\,\lambda^* \le z_m - \alpha^* \phi'(0)\}$, with $\cN=\emptyset$ if $\phi'(s)\tndni$ as $s\tndo+$. Set
		\begin{equation}\label{p_star}
		p^*_{t,m} = \left\{ 
		\begin{array}{ll}
		\frac{1}{M_t} (\phi')^{-1}\left(\frac{z_m - \lambda^*}{\alpha^*}\right), 
		& \,\,\, m\in\cN \\
		0 & \,\,\,m\notin\cN
		\end{array} 
		\right. .
		\end{equation}
(Note that \refprop{prop:im_cmp_bnd} explains the expression in~\eqref{p_star}.)
		\textbf{Return} ${P}_t^*$.
	\end{enumerate}
\end{enumerate}


\begin{figure*}[htbp]
	\begin{minipage}{.24\textwidth}
		\centering
		\includegraphics[width=0.9\columnwidth]{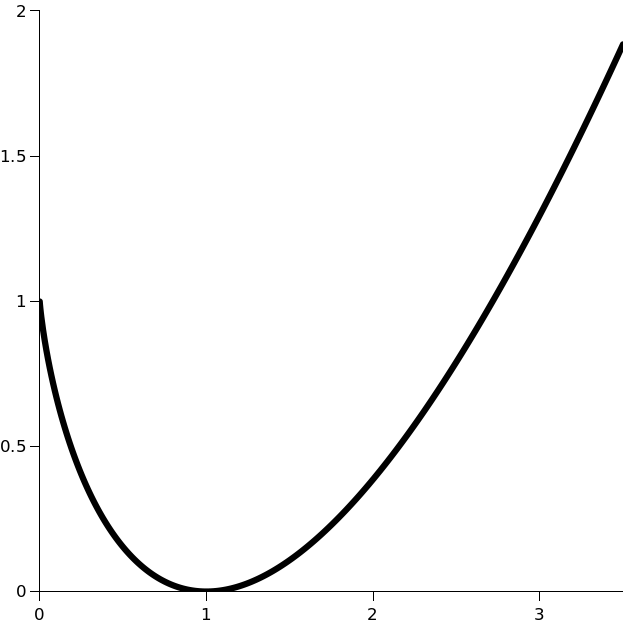}
	\end{minipage}%
	\begin{minipage}{.24\textwidth}
		\centering
		\includegraphics[width=0.9\columnwidth]{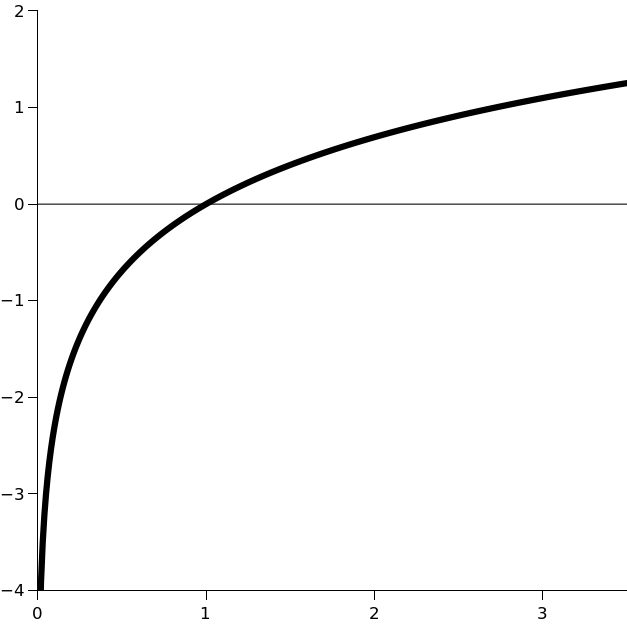}
	\end{minipage}%
	\begin{minipage}{.24\textwidth}
		\centering
		\includegraphics[width=0.9\columnwidth]{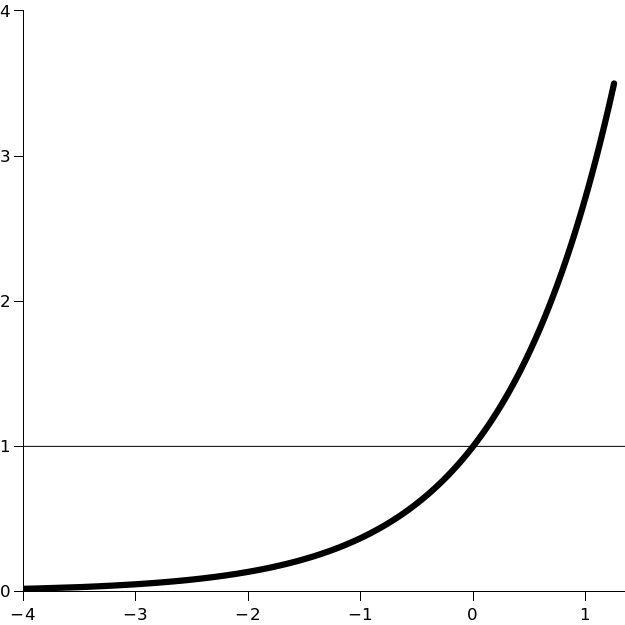}
	\end{minipage}%
	\begin{minipage}{.24\textwidth}
		\centering
		\includegraphics[width=0.9\columnwidth]{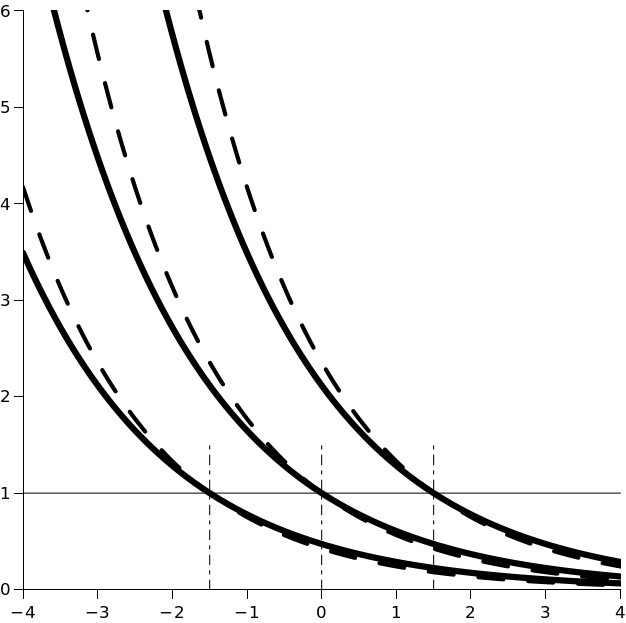}
	\end{minipage}%
	\vskip .1in
	\begin{minipage}{.24\textwidth}
		\centering
		\includegraphics[width=0.9\columnwidth]{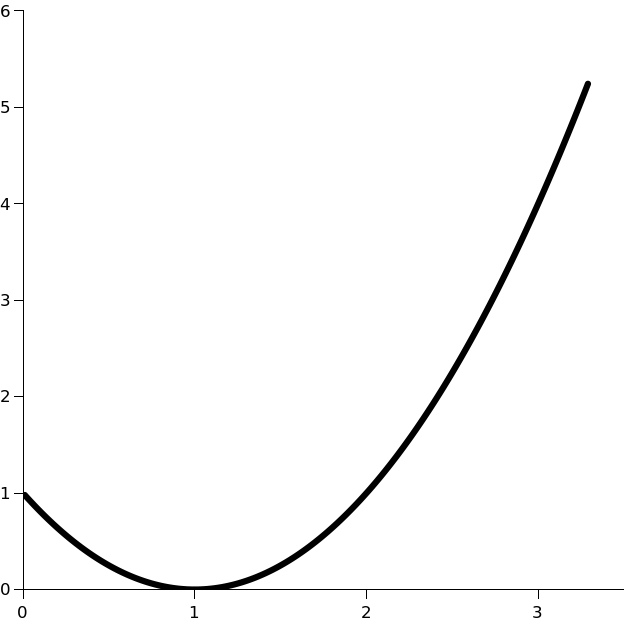}
	\end{minipage}
	\begin{minipage}{.24\textwidth}
		\centering
		\includegraphics[width=0.9\columnwidth]{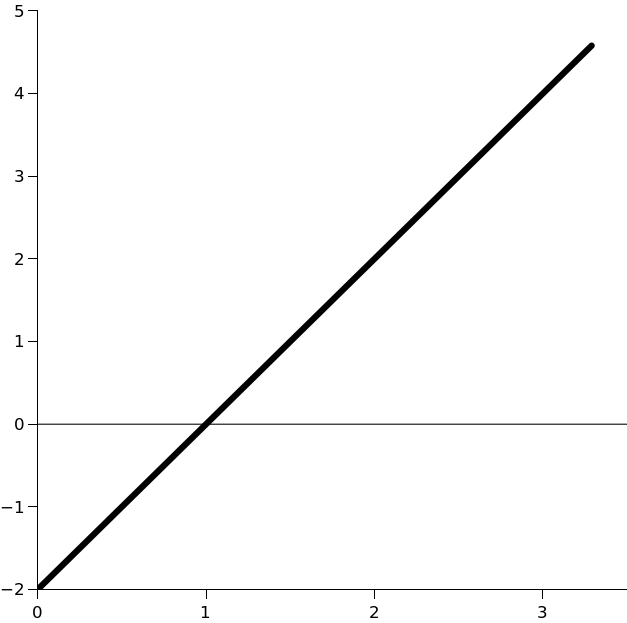}
	\end{minipage}
	\begin{minipage}{.24\textwidth}
		\centering
		\includegraphics[width=0.9\columnwidth]{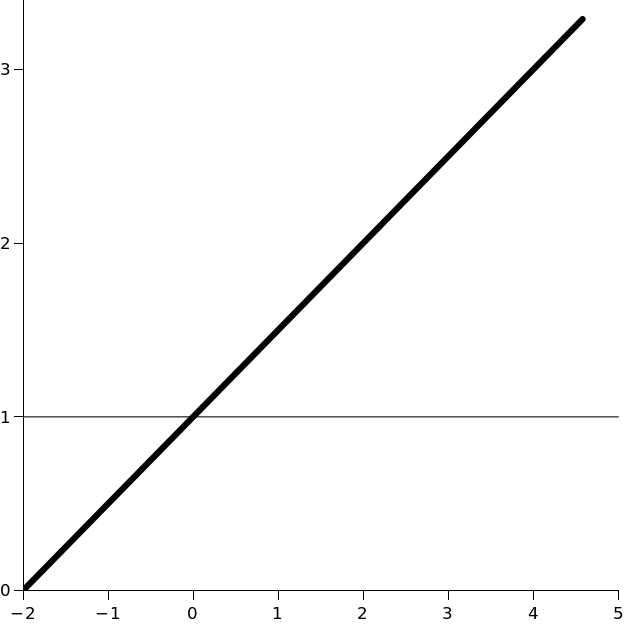}
	\end{minipage}
	\begin{minipage}{.24\textwidth}
		\centering
		\includegraphics[width=0.9\columnwidth]{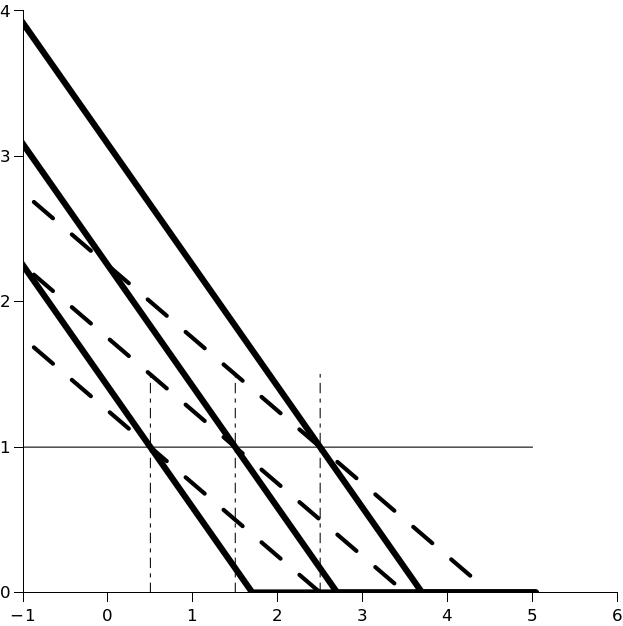}
	\end{minipage}
	\caption{The plots on the top row 
		are for the Kuhlback-Leibler divergence,
		and those on the bottom 
		are for the modified $\chi^2$-divergence.
		Specifically, the top row from left-to-right considers:
		$\phi(s) = s\log s -s +1,\,\, s\ge 0$;
		$\phi'(s) = \log s,\,\, s > 0$;
		$(\phi')^{-1}(y) = e^y,\,\,y\in\real$;
		$p^*_{z,\alpha}(\lambda) = (\phi')^{-1}( (z-\lambda) / \alpha ),\,\,\lambda\in\real$.
		The bottom row from left-to-right considers:
		$\phi(s) = (s-1)^2,\,\,s\ge 0$;
		$\phi'(s) = 2(s-1),\,\, s\ge 0$;
		$(\phi')^{-1}(y) = 1+ y/2,\,\,y\in\real$;
		$p^*_{z,\alpha}(\lambda) = (\phi')^{-1}( (z-\lambda) / \alpha),\,\,\lambda\in\real$.
		Note: On the top, $\phi'(s)\rightarrow-\infty$, so the inverse is always positive;
		for modified $\chi^2$-divergence on the bottom, it is positive only when $y\ge -2$.
		The last column plots $p^*_m$ from~\eqref{p_star} for three $z_m$ and two $\alpha$.
	}
	\label{fig:phi_conv}
\end{figure*}

\ProofOf{\refprop{prop:im_cmp_bnd}}
We eliminate the subscript $t$ in this proof for clarity of exposition; thus, the support is indexed over $m=1,\ldots,M$.

To start, order all the $z_{m}$ into the increasing sequence $z_{(1)}\le z_{(2)} \le \ldots\le z_{(M)}$, where the subscript notation $(i)$ denotes the index of the $i$-th smallest $z_m$ value.
The tightest bound on the cost of sorting the vector $(z_m)$ in increasing order is of order $O(M\log M)$.

We first handle the case when the $\phi$-divergence constraint is not tight and $\alpha^*=0$.
Substituting this in~\refeq{roblag} shows that any optimal solution $\hat{P}^*$ places mass only within the set $\cM^\prime$ as defined in Step 1(a).
Consider any such ${P}^*$, and let $U_{M}$ be as defined in the statement of~\reflem{lem:unif_small_phi}.
Then the lemma provides that, among all optimal solutions, $U_{M'}$ obtains the smallest divergence,
and thus it is the best optimal candidate to meet the divergence constraint with slack.
This is why the solution procedure stops in Step 1(b). 
The computational complexity of Step 1 is mainly due to determining the set $M'$, which can be part of the sorting operation that determines the sequence $z_{(m)}$ above.
 
Note that this case is precluded by the assumptions of~\refprop{prop:rgrad}. With the assumption that $\rho < \bar{\rho}(N,\phi)$, we have that the feasible region only allows for pmfs that have non-zero mass over the full support $M$. This in turn implies that for this case to hold, we need $l(\theta,\xi_n)=\ell$ for all $n$, which is also ruled out by the assumptions of~\refprop{prop:rgrad}. 

For the case when the $\phi$-divergence constraint is tight, we proceed according to the corresponding three steps in \textbf{Procedure 1}. 
\paragraph{Step 2(a).} Identify the set of indices $\cN$ as defined above. Setting to zero the gradient of $\cL(\alpha,\lambda,P)$ with respect to $P$, we obtain the expression in~\eqref{p_star} for $P^*(\lambda, \alpha)$.
For the case where $\phi'(s)\tndni$, all $p^*_{m} > 0$ since $\cN=\emptyset$.
In the case where $\phi'(s)\rightarrow K > -\infty$ for some constant $K$, the probability values need to observe the check on $(\lambda, \alpha)$ as given in~\eqref{p_star} in order to satisfy the non-negativity constraint on ${P}^*$. 
The set $\cN = \cN(\lambda,\alpha)$ can be equivalently represented as $\cN = \{ (m) \,\,\mid\,\,(m) \ge (m_g)\}$
where $(m_g) = \inf_m \{ (m)\,\,\mid\,\, z_{(m)} \ge \lambda + \alpha \phi'(0) \} $, i.e., the smallest ordered value $z_{(m)}$ satisfying the defining condition of $\cN$.
From~\eqref{p_star}, we observe that $p^*_{(m)}$ are strictly increasing in $m$ for any fixed $(\lambda,\alpha)$.
Hence, the optimal probability allocation to support points increases in accordance with their $z_m$ values. 

\paragraph{Step 2(b).} Define
\[
h_{\alpha}(\lambda) \,\,\DefAs\,\, \sum \hat{p}^*_m \,\,=\,\, \frac 1 {M} \sum_{(m)\ge (m_g)} (\phi')^{-1}\left(\frac{z_{(m)} - \lambda}{\alpha}\right) .
\]
We seek the $\lambda$ that attains $h_{\alpha}(\lambda) =1$, i.e., $\sum \hat{p}^*_{m} =1$. The
above
expression
for
$h_{\alpha}(\lambda)$
is a decreasing function of $\lambda$ for fixed $\alpha$, given the strict convexity of $\phi$.
When $\phi'(0)> - \infty$, at $\bar{\lambda} = \bar{\lambda}(\alpha) = z_{(M)} - \alpha \phi'(0)$, the summation $h_{\alpha}(\bar{\lambda}) = 0$.
For the $\phi'(0)\rightarrow-\infty$ case, it is sufficient to consider a point $\bar{\lambda}(\alpha) = z_{(M)} - \alpha (z_{(1)} - z_{(M)})$ to obtain that $h_{\alpha}(\bar{\lambda}) < 1$.
On the other hand, at $\underline{\lambda}= z_{(1)}$, we have (note the properties of $\phi'$ in \reffig{fig:phi_conv}) that
\begin{align*}
h_{\alpha}(z_{(1)}) &= \frac 1 {M} \bigg( \underbrace{(\phi')^{-1}\Big(\frac{z_{(1)} - z_{(1)}}{\alpha}\Big)}_{=1}
+ \sum_{m>1} \underbrace{(\phi')^{-1}\Big(\frac{z_{(m)} - z_{(1)}}{\alpha}\Big)}_{>1}\bigg) \quad>\quad 1. 
\end{align*}

Hence, we only need to perform a bisection search on $[\underline{\lambda}, \bar{\lambda}]$,
which can be performed with a computational effort of at most $O(\log 1/\epsilon)$ to get to within $\epsilon$ precision.
The size $|\bar{\lambda} - \underline{\lambda}|$ of the search interval appears in the order constant, but does not change with $M$.

\paragraph{Step 2(c).} To obtain $\alpha^*$, substitute the $P^*$ from~\eqref{p_star} into
the divergence constraint satisfied as an equality:
\[
 D_{\phi}(\alpha) = \frac 1 {M} \sum_m \phi(M p^*_m (\alpha)) = \rho .
\]
We show below that the function $D_{\phi}(\alpha)$ is decreasing in $\alpha$, and hence a bisection search leads us to the optimal $\alpha^*$.
The total computational effort then in estimating the correct $\alpha^*$ involves $\log (\frac{1}{\epsilon})$ function calls to the bisection search to find the $\lambda(\alpha)$ for the current iterate of $\alpha$.
As noted before, each of these function calls takes at most $\log (\frac{1}{\epsilon})$.
This, in addition to the time taken to sort the $z_m$ once in each run of the complete algorithm, provides us with the bound on the computational effort.

Consider any pair of $\alpha_1<\alpha_2$. Let $\lambda^*(\alpha_i) = \lambda^*_i$ be the optimal value that attains $h_{\alpha_i}(\lambda^*_i) = 1,\,\,i=1,2$.
The summation $h_{\alpha}(\lambda)$ is decreasing in $\alpha$ for a fixed $\lambda$, and thus $h_{\alpha_2}(\lambda^*_1) < 1$.
Since $h_{\alpha}(\lambda)$ is a decreasing function of $\lambda$ for a fixed $\alpha$, we have that $\lambda^*_2<\lambda^*_1$.

Now, let $\underline{\lambda}_2$ be the value that satisfies the equality $(z_{(M)} - \underline{\lambda}_2) / \alpha_2 = (z_{(M)} - \lambda^*_1) / \alpha_1$,
so that at $\underline{\lambda}_2$ the support point corresponding to the largest value $z_{(M)}$ has the same probability allocation~\eqref{p_star} under
$\alpha_2$ as under $\alpha_1$.
For any $m<M$, we then have
\begin{align*}
\frac {z_{(m)} - \underline{\lambda}_2} {\alpha_2} &\;\;=\;\; \frac {z_{(m)} - z_{(M)}}{\alpha_2} + \frac {z_{(M)} - \underline{\lambda}_2} {\alpha_2} \;\;
 =\;\; \underbrace{(z_{(M)}-z_{(m)}) \left(\frac 1 {\alpha_1} - \frac 1 {\alpha_2}\right)}_{>0} + \frac{z_{(m)}-\lambda^*_1}{\alpha_1}.
\end{align*}
Therefore, at $\underline{\lambda}_2$, we have that $h_{\alpha_2}(\underline{\lambda}_2) > 1$ and that $\lambda^*_2\in[\underline{\lambda}_2,\lambda^*_1]$.
As a consequence, the mass allocated to $p^*_{(M)}$ also decreases.
Let $\Delta\DefAs p^*_{(M)}(\alpha_1) - p^*_{(M)}(\alpha_2)$, and then from the preceding discussion $\Delta > 0$.
Moreover, we have $p^*_{(M)} (\alpha_2) > 1/M$, else given the strict ordering of \eqref{p_star} over $\{(m)\}$, the total probability assigned over all support points will sum up to less than $1$.
Hence, $M\, p^*_{(M)}(\alpha_2) > 1$, to the right of the minima at $s=1$ of $\phi(s)$. 

The change in $D_{\phi}$ is bounded as follows:
\begin{align*}
M \left(D_{\phi}(\alpha_2) - D_{\phi}(\alpha_1) \right)
&=\left(\phi(p^*_{(M)}(\alpha_2)) - \phi(p^*_{(M)}(\alpha_1))\right)
+ \sum_{m<M} \left(\phi(p^*_{(m)}(\alpha_2)) - \phi(p^*_{(m)}(\alpha_1))\right)\\
&= \phi'(\xi_M) (-\Delta) + \sum_{m<M} \left(\phi(p^*_{(m)}(\alpha_2)) - \phi(p^*_{(m)}(\alpha_1))\right)\\
&\le -\phi'(\xi_{M}) \Delta + \left(\phi(\,\,p^*_{(M)}(\alpha_2)\,\,) - \phi(\,\,p^*_{(M)}(\alpha_2)-\Delta\,\,)\right) \\
&= - \phi'(\xi_{M})\, \Delta + \phi'(\underline{\xi}_{M}) \, \Delta\quad <\quad 0.
\end{align*}
Here, the second equality applies the mean value theorem for some $\xi_M\in [ p^*_{(M)} (\alpha_2), p^*_{(M)} (\alpha_1)]$.
The first inequality is due to the highest increase in $\phi$ occurring if all the reduction in probability $\Delta$ is picked up by the next support point $p^*_{(M-1)}$.
Since the mass allocations are ordered, the best case is that $p^*_{(M-1)}(\alpha_1) = p^*_{(M)}(\alpha_2)-\Delta$ increases to $p^*_{(M-1)}(\alpha_2) = p^*_{(M)}(\alpha_2)$.
The third equality again applies the mean value theorem for some $\underline{\xi}_{M} \in  [p^*_{(M)}(\alpha_2)-\Delta, p^*_{(M)}(\alpha_2)]$, and the last inequality is due to the strict convexity of $\phi$.

Therefore, $D_{\phi}(\alpha)$ is a decreasing function of $\alpha$.
To obtain a finite range $[\underline{\alpha}, \bar{\alpha}]$ that contains $\alpha^*$, first consider the $\phi'(s)>-\infty$ case.
There exists an $\alpha_{\tau}$ such that $\lambda^*_{\tau} = \lambda^* (\alpha_{\tau}) = z_{(M)} -\tau$, for a small $\tau>0$,
and the corresponding $P^*_{\tau}$ places mass $1$ on $z_{(M)}$ and zero elsewhere;
this leads to $D_{\phi}(\alpha_{\tau}) = D_{\phi}(U_1, U_{M})$, which is the highest distance possible as per the discussion following~\reflem{lem:unif_small_phi}.
On the other hand, as $\alpha\tndi$, each $p^*_m\rightarrow 1/M$ as observed from~\eqref{p_star} and~\reffig{fig:phi_conv}, and thus $D_{\phi}(\alpha)\tndo$ as $\alpha\tndi$.
Hence, there exists a range $[\underline{\alpha}, \bar{\alpha}]$ that contains $\alpha^*$. 

The upper limit $\bar{\alpha}$ also applies when $\phi'(s)\rightarrow -\infty$ , so only the lower limit $\underline{\alpha}$ needs be modified for this case.
There exists a small $\delta_{\tau}$ such that $P^*_{\tau}$ places mass $1-M\delta_{\tau}$ on $z_{(M)}$ and at most $\delta_{\tau}$ mass on $z_{(m)},\,\,m<M$.
As $\tau\tndo$, $\delta_{\tau}\tndo$ and $D_{\phi}(\alpha_\tau)\rightarrow D_{\phi}(U_1,U_{M_1})$. 

Note that when the conditions for Case $1$ and Case $2$ both hold,
then the steps of Case $1$ and Case $2$ will both result in the same final outcome.
\ProofEnd



%% file: apdx_bias.tex
\subsection{Small-Sample Approximation of $\nabla_{\theta} R(\theta)$}\label{apdx:bias}

The setup of the robust loss function enables us to extensively exploit the mathematical properties of statistically sampling a finite set without replacement, and we therefore provide a brief summary here.
Let $\{x_1,\ldots,x_N\}$ be a set of $N$ one-dimensional values with mean $\mu = \frac{1}{N} \sum_n x_n$ and variance $\sigma^2 = \frac{1}{N-1} \sum_n(x_n-\mu)^2$.
Suppose we sample $M<N$ of these points uniformly without replacement to construct the set $\cM= \{X_1,\ldots, X_M \} $.
The probability that any particular set of $M$ subsamples was chosen is given by ${(N-M)!\choose N!}$.
Denote by $\E_{\cM}$ the expectation under this probability measure, and
let $\bar{X} = \frac{1}{M} \sum_{m=1}^M X_m$ and $\bar{S}^2 = \frac{1}{M-1} \sum_{m=1}^M (X_m - \bar{X})^2$ represent
the sample mean and sample variance, respectively.
We then know that the expectation of the sample mean $\E_{\cM}[\bar{X}] = \mu$ and of the sample variance $\E_{\cM}[\bar{S}^2] = \sigma^2$ are both unbiased;
refer to~\cite{wilks}.
On the other hand, the variance of the sample mean
\[
\E_{\cM}[(\bar{X}-\mu)^2] = \left(\frac{1}{M} - \frac{1}{N}\right)
\sigma^2
\]
reduces to zero as $M\rightarrow N$. The term on the right above is fundamental to our results.

With this overview, we now begin our proof by addressing the feasibility of the restriction
$\tilde{P}^{*}$ of the (unique, by assumption) optimal solution $P^*$ of the
full-data problem onto the (randomly sampled) subset $\cM_t$, where
$$ \tilde{p}^*_m = \frac {p^*_m}{\sum_{j\in\cM_t}p^*_j},\,\,\,\forall m\in\cM_t.$$
Denote by $\cP_{t}$ the feasibility set of \eqref{restrob} for the sampled $\cM_t$,
and recall from Section~\ref{ssec:ssapx} the notational simplification that $\E_{\cM_t} = \E_t$ and $\Prb_{\cM_t}=\Prb_{t}$.
\begin{lemma} \label{lem:phifeas} Suppose the $\phi$-divergence
  function has strictly convex level sets and $\rho < \bar{\rho}(N,\phi)$, where
  $\bar{\rho}$ is given by~\eqref{rhomax}. Let the $D_{\phi}$-constraint
  target $\rho_t$ of the restricted problem~\eqref{restrob} be set as stated
  in~\refthm{thm:bias}.  Then, for $M \ge M'_t$, we have
  \begin{equation}\label{eq:phifeas}\Prb_{t} (\tilde{P}^* \in \cP_{t}) \ge
  \max\left\{0, 1- \tau_1\right\}\cdot
  \max\left\{0, 1- \tau_2\right\},
  \end{equation}
where
\begin{align}
 \tau_1 &\DefAs \eta_t^{2\delta/(1-\delta)} \sigma^2(P^*),
\nonumber \\
  \tau_2 
&
\DefAs \eta_t^{2\delta / (1-\delta)}\frac{\sigma^2(\phi)} {{c_3^2}} , \nonumber \\
  \sigma^2 (P^*) &\DefAs \frac{1}{N-1} \sum_{n=1}^N (Np^*_n-1)^2  ,
\nonumber \\
  \sigma^2(\phi) 
&
\DefAs \frac 1 {N-1} \sum_{n=1}^N (\phi(Np^*_n)-\rho)^2 , \nonumber \\
  M' &\DefAs \min \left\{\,M\,|\, \eta_t \le \max\left\{ 1- \frac 1 {k_0} ,\,\, \zeta_0 ,\,\, \frac 1 {k_0\kappa_1}\right\}\right\} , \label{eq:Mprime} \\
c_3 & \DefAs \frac {c-k_0(\kappa_1\rho+\kappa_2)} 2 \nonumber
\end{align}
and the constants $c$ and $k_0$ are chosen to yield $ c_3 > 0$.
\end{lemma}
Note here that $\sigma^2(\phi)$ calculates the variance in the vector $\phi(Np^*_n)$ for \emph{any} $\phi$, though the formula makes it resemble the modified $\chi^2$-divergence $\phi(s)= (s-1)^2$.
Recall that the constants $\zeta_0$, $\kappa_1$ and $\kappa_2$ arise from the local continuity of Assumption~\ref{asm:phicont}, and that $\eta_t = c (\frac 1 M - \frac 1 N)^{(1-\delta)/2}$ from the statement of~\refthm{thm:bias}.
Then $M'$ is defined in~(\ref{eq:Mprime}) as the \emph{smallest} support size $M$ that leads to a prescribed positive value for the difference $(\frac 1 M - \frac 1 N)$.
Hence, as $N\tndi$, $M' = o(N)$ in order to match this positive difference.

\ProofOf{\reflem{lem:phifeas}} In the notation of sampling without-replacement introduced above,
define a set of scalar values $x_n (P^*)= Np^*_n,\,\,\forall n=1,\ldots,N$.
We use $\mu (P^*) = \frac{1}{N}\sum_n Np^*_n  = 1$ and
the variance $\sigma^2(P^*)$ as in the statement of the lemma.  
By Chebychev's inequality, the sample-average $\bar{X}$ of an
$M$-subsample chosen uniformly without replacement from this set satisfies
\begin{align*}
  \Prb_{t}\left(\left| \bar{X}(P^*) - 1\right| > \eta_t\right)
  &\le \frac{1}{\eta_t^2} \E_{t} [ (\bar{X}(P^*) - \mu)^2]
\\ 
  &
\le \left(\frac{1}{M_t}- \frac{1}{N}\right)^{\delta} \sigma^2(P^*). 
\end{align*}
Hence,  $|\bar{X} (P^*) - 1| \le \eta_t$ with probability at least $(1-\tau_1)$.
One implication of this is that $\sum_{\cM_t}p^*_j \ge (M_t/N) (1-\eta_t)$ with probability at least $(1-\tau_1)$,
and thus the restriction $\tilde{P}^*$ of $P^*$ to $\cM_t$ is a pmf (i.e.,
its denominator is greater than zero) with
probability converging to one as $M\rightarrow N$.


We now check if the restriction $\tilde{P}^*$ is feasible for the
problem~\eqref{restrob}, namely whether
$\Prb_{t} ( D_{\phi}(\tilde{P}^*, U_{M_t})> \rho_t)$ is small.
  Note that $\eta_t\searrow 0$ as $M_t\rightarrow N$; and rearrange
  $|\bar{X}-1| \le \eta_t$ to yield
  $\bar{X}^{-1} = (\frac{1}{M_t}\sum_{j\in\cM_t}Np^*_j)^{-1}\le 1 + k_0
  \eta_t$ for all $M_t\ge M'_t$.
   Then for all $M_t \ge M'_t$, we obtain on the event
  $\cG \, \DefAs \,\{|\bar{X}(P^*)-1| \le \eta_t\}$ that 
\begin{align}
D_{\phi}(\tilde{P}^*, U_{M_t}) &= \frac{1}{M_t}\sum_m\phi\left(M_t\frac{p^*_m}{\sum_jp^*_j}\right) 
\nonumber\\
  & \hspace*{-1.0in} 
= \frac{1}{M_t} \sum_m \phi\left( Np^*_m\frac{1}{\frac{1}{M_t}\sum_{j\in\cM_t}Np^*_j} \right)\nonumber\\
  & \hspace*{-1.0in} 
 \le \frac{1}{M_t}\left( \sum_m \phi(Np^*_m)(1+\kappa_1k_0\eta_t) \right) +  \kappa_2k_0\eta_t, \label{ccc}
\end{align}
where the last inequality follows
from~\refasm{asm:phicont},
applied in the form $\phi(s(1+\zeta)) \le \phi(s) (1+\kappa_1\zeta) + \kappa_2\zeta$.  

Let $\{x_n(\phi) = \phi(Np^*_n)\}_{n=1}^N$ be a vector of values
associated with the $N$ support points. Then, for the (random) set of
indices $\cM_t$ chosen uniformly without replacement,
$\E_{t} \bar{X}(\phi) = \E_{t}[ \frac{1}{M_t} \sum_{m\in\cM_t}
\phi(Np^*_m) ]= \mu(\phi) = \frac{1}{N}\sum_n \phi(Np^*_n) =
D_{\phi}(P^*, U_N) = \rho$, again exploiting the fact that the
$D_{\phi}$-constraint is tight at the optimal solution $P^*$ for the
full-support optimization problem. Note that the corresponding
population variance is given by $\sigma^2(\phi)$ as defined in the
statement of the lemma.

Now define the desired event as $\cD \,\DefAs\, \{D_{\phi}(\tilde{P}^*, U_{M_t}) \le (\rho +
   c\eta_t)\}$. Then, 
inequality~\eqref{ccc} renders the following bound on its complement   
\begin{align*}
&
\Prb_{t}\left( \cD^c | \cG \right)
\\
&= \Prb_{t} \left( D_{\phi}(\tilde{P}^*, U_{M_t}) > (\rho + c\eta_t) \,\,\left|\,\, \cG \right.\right)\\
&\le \Prb_{t} \bigg( \frac{1+\kappa_1k_0\eta_t}{M_t} \sum_{m\in\cM_t} \phi(Np^*_m) + \kappa_2 k_0\eta_t > (\rho + c\eta_t) \bigg) \\
&\le \Prb_{t} \bigg(  \Big| \frac 1 {M_t} \sum_{m\in\cM_t} \phi(Np^*_m) - \rho \Big| > \frac{(c-\kappa_2k_0-\rho\kappa_1k_0)}{1+\kappa_1k_0\eta_t}\eta_t \bigg) \\
&\le \Prb_{t} \bigg(  \Big| \frac 1 {M_t} \sum_{m\in\cM_t} \phi(Np^*_m) - \rho \Big| > c_3\eta_t \bigg)
\\
&
\le \eta_t^{2\delta/(1-\delta)}\frac {\sigma^2(\phi)}{{c_3^2}} ,
\end{align*}
where $c_3 = (c-\kappa_1k_0\rho-\kappa_2k_0)/2 >0$ by the assumptions on $c$
and the minimum $M'_t$, and where Chebychev's inequality is used in the last step.
This yields the desired high probability guarantee from the
elementary probability identity that $\Prb_{t}(\cD) \ge \Prb_{t}(\cD | \cG)
\Prb_{t} (\cG)$. 
\ProofEnd

\reflem{lem:phifeas} shows that the specific choice of $\rho_t$ leads to
the restriction $\tilde{P}^*_{t}$ of the unique optimal $P^*$ to be feasible for
\eqref{restrob} with probability converging to $1$ as $M\rightarrow N$.
We next establish that the bias in the estimation of the optimal objective is
$O_p(\eta_t)$.
\begin{lemma} \label{lem:objbias}
  Under the assumptions of~\reflem{lem:phifeas}, there exists a $c_1>0$ such that 
  \[
    \Prb_{t}\left( {\eta_t}^{-1}|\hat{R}_{t}(\theta) - R(\theta)| \le c_1\right) \ge \bar{\tau}_t
  \]
  where 
  \begin{align*}
  \bar{\tau}_t &\DefAs \max\{0,1- \tau_3\} \cdot \hat{\tau} \\
  \tau_3 & \DefAs \eta_t^{2\delta / (1-\delta)}\frac{\sigma^2(R)} {{c_1^2/4}}\\
  \hat{\tau} &\DefAs \max\left\{0, 1- \tau_1\right\}\cdot
  \max\left\{0, 1- \tau_2\right\} \\
	\sigma^2(R) &\DefAs \frac 1 {N-1} \sum_{n=1}^N ((z_nNp^*_n)-\mu(R))^2 \\
  \mu(R) &\DefAs \frac 1 N \sum_n z_nNp^*_n.
  \end{align*}
\end{lemma}
Note the definition of $\bar{\tau}_t$ in Lemma~\ref{lem:objbias}, which appears in the statement of~\refthm{thm:bias}.
Further note that $C$ in \refthm{thm:bias} corresponds to $c_1$ in Lemma~\ref{lem:objbias}.
The term $\hat{\tau}$ is the probability on the right hand side of~\eqref{eq:phifeas}.

\ProofOf{\reflem{lem:objbias}} We split the optimality gap as
$|\hat{R}_{t}(\theta) - R(\theta)| \le | z^T{P}^*_{t} - z^T\tilde{P}^*| + |z^T\tilde{P}^*_t - z^TP^*|,$
where $P^*$ is the (unique) solution to the full-support problem~\eqref{absfmln},
$\tilde{P}^*_t$ its restriction to a sampled subset $\cM$ of size $M$, and
$P^*_{t}$ the (unique) solution to the optimization problem~\eqref{restrob} on the
subsampled support $\cM_t$.
We then rewrite the required probability and analyze each summand as follows
\begin{align}
\Prb_{t} \left(  |\hat{R}_{t}(\theta) - R(\theta)| > c_1 \eta_t\right) \;\;
& \le \;\; \Prb_{t} \left( | z^T{P}^*_{t} - z^T\tilde{P}^*| > \frac {c_1} 2 \eta_t\right) 
+ \Prb_{t} \left( |z^T\tilde{P}^*_t - z^TP^*| > \frac{c_1} 2 \eta_t\right).\label{tspl}
\end{align}

For the first term, note that $z^T\tilde{P}^* \le z^T{P}^*_{t}$ since $\tilde{P}^*$ is a
feasible solution to the restricted problem~\eqref{restrob} with probability $\hat{\tau}$
as shown in~\reflem{lem:phifeas}, while ${P}^*_{t}$ is its optimal solution.
On the other hand, ${P}^*_{t}$ satisfies the $D_{\phi}$-divergence constraint at $\rho+\eta_t$,
using the same arguments as for $P^*$ at $\rho$.
Following along the same lines that lead up to the expression~\eqref{firstder} of $\nabla_{\theta} R(\theta)$, we can similarly derive that $\D R / \D \rho = \D \cL / \D \rho = \alpha^*$. As noted in the proof of~\refprop{prop:im_cmp_bnd}, $\alpha^* > 0$. Thus, for sufficiently large $M_t$ such that $\eta_t$ is small,
we obtain from a first-order Taylor expansion that $z^T{P}^*_{t} \le z^TP^* + c_2\eta_t$ for some $c_2\ge 1$.

Hence, there exists a $c_1> 2c_2$ such that
$$\Prb_{t} \left( |z^T{P}^*_{t} ]- z^T\tilde{P}^*| > \frac {c_1} 2
  \eta_t \right) = 0.
$$
We rewrite the second term on the right hand side of~\eqref{tspl} as
\begin{align}
  z^T\tilde{P}^*_{t} - z^T{P}^*
  &= \sum_{m\in\cM_t}z_m\frac{p^*_{m}}{\sum_j p^*_j} - \sum_{n=1}^N z_np^*_n 
\;\;= \;\; \frac{\frac{1}{M_t} \sum_{m\in\cM_t}z_m N p^*_{m}}{\frac{1}{M_t} \sum_j Np^*_j} - \frac{1}{N} \sum_{n=1}^N z_n Np^*_n\nonumber\\
  & = \frac{\bar{X}(R)}{\bar{X}(P^*)} - \frac{\mu(R)}{\mu(P^*)}  , \label{lastrr}
\end{align}
  where the last equality makes use of the sample and population means of the
  two $N$-dimensional vectors, namely $\{x_n(R) = z_n Np^*_n\}$ and the
  vector with components $x_n(P^*)$ introduced in the proof of~\reflem{lem:objbias}.

The Taylor expansion of any smooth function $h(u,v)$ is given by  
\begin{align*}
h(u,v) &= h(u_o,v_o) + \nabla_{\theta} h(u_o,v_o) {(u-u_o) \choose (v-v_o)}
+ {(u-u_o) \choose (v-v_o)}^T\nabla_{\theta}^2 h(u_o,v_o) {(u-u_o) \choose (v-v_o)}
+ r(u,v, u_o,v_o),
    \end{align*}
where the higher order terms $r(u,v, u_o,v_o)$ are $o(\|u-u_o\|\cdot\|v-v_o\|)$.
Applying this to $h(u,v)=u/v$ with $u=\bar{X}(R)$,  $u_o=\mu(R)$,
$v=\bar{X}(P^*)$, $v_o=\mu(P^*)$ 
and $Y=r( u,v, u_o,v_o )$, we obtain
\begin{align*}
|h(u,v) - h(u_o,v_o)| 
&
= \left|\frac{\bar{X}(R)}{\bar{X}(P^*)} -  \frac{\mu(R)}{\mu(P^*)}\right| \\
&  \le \frac{1}{\mu(P^*)} \left| \bar{X}(R)-\mu(R) \right| + \frac{\mu(R)}{\mu(P^*)^2}\left|  \bar{X}(P^*)-\mu(P^*)\right| 
\\&\;\;\;
+ \left|\frac{2\mu(R)}{\mu(P^*)^3} (\bar{X}(P^*)-\mu(P^*))^2 
\;\; 
- \;\;\frac{1}{\mu(P^*)^2} (\bar{X}(R)-\mu(R))(\bar{X}(P^*)-\mu(P^*)) + Y\right|.
\end{align*}
Note that $\mu(P^*)=1$ and $\mu(R)>0$ since the individual scenario losses are non-negative.
Further, the proof of \reflem{lem:phifeas} shows that $\bar{X}(P^*) > 0$ with probability at least $(1-{\tau}_1)$ as assumed here.
This avoids the pathological case where the Taylor expansion of $h(u,v)$ above is undefined because $v$ or $v_o$ is zero. 
The above expression then yields
\begin{align*}
\Prb_{t} \left( \eta_t^{-1}\left| z^T\tilde{P}^*_{t} - z^T{P}^* \right| > \frac {c_1} 2 \right) 
& \le  \Prb_{t} \left( \eta_t^{-1}\left| \bar{X}(R) - \mu(R) \right| > \frac {c_1} 6 \right)
+\Prb_{t} \left( \eta_t^{-1}\left| \bar{X}(P^*) - 1 \right| > \frac {c_1} {6|\mu(R)|} \right)   \\
& \;\;
+ \Prb_{t} \bigg( \eta_M^{-1}\bigg| \frac{2\mu(R)}{\mu(P^*)^3} (\bar{X}(P^*)-\mu(P^*))^2
- \frac{1}{\mu(P^*)^2} (\bar{X}(R)-\mu(R))(\bar{X}(P^*)-\mu(P^*)) + Y \bigg| > \frac {c_1} 6 \bigg) .
\end{align*}
From the previous applications of Chebychev's inequality in the
proof of \reflem{lem:phifeas}, we have that
the first two terms are of order $\eta_t^{2\delta/(1-\delta)}$.
The probability of the second term is already included in $\hat{\tau}$
of~\reflem{lem:phifeas}, and thus $\tau_3$ is the additional probability
term that arises from the first term. Hence, each of the random variables in the
first two terms are $O_p(\eta_t)$. The last term involves higher powers of the same random variables that appear in the first two terms.
We show that, as expected, the probability of these terms is $o(\eta_t^{2\delta/(1-\delta)})$,
or in other words, the random variables represented in these terms are $o_p(\eta_t)$.

Taking the symmetric random variable\ $Z=|\bar{X}(P^*)-1|$ and an integer $j>1$, then
 $\Prb(|Z|^j > \eta_t) = \Prb(|Z| > \eta_t^{1/j}) \le \Prb( |Z| > \eta_t^{1-\delta'})$,
where $\delta' < \delta$ is smaller than the $\delta$ used in the definition of $\eta_t$.
Applying Chebychev's inequality renders $\Prb(|Z|^j>\eta_t) \le O(\eta_t^{2\delta/(1-\delta)+2\delta'/(1-\delta)}) = o(\eta^{2\delta/(1-\delta)}_t)$.
The same logic holds for each of the other summands in the remainder,
  and thus the last term is of smaller order than the first two.
Hence, the total probability in this tail term is $o(\eta_t^{2\delta/(1-\delta)})$,
which yields the final result. \ProofEnd

\ProofOf{\refthm{thm:bias}}
Recall that the constants $M'$ and $\bar{\tau}_t$ are defined in the statements of~\reflem{lem:phifeas} and~\reflem{lem:objbias}, respectively.
Given the robust loss function $R(\theta) = \sum_n l(\theta, \xi_n)
p^*_n$ and our approximation 
$\hat{R}_{t}(\theta) = \sum_{m\in\cM} l (\theta,\xi_m) \hat{p}_m$
constructed from the subsampled $\cM$, the mean-value theorem of calculus yields
\begin{align*}
(\nabla_{\theta} l(\theta, \xi_n))_u &= \frac{\partial l(\theta,\xi_n)}{\partial \theta_u} 
= \frac{1}{h_{u,n}} (l(\theta+h_{u,n} \be_u, \xi_n)-l(\theta, \xi_n)),
\end{align*}
where $h_{u,n}$ is a small positive value that depends on the component $\theta_u$
and on the sample $\xi_n$, with $\be_u$ the unit-vector in the $u$-th coordinate. 
Let $\underbar{h} = \min_{u,n} h_{u,n}$.
We then have
\begin{align*}
\left| ( \nabla_{\theta}\hat{R}_{t}(\theta) - \nabla_{\theta} R(\theta))_u \right|
& \le \frac{1}{\underbar{h}}   \bigg| \sum_n l(\theta+h_{u,n}\be_u,\xi_n)^T(p^*_n-\hat{p}_n)
- l(\theta,\xi_n)^T (p^*_n-\hat{p}_n) \bigg| \\
& \le \frac{1}{\underbar{h}} \bigg| \Big[ \sum_n  l(\theta+h_{u,n}\be_u,\xi_n)^T(p^*_n-\hat{p}_n )\Big] \bigg|
+ \frac{1}{\underbar{h}} \bigg| \Big[ l(\theta,\xi_n)^T (p^*_n-\hat{p}_n)\Big]\bigg|.  
\end{align*}
Applying the same arguments as those used in the proof of \reflem{lem:objbias},
together with squaring and combining these terms over all $u$, renders
the desired final result. 
\ProofEnd

%% file: apdx_cvg.tex
\subsection{Convergence of \refeq{ssgd}}\label{apdx:cvg}

\ProofOf{\refthm{thm:main-gen}}

For any $\theta$ and a set $\cM_t$ sampled to have $M_t$ support points, \refthm{thm:bias} and~\refasm{asm:props}(iii) show that
\begin{align*}
\E_{t} \left[ \|\nabla_{\theta}\hat{R}_{t}(\theta) - \nabla_{\theta}{R}(\theta)\|^2_2 \right]
&
\le \E_{t} \left[ \|\nabla_{\theta}\hat{R}_{t}(\theta) - \E_{t} [\nabla_{\theta}\hat{R}_{t}(\theta)]\|^2_2 \right]
+ \| \E_{t} [\nabla_{\theta}\hat{R}_{t}(\theta)]  - \nabla_{\theta}{R}(\theta)\|^2_2 \\
&
\le O(\eta_t^{2/(1-\delta)}) + O(\eta_t^{2}) \; = \; O(\eta_t^{2}). 
\end{align*}
Hence, the slower rate of decrease in the bias prevails as the rate at which the
mean squared error decreases to zero.
Elementary algebraic manipulations yield the following two implications:
\begin{align}
\E_{t} \left[ \|\nabla_{\theta}\hat{R}_{t}(\theta) \|^2_2 \right] \quad &\le \quad C \eta_t^2 + \| \nabla_{\theta}{R}(\theta) \|^2_2 \label{biasimpl1} \\
%
\mbox{and}\qquad
%
- \E_{t} \left[ \left(\nabla_{\theta}\hat{R}_{t}(\theta)\right)^T\nabla_{\theta}{R}(\theta) \right]  \quad &\le \quad C {\eta_t^2} - \| \nabla_{\theta}{R}(\theta) \|^2_2
-  \E_{t} \left[ \|\nabla_{\theta}\hat{R}_{t}(\theta) \|^2_2 \right] . \label{biasimpl2}
\end{align}

We can therefore bound the expected robust loss at step $(t+1)$ using
\begin{align}
R_{\inf} &\le \E_{t}[ R(\theta_{t+1}) ] \nonumber\\
&\le \E_{t}[ R(\theta_t) ]  - \gamma \E_{t} [ \nabla_{\theta} R(\theta_t)^T\nabla_{\theta} \hat{R}_{t}(\theta_t)] 
+ \frac{L \gamma^2}{2} \E_{t}\Big[\|\nabla_{\theta}\hat{R}(\theta_t)\|_2^2\Big] \nonumber\\
&\le \E_{t}[ R(\theta_t) ] + \frac{C\gamma\eta_t^2}{2} - \frac{\gamma}{2} \|\nabla_{\theta} R(\theta_t)\|_2^2  
+ \bigg(\frac{L\gamma^2-\gamma}{2} \bigg)\bigg(\|\nabla_{\theta} R(\theta_t)\|_2^2 + C\eta_t^2\bigg)\nonumber\\
&= \E_{t}[ R(\theta_t) ] + \frac{C\gamma\eta_t^2}{2} (L\gamma+1) 
-\frac{\gamma}2 (2-L\gamma) \|\nabla_\theta R(\theta_t)\|^2_2 .\label{main:funda}
\end{align}
Upon rearranging~\eqref{main:funda} and telescoping back to the initial
iterate $\theta_0$, we obtain the desired final result.
\ProofEnd

We now present a structural lemma that builds up to the proof of our final
result, namely~\refthm{thm:main}.
Recall from~\refprop{prop:im_cmp_bnd} that, in our case, the $t$-th iterate has a computational burden of $w_t = O(M_t \log M_t)$. 

\begin{lemma}\label{lem:workdone}
~~(i) If $M_t$ is a constant-factor growth sequence with $\nu_t=\nu>1$, then
there exist constants $c_1,c_2$ such that $c_1 \leq \frac{W_t}{w_t} \leq c_2$
as $t \rightarrow \infty$;
~~(ii) If $M_t$ is a diminishing-factor growth sequence, then
$\frac{W_t}{w_t} \rightarrow \infty$ as $t \rightarrow \infty$.
\end{lemma}
%

%
\begin{proof}
For (i), 
when $w_t=O(M_t\log M_t)$ and $M_t=\nu^{t}M_0$, we obtain
\begin{align*}
W_t &= \sum_{s\le t} M_s\log M_s = M_0 \log (M_0\nu)\,\, \sum_{s\le t} {s} \nu^{s}
= \frac{M_0\log(M_0\nu)}{\nu} \,\,\sum_{s\le t} (s)\nu^{(s-1)} 
\\ &
= \frac{M_0\log(M_0\nu)}{\nu}\frac{\partial}{\partial\nu}\left(\sum_{s\le t}\nu^{s}\right)
\;\;=\;\; \frac{M_0\log(M_0\nu)}{\nu} \frac{t\nu^{(t+1)}-\nu^{t}+1}{(\nu-1)^2}. \\
\end{align*}
Dividing the last expression by $w_t=O(t\nu^{t}M_0\log(M_0\nu))$ renders
\begin{equation}
\frac {W_t}{w_t} = \frac 1 {\nu -1} \left( 1+ \frac 1 t + \frac 1 {t\nu^t} \right), 
\end{equation}
from which the requisite constants can be worked out.

For (ii), first consider any small $\epsilon > 0$.
Since we have that $\nu_t\nearrow 1$ for sub-geometric growth of $\{M_t\}$,
then for a sufficiently large $t$ there exists $t_0(\epsilon)$ such that
$$1 \,\,\ge\,\, \nu_s \,\,\ge\,\, \left(\frac{1}{t^{\epsilon/2}}\right)^{1/t^{\epsilon}} \,\,\ge\,\, 0,
\quad \forall t\ge s \ge t_0(\epsilon).$$
We then obtain
\[ \frac { W_t} {w_t} \ge \sum_{s=t-t^{\epsilon}}^t \prod_{u=s+1}^t \nu_u \ge
t^{\epsilon} \prod_{u=t-t^{\epsilon}}^t \nu_u \ge
t^{\epsilon/2}.
\]
Hence, as $t\tndi$, we have that $W_t/w_t\tndi$, thus proving the result.
\end{proof}

%

%
\ProofOf{\refthm{thm:main}}
We begin with the $c$-strong convexity of $R(\theta)$ under~\refasm{asm:props}(iii).
Since each $l(\theta,\xi_n)$ is $c$-strongly convex, we have
\begin{align*}
&
l(\theta_1,\xi_n) + \nabla_{\theta} l(\theta_1,\xi_n)^T(\theta_2-\theta_1) + \frac{c}{2} \|\theta_2-\theta_1\|^2_2
\le l(\theta_2,\xi_n).
\end{align*}
Taking any pmf $P$ with components $p_n$ and summing up each side, we obtain
\begin{align*}
&
\sum_n p_n \Big(l(\theta_1,\xi_n) + \nabla_{\theta} l(\theta_1,\xi_n)^T(\theta_2-\theta_1)  + \frac{c}{2} \|\theta_2-\theta_1\|^2_2\Big)
\le \sum_n p_n l(\theta_2,\xi_n).
\end{align*}
Since the above holds for any $P$, apply this for $P^*(\theta_1)$,
the optimal pmf for the inner maximization that defines $R(\theta_1)$,
with components $p^*_n(\theta_1)$.
As previously discussed, if the $D_{\phi}$-constraint is tight enough,
i.e., $\rho < \rho_1$, then $P^*(\theta_1)$ is unique for $\theta_1$, and thus the subgradient $\nabla_{\theta} R(\theta)$ corresponds with the gradient from Danskin's Theorem
(see~\cite[Theorem 7.21, p.~352]{shap09}).
We then derive
\begin{align*}
&
\bigg(\sum_n p^*_n(\theta_1) l(\theta_1,\xi_n)\bigg)
+ \sum_n p^*_n(\theta_1)\nabla_{\theta} \left(l(\theta_1,\xi_n)\right)^T (\theta_2-\theta_1) + \frac{c}{2} \|\theta_2-\theta_1\|^2_2
\;\;\;\le\;\;\; \sum_n p^*_n(\theta_1) l(\theta_2,\xi_n),
\end{align*}
which verifies that $R(\theta)$ is $c$-strongly convex:
\begin{align*}
&
R(\theta_1) + \nabla_{\theta} R(\theta_1)^T(\theta_2-\theta_1) + \frac{c}{2} \|\theta_2-\theta_1\|^2_2 
\;\;\le\;\; \max_{P\in\cP} \sum_n p_n l(\theta_2,\xi_n) = R(\theta_2).
\end{align*}

Let us next return to the point at which the proof of~\refthm{thm:main-gen} terminated, namely \eqref{main:funda}.
Recall that, for the strongly convex case, there exists a unique minimizer $\theta_{\rob}$ such that $R(\theta_{\rob})=R_{\inf}$.
We rewrite~\eqref{main:funda} to obtain
\begin{align}
\E_{t}[ R(\theta_{t+1}) ] - R(\theta_{\rob}) 
&\le 
\bigg(\E_{t}[ R(\theta_t) ] - R(\theta_{\rob})\bigg) - \gamma \E_{t} [ \nabla_{\theta} R(\theta_t)^T\nabla_{\theta} \hat{R}_{t}(\theta_t)]  
+  \frac{L \gamma^2}{2} \E_{t}\Big[\|\nabla_{\theta}\hat{R}(\theta_t)\|_2^2\Big] \nonumber\\
&\le \bigg(\E_{t}[ R(\theta_t) ] - R(\theta_{\rob})\bigg) + \frac{C\gamma\eta_t^2}{2} - \frac{\gamma}{2} \|\nabla_{\theta} R(\theta_t)\|_2^2 
+ \bigg(\frac{L\gamma^2-\gamma}{2} \bigg)\bigg(\|\nabla_{\theta} R(\theta_t)\|_2^2 + C\eta_t^2\bigg)
\nonumber\\&
\le \bigg(\E_{t}[R(\theta_t)] - R(\theta_{\rob})\bigg)\Big(1-\frac{2\gamma - L\gamma^2}{4c}\Big) + \frac{CL\gamma^2\eta_t^2}{2}\nonumber \\
&\le \bigg(1 - \frac{\gamma}{4c}\bigg) \bigg( \E_{t}[R(\theta_t)] -
R(\theta_{\rob})\bigg) + \frac{CL\gamma^2\eta_{t}^2}{2}. \label{onestep}
\end{align}
The first inequality starts with the $L$-Lipschitzness of $\nabla_{\theta} R(\cdot)$, and the second inequality substitutes the relations in \eqref{biasimpl1} and \eqref{biasimpl2}.
The third inequality uses the $c$-strong
convexity of $R(\theta)$, specifically the implication that $\|\nabla_{\theta} R(\theta)\|_2^2/2c \ge (R(\theta) - R(\theta_{\rob})) $. The final
inequality utilizes the conditions imposed on $\gamma$.

Let $r=1-\gamma/4c < 1$.
The form of~\eqref{onestep} is quite informative, in that it clearly displays the tradeoff
being addressed by the algorithm: the first summand provides a geometric reduction in
the optimality gap, which is to be balanced with the stochastic error
in the second summand. Note that, for growth sequences $M_t = M_o\prod_{s=1}^t \nu_s$ where $\nu_s>1$, we have
\begin{align*}
\frac {\eta_t^2}{\eta^2_{t-1}} & = \left( \frac {\frac 1 {M_t} - \frac 1 N} {\frac 1 {M_{t-1}} - \frac 1 N}\right)^{(1-\delta)}
= \frac 1 {\nu_t^{(1-\delta)}} \left( \frac {\frac 1 {M_0\prod_{s=1}^t\nu_s} - \frac 1 N} {\frac 1 {M_0\prod_{s=1}^t\nu_s} - \frac 1 N + \frac 1 N \left(1-\frac 1 {\nu_t}\right) } \right)^{(1-\delta)}
\le \frac 1 {\nu_t^{(1-\delta)}}.
\end{align*} 
Therefore, the $\eta^2_t$ decreases at least as fast as a factor of $\nu_t^{-(1-\delta)}$ as $t\uparrow$.

Recalling the definition $E_{t} \,\DefAs\, \E_{t}[R(\theta_{t})]-R(\theta_{\rob})$, 
we telescope the optimality gap in the first summand of~\eqref{onestep} to render it as
\begin{align}
E_{t+1}
&\le r E_t + \frac{CL\gamma^2\eta_{t}^2}{2} 
\le r^t E_0  + \frac{CL\gamma^2}{2} \sum_{s=0}^t  \eta_{s}^2 r^{t-s} 
\;\;\le\;\; r^t E_0  + \frac{CL\gamma^2}{2} \underbrace{\sum_{s=0}^t r^{t-s} \left(\prod_{u=0}^s \nu_u^{-(1-\delta)}\right)}_{I}. \label{eq:DAstochbal}
\end{align}
From~\eqref{eq:DAstochbal}, we observe that the error after $t$ steps is a result of the balance
between $r$ and $\nu_t$, and the key term in analyzing this balance is denoted by $I$.

Consider the constant growth sequences where $\nu_s = \nu^s$ for a $\nu>1$.
For case (i) in the theorem, further assume that $\nu^{(1-\delta)} < 1 /r$.  Then, the term $I$ can be written as
\begin{align*}
&
I = \frac 1 {\nu^{t(1-\delta)}} \Bigg[ 1 + r \nu^{(1-\delta)} + \ldots+ (r\nu^{(1-\delta)})^{t-1} \Bigg]
=  \frac 1 {\nu^{t(1-\delta)}} \Bigg[\frac {1-(r \nu^{(1-\delta)})^t }{1-r \nu^{(1-\delta)} } \Bigg] 
\le \frac 1 {\nu^{t(1-\delta)}} \Bigg[\frac 1{1-r \nu^{(1-\delta)} } \Bigg] .
\end{align*}
The last expression in turn shows that the optimality gap $E_{t+1}$ decreases at the dominating (slower) rate of $\nu^{t(1-\delta)}$ such that the term $I$ vanishes to zero.
On the other hand, we have from~\reflem{lem:workdone}(i) that, for constant growth sequences, $W_t \le w_t c_2$. Combining both of these yields that there exists a constant $K_1$ such that
$W_t E_{t+1} \le K_1 t \nu^{t\delta},$
as desired, where
\begin{equation}\label{geomk1}
K_1 = \frac {c_1}{\nu -1} M_o\nu \log (M_0 \nu)\bigg(E_0 + \frac {CL\gamma^2} {2 (1-r \nu^{(1-\delta)} )} \bigg).
\end{equation}

For (ii), we now assume that $\nu^{(1-\delta)} > 1 /r$.
In this case, the term $I$ can be written as
\begin{align*}
I = r^t \Bigg[ 1 + \frac 1 {r \nu^{(1-\delta)}} +  \ldots+ \frac 1 {(r\nu^{(1-\delta)})^{t-1}} \Bigg]
=  r^t  \Bigg[\frac {1-\frac 1 {(r \nu^{(1-\delta)})^t} }{1-\frac 1 {r \nu^{(1-\delta)}} } \Bigg]
\;\;\le\;\; r^t  \Bigg[ \frac 1 { 1- \frac 1 {r \nu^{(1-\delta)}} } \Bigg].
\end{align*}
The
last expression shows that the optimality gap $E_{t+1}$ is dominated by the (slower) rate of the deterministic convergence $r^t$.
Combining this with~\reflem{lem:workdone}(i) yields that 
$W_t E_{t+1} \le K_2 t (r\nu)^{t},$
as desired, where
$$K_2 = \frac {c_1}{\nu -1} M_o\nu \log (M_0 \nu)\bigg(E_0 +  \frac {CL\gamma^2} {2(1-\frac 1 {r \nu^{(1-\delta)} } ) }\bigg).$$

Finally, we consider the case (iii) where the growth factor sequence is diminishing in size: $\nu_s\searrow 1$. 
This case is similar to case (i) in that there exists sufficiently large $s_o$ such that  $\nu_s^{(1-\delta)} r <1 $ for all $s\ge s_o$.
A similar analysis then shows that the dominating (slowly converging) term in the expression for $E_{t+1}$ is the summand for $I$ of size $\prod_{s=0}^t \nu_s^{-(1-\delta)} = O(M_t^{-(1-\delta)})$. 
On the other hand, we know from~\reflem{lem:workdone} that the computational effort $w_t=O(M_t\log M_t)$ is also $w_t=o(W_t)$. These together imply the desired result for this case.
\ProofEnd

We note that the arguments in the proof of \refthm{thm:main} also provide a method to obtain a result in the opposite direction,
namely a method for setting the growth sequence that ensures a minimal amount of computation and yields a certain level of accuracy.
More specifically, given a desired level of accuracy, these arguments can be applied to determine the optimal sequence of $M_t$ that achieves the desired level of accuracy with a minimal amount of total effort $W_t$.

For strongly convex $R(\theta)$,
the major implication of~\refthm{thm:main} is that the best setting of the growth parameter sequence is $\nu_t=\nu$ for some $\nu\in(1, r^{-1/(1-\delta)})$
in order to balance the computational effort $W_t$ with the expectation of the optimality gap $E_t$.
Here $r$ is defined from~\eqref{onestep} to be the geometric (also called linear) decrease in the deterministic error and $\delta$ is the small constant introduced by~\refthm{thm:bias}.
The reason for this setting of the growth parameter sequence is because only under such growth sequences can the product $E_t W_t$ be bounded above by slowly growing functions of the iteration count $t$,
and this upper bound is nearly linear in $t$ because $\delta$ is usually chosen to be small.
When $\nu_t$ grows in a slower fashion, we have that $E_t = o(W_t^{-1})$;
and when $\nu_t$ is chosen to grow faster, the upper bound on $E_tW_t$ shows exponential growth in $t$ or, in other words, $E_t$ drops to zero slower than the inverse of the effort $W_t$. 

The constant associated with this near-linear upper bound when $\nu \in (1, r^{-1/(1-\delta)})$ is given as $K_1:= K_1(\nu)$ defined in~\eqref{geomk1}.
This constant is finite and positive everywhere in the open interval $(1, r^{-1/(1-\delta)})$, but grows to infinity as $\nu$ approaches either end point.
Hence, non-linear optimization techniques can be employed to find a good local (or even a global) minima of the $K_1(\nu)$ in this interval that will yield the best balance between $E_t$ and $W_t$. 


%% file: apdx_expt.tex
%

\begin{figure}[htbp]
	\vskip -0.1in
	\begin{center}
		\includegraphics[width=0.5\columnwidth]{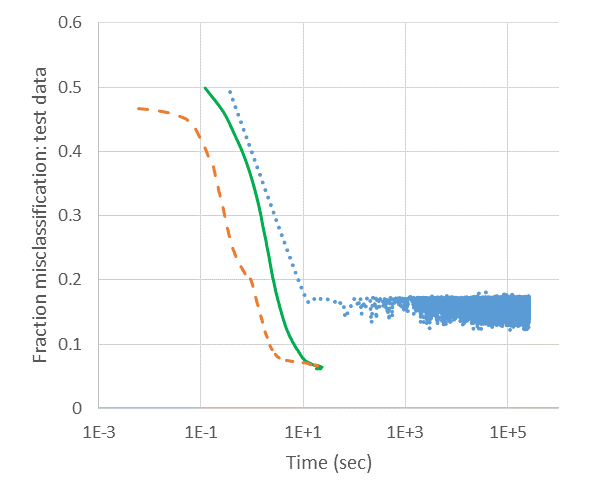}\hskip 0.6in
		\caption{Comparisons of DSSG (orange, dashed lines), FSG (green, solid lines) and single run of \citeN{nd16} algorithm (blue) on fraction of misclassification in testing ($y$-axis) versus computation times ($x$-axis) over HIV-1 dataset with $\rho=0.1$ and log-scale $x$-axis.}
		\label{fig:rho0.1:HIV+nd16}
	\end{center}
	\vskip -0.1in
\end{figure}

In this section we present additional experimental results that complement those in the main body of the paper. 
Section~\ref{ssec:droexpt} provides comparisons among the DSSG, FSG and standard SGD algorithms used to solve the DRO formulation~(\ref{absfmln}) over three additional datasets,
together with further results and details that include the dataset from the main paper.
Section~\ref{ssec:genexpt} provides additional results and details on the Generalization comparison in the main body of the paper. 
%
%
%
%
We will open our github repository for all code used to produce these results for public consumption once the anonymity requirement is resolved.

\subsection{Solving the DRO Formulation}\label{ssec:droexpt}
\begin{figure*}[htbp]
	\vskip -0.1in
	\begin{center}
		\includegraphics[width=0.245\textwidth]{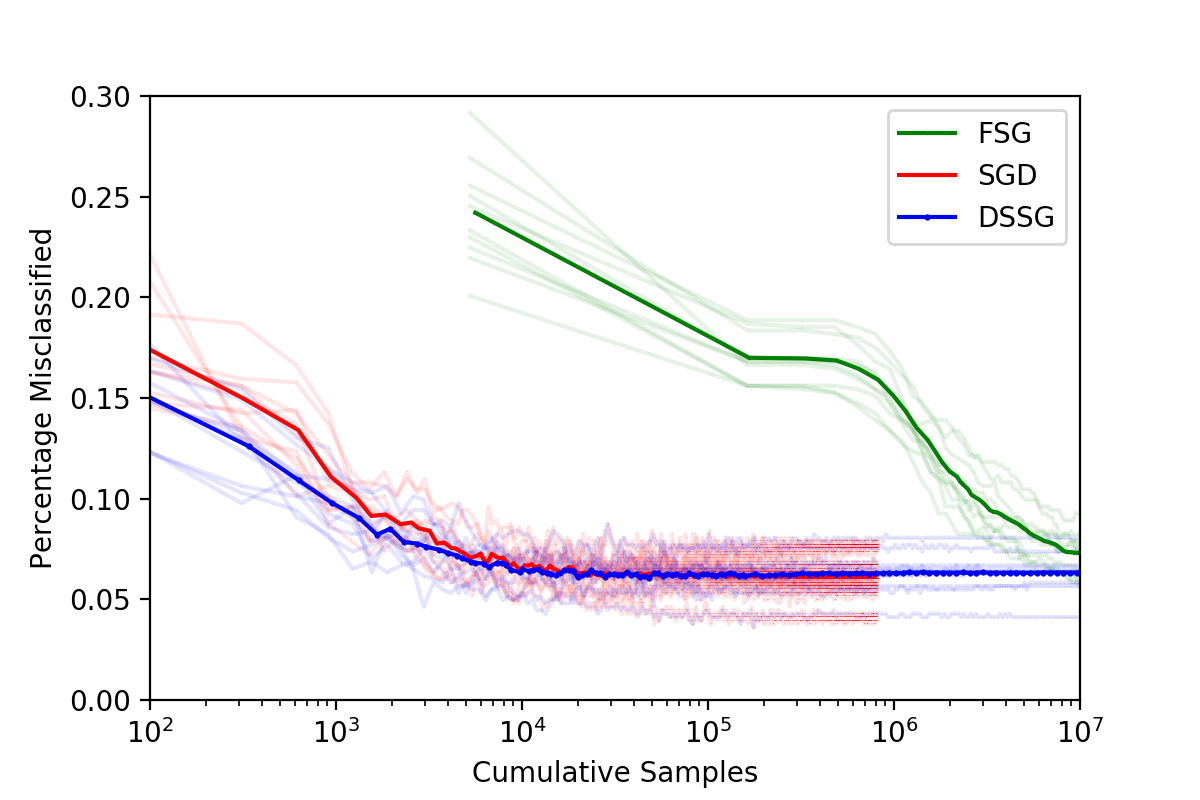}
		\includegraphics[width=0.22\textwidth]{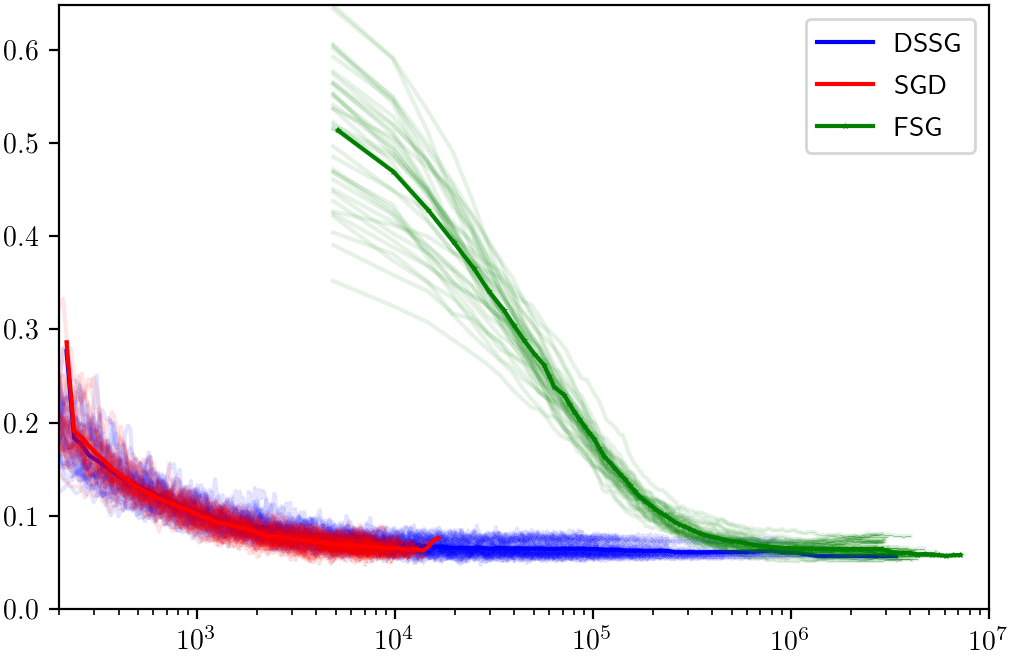}
		\includegraphics[width=0.245\textwidth]{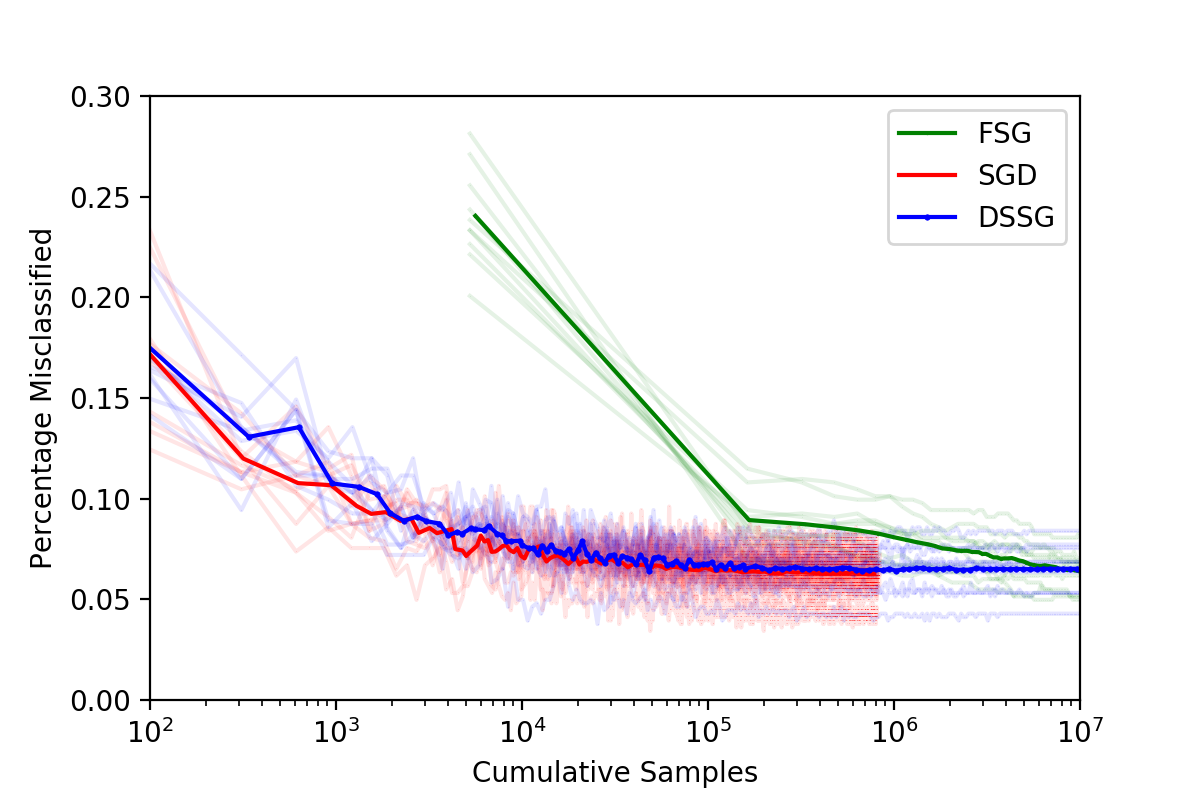}
		\includegraphics[width=0.245\textwidth]{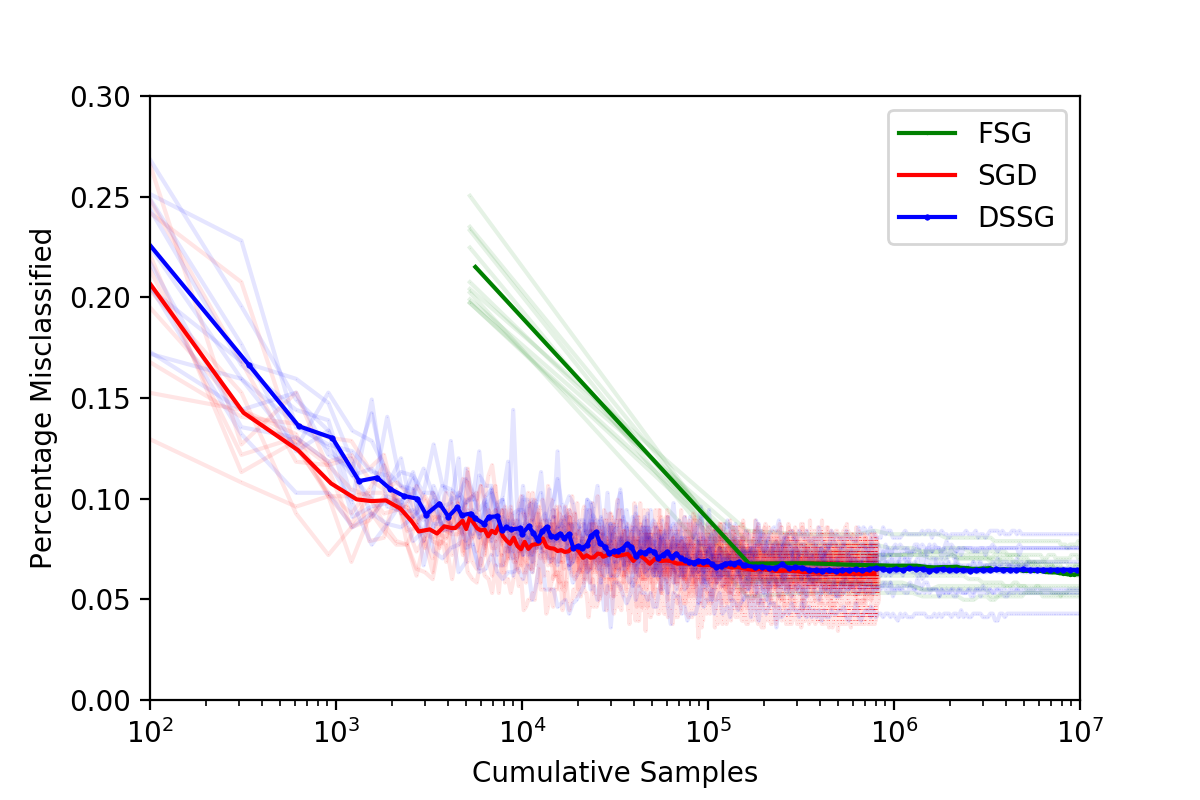}
		\caption{Comparisons of DSSG (blue), FSG (green) and standard SGD (red) on fraction of misclassification in testing ($y$-axis) versus cumulative samples ($x$-axis) over HIV-1 dataset with (first) $\rho=0.01$, (second) $\rho=0.1$, (third) $\rho = 0.5$ and (fourth) $\rho=1.0$.
			Log-scale cumulative samples $x$-axis.}
		\label{fig:hiv1all}
	\end{center}
	\vskip -0.2in
\end{figure*}

\begin{figure*}[htbp]
	\begin{center}
		\includegraphics[width=0.245\textwidth]{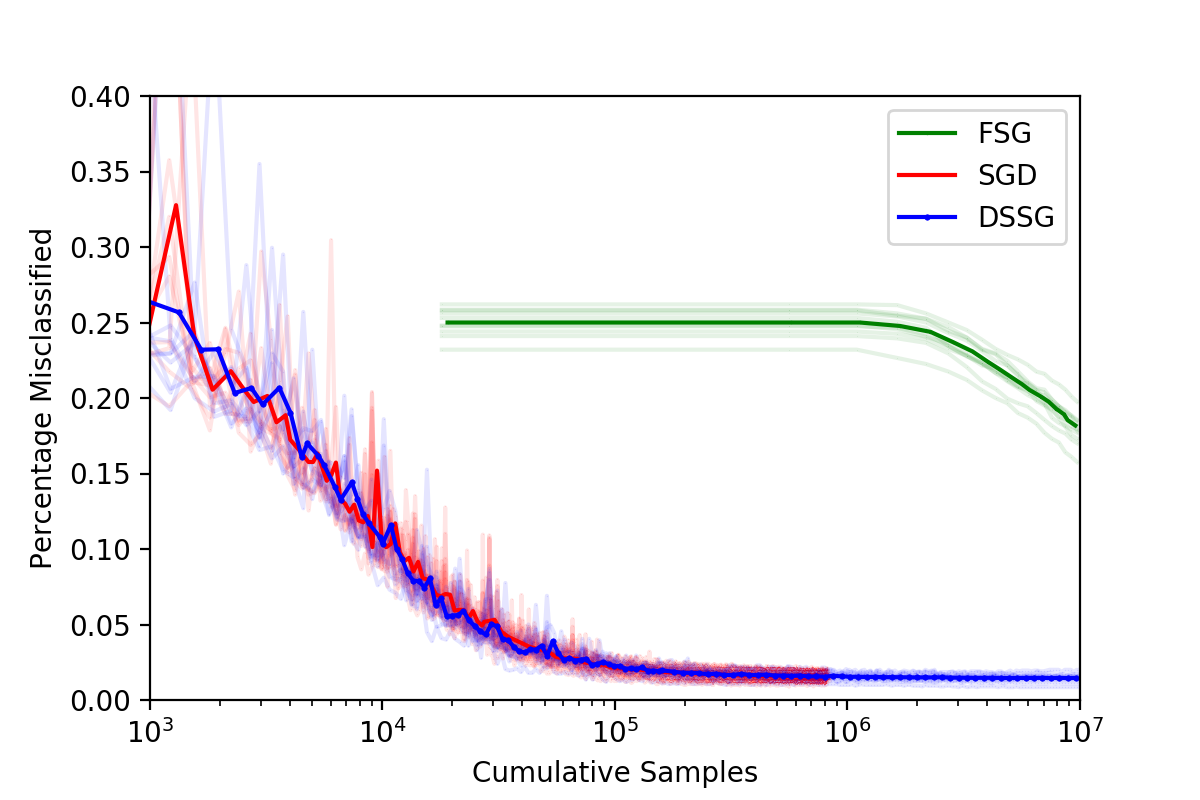}
		\includegraphics[width=0.22\textwidth]{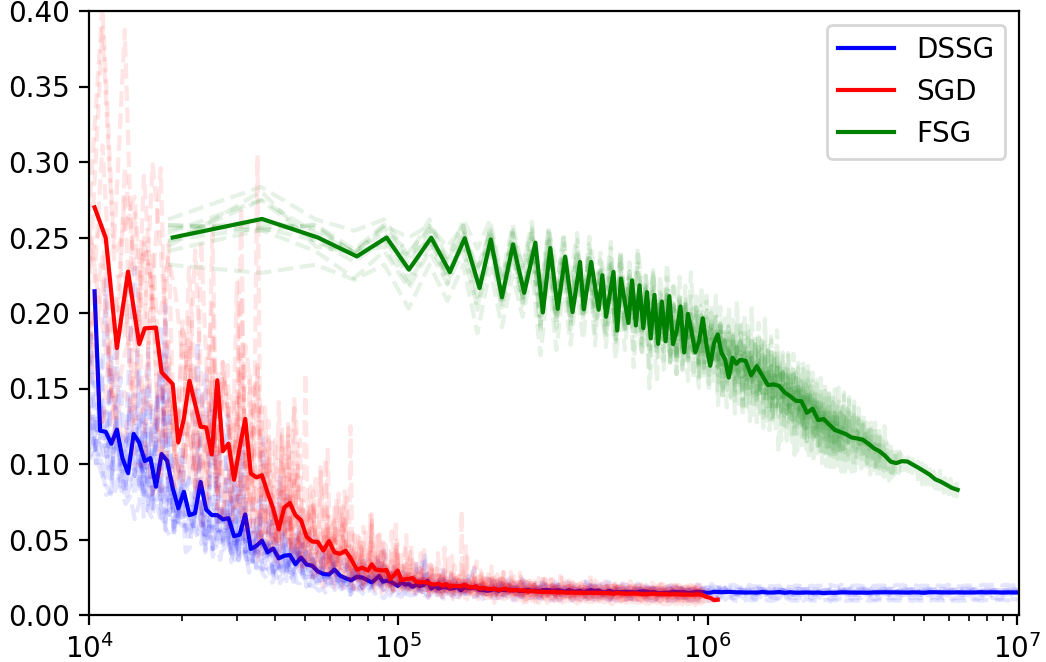}
		\includegraphics[width=0.245\textwidth]{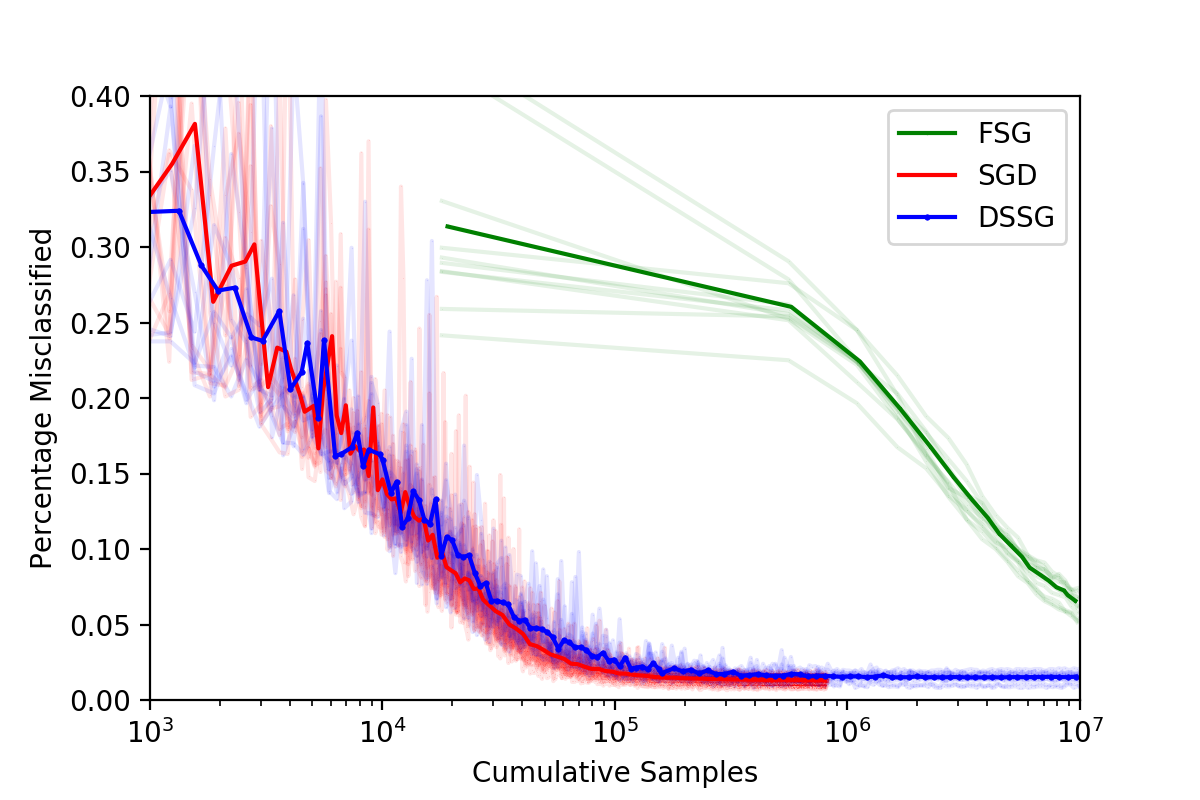}
		\includegraphics[width=0.245\textwidth]{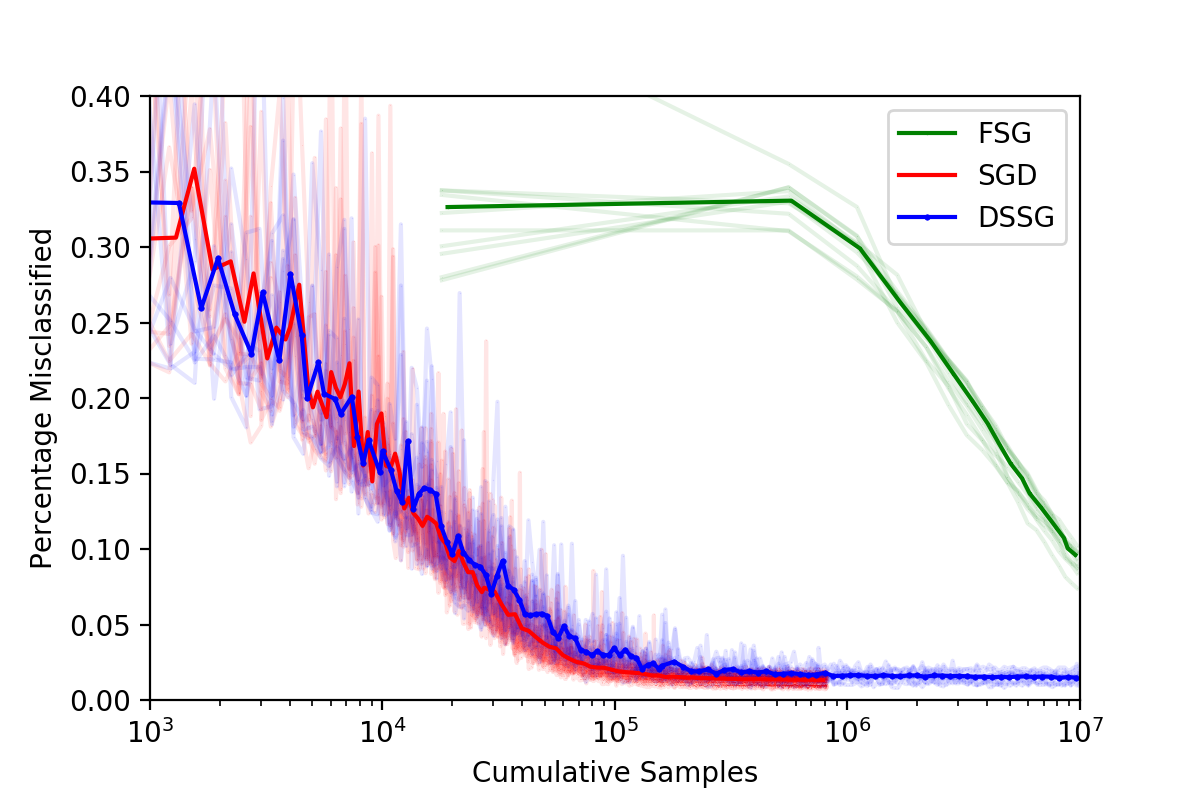}
		\caption{Comparisons of DSSG (blue), FSG (green) and standard SGD (red) on fraction of misclassification in testing ($y$-axis) versus cumulative samples ($x$-axis) over Riccardo dataset with (first) $\rho=0.01$, (second) $\rho=0.1$, (third) $\rho = 0.5$ and (fourth) $\rho=1.0$.
			Log-scale cumulative samples $x$-axis.}
		\label{fig:riccardo.multirho.class}
	\end{center}
	\vskip -0.2in
\end{figure*}

\begin{figure*}[htbp]
	\begin{center}
		\includegraphics[width=0.32\textwidth]{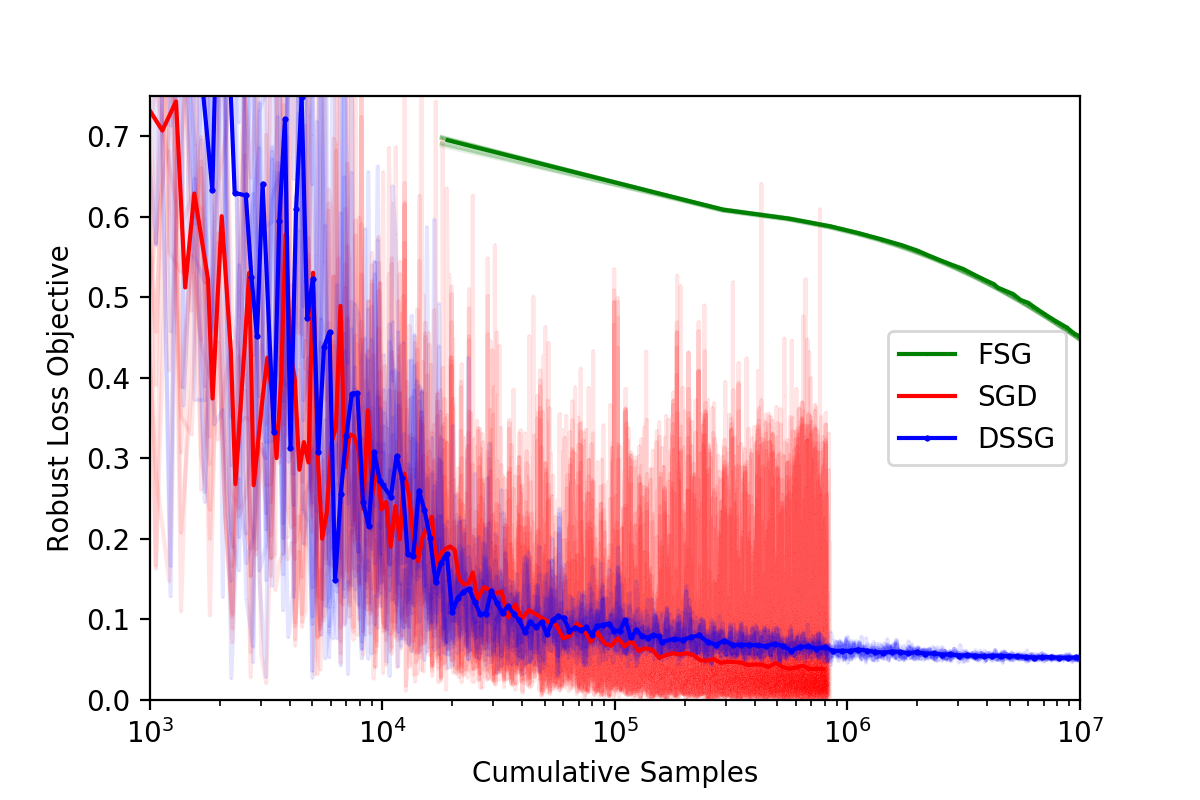}
		\includegraphics[width=0.32\textwidth]{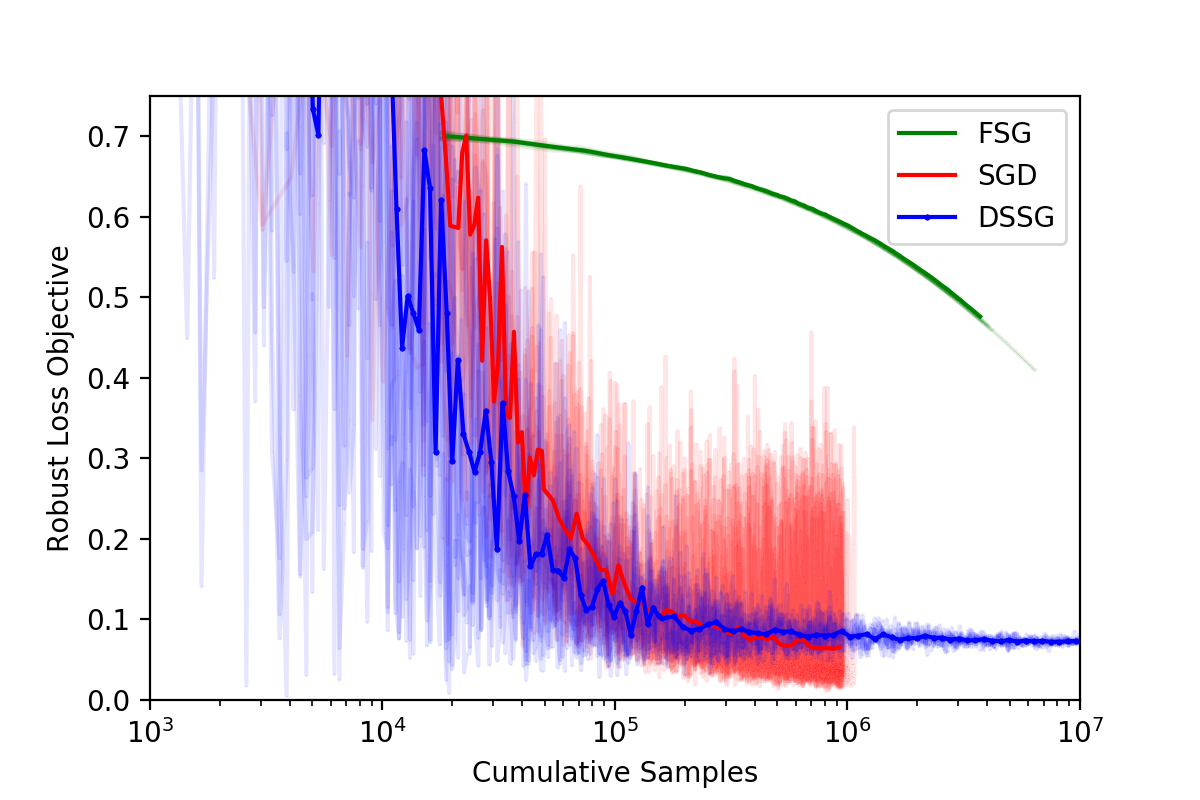}
		\includegraphics[width=0.32\textwidth]{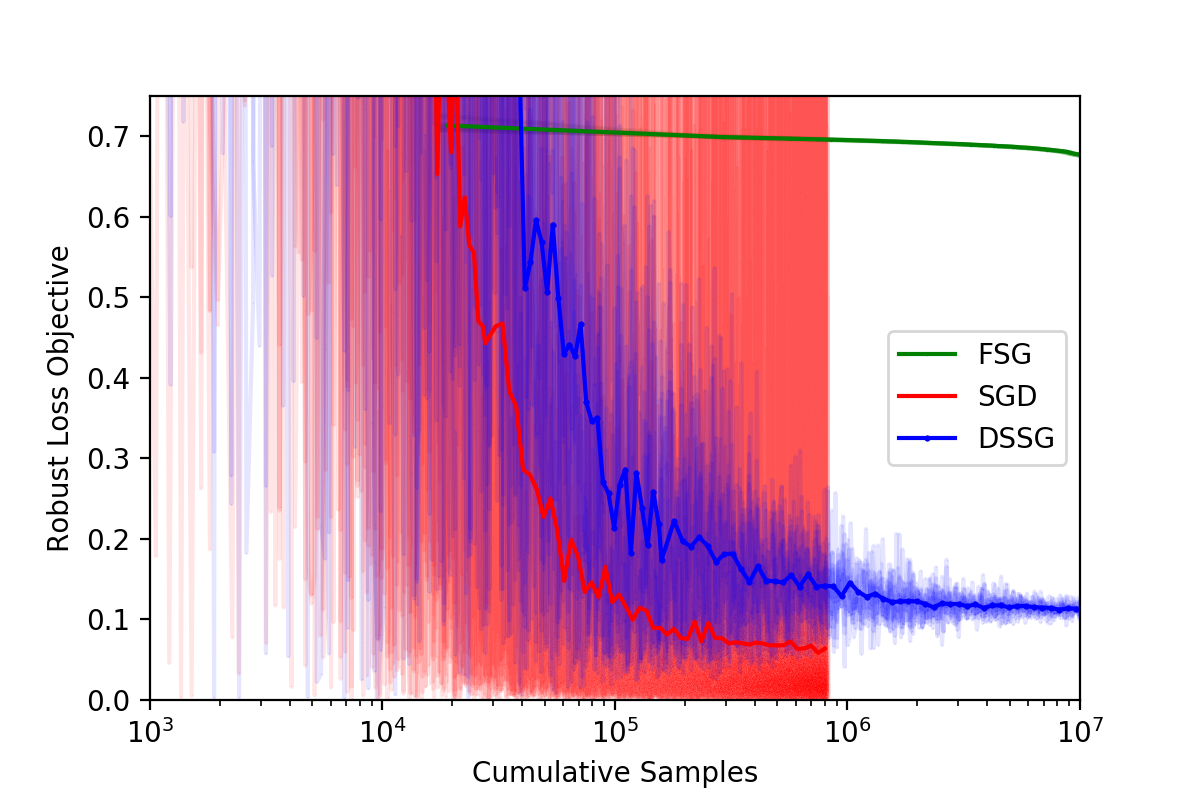}
\caption{Comparisons of DSSG (blue), FSG (green) and standard SGD (red) on training robust loss (minimization objective) ($y$-axis) versus cumulative samples ($x$-axis) over Riccardo dataset with (left) $\rho=0.01$, (center) $\rho = 0.5$ and (right) $\rho=1.0$.
	Log-scale cumulative samples $x$-axis.}
\label{fig:riccardo.multirho.robloss}
	\end{center}
	\vskip -0.2in
\end{figure*}

\begin{figure*}[htbp]
	\vskip -0.1in
	\begin{center}
		\includegraphics[width=0.32\textwidth]{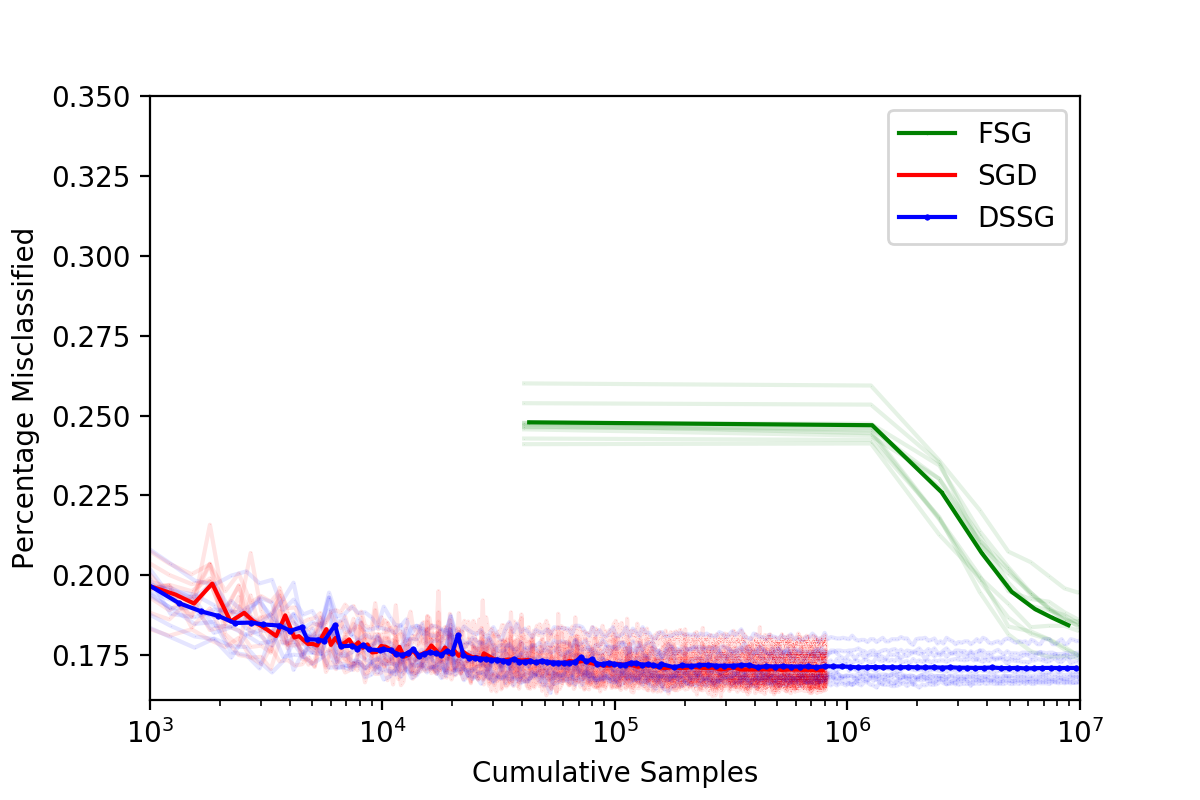}
		\includegraphics[width=0.32\textwidth]{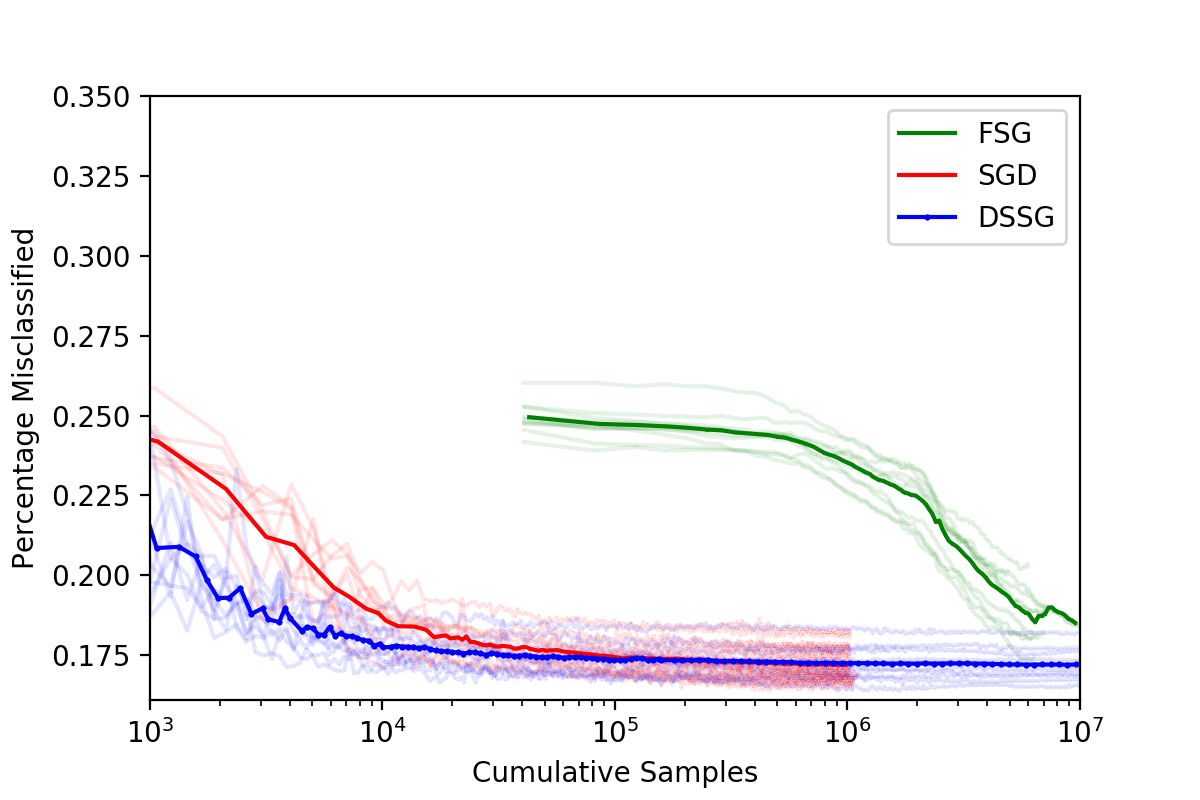}
		\includegraphics[width=0.32\textwidth]{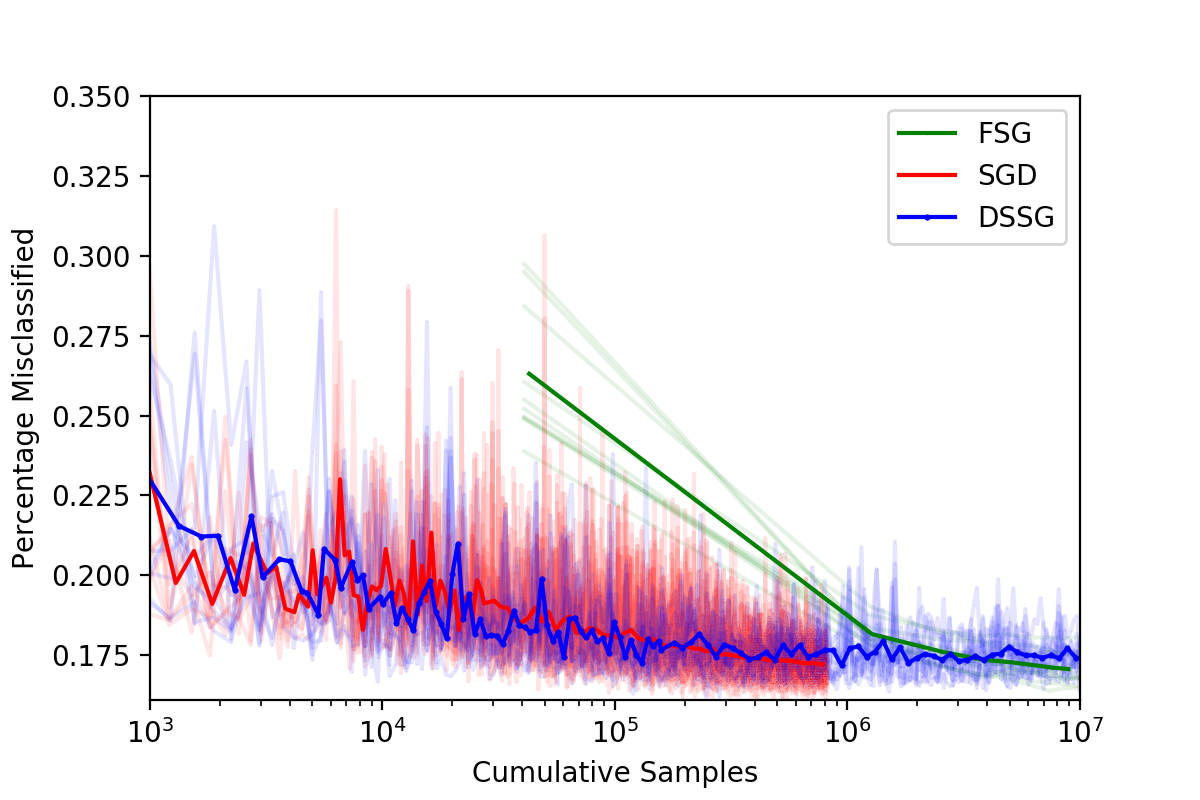}
		\caption{Comparisons of DSSG (blue), FSG (green) and standard SGD (red) on fraction of misclassification in testing ($y$-axis) versus cumulative samples ($x$-axis) over Adult Income dataset with (left) $\rho=0.01$, (center) $\rho=0.1$ and (right) $\rho = 0.5$.
			Log-scale cumulative samples $x$-axis.}
		\label{fig:adult.multirho}
	\end{center}
	\vskip -0.2in
\end{figure*}

\begin{figure*}[htbp]
	\begin{center}
		\includegraphics[width=0.32\textwidth]{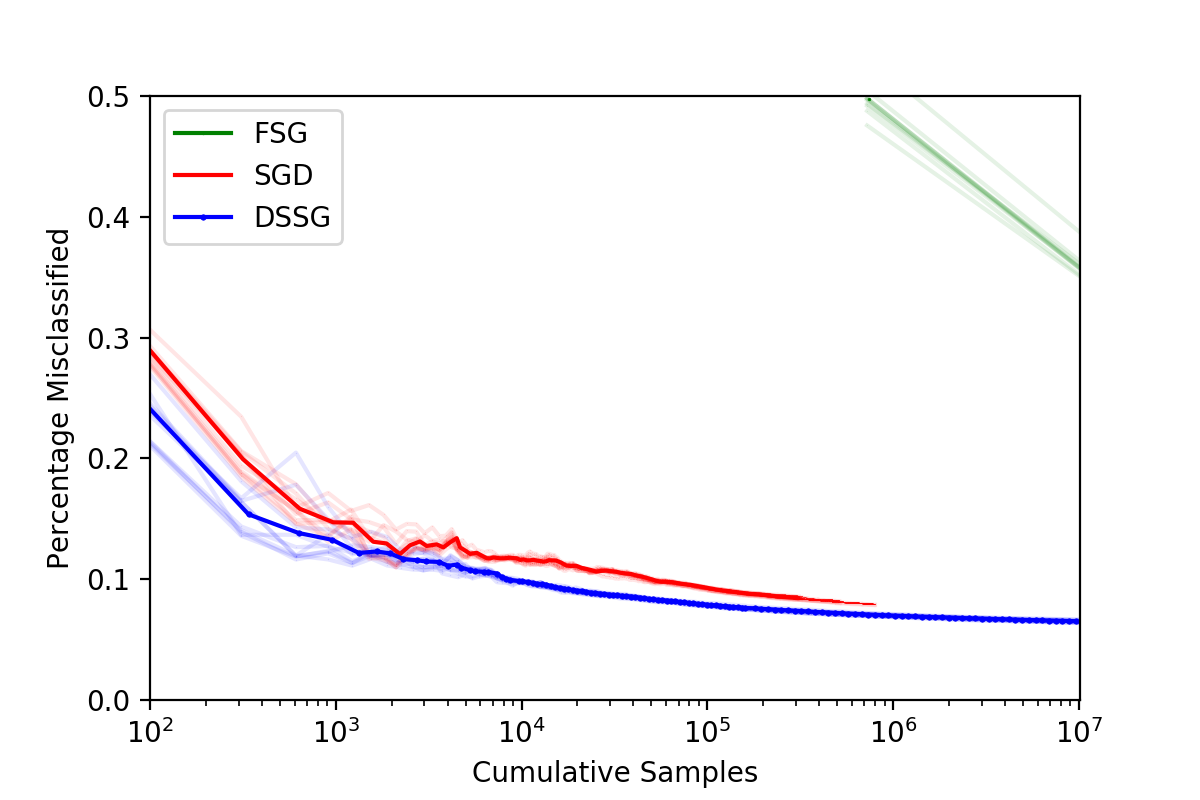}
		\includegraphics[width=0.32\textwidth]{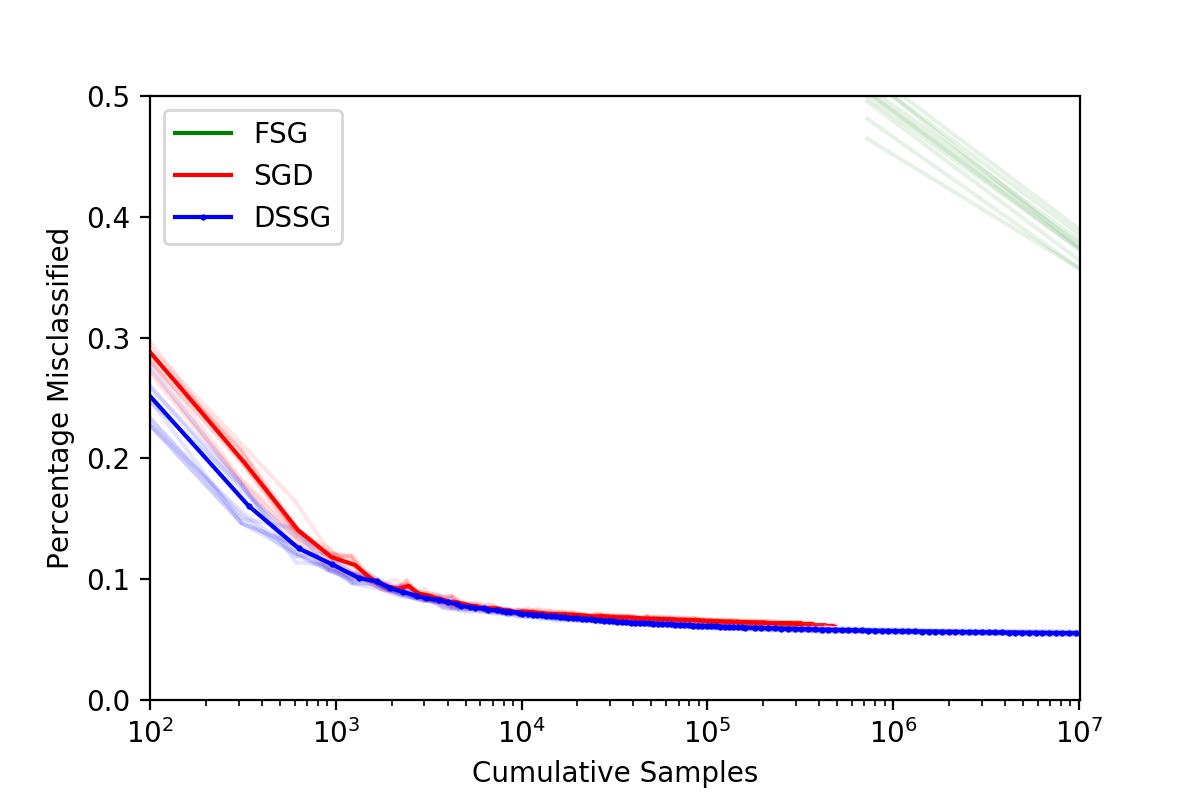}
		\includegraphics[width=0.32\textwidth]{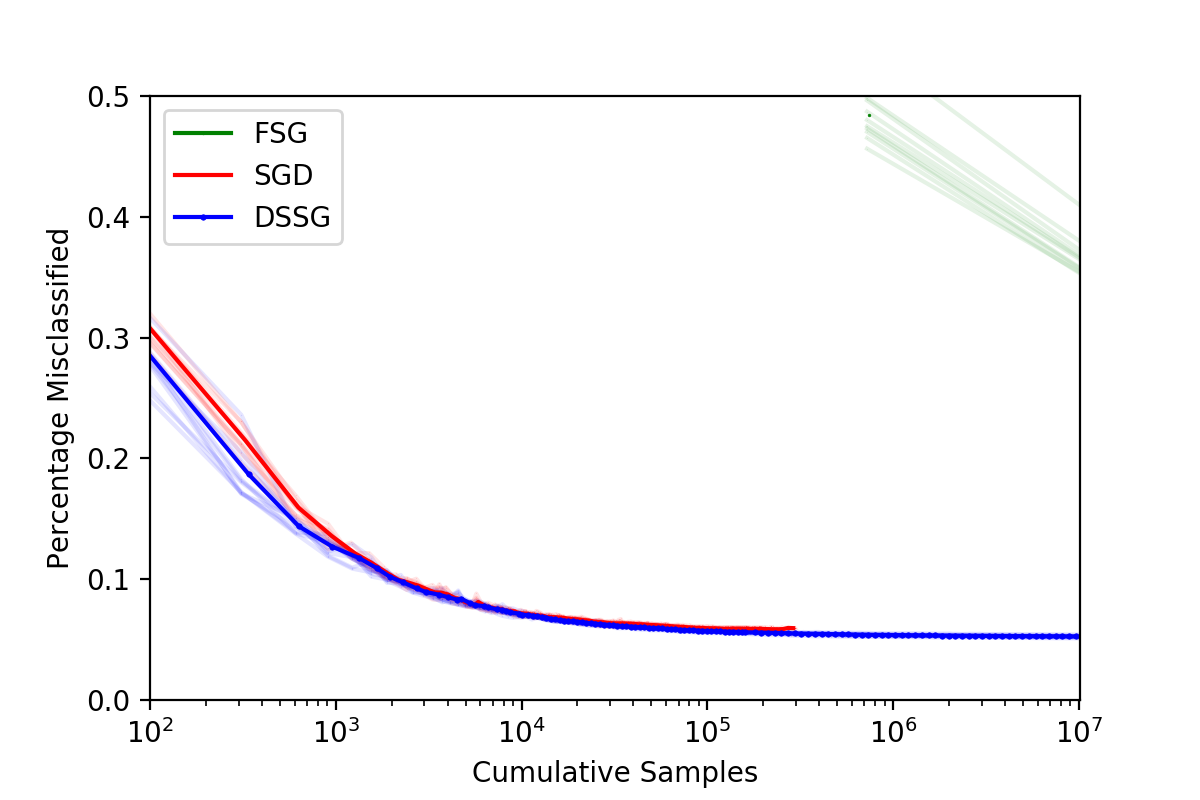}
		\caption{Comparisons of DSSG (blue), FSG (green) and standard SGD (red) on fraction of misclassification in testing ($y$-axis) versus cumulative samples ($x$-axis) over RCV1 dataset with (left) $\rho=0.01$, (center) $\rho = 0.5$ and (right) $\rho=1.0$.
			Log-scale cumulative samples $x$-axis.}
		\label{fig:rcv1.multirho}
	\end{center}
	\vskip -0.2in
\end{figure*}

We first present detailed results that compare the performance of DSSG against FSG and SGD in solving~(\ref{absfmln}) over three additional datasets.
This includes comparisons across additional $\rho$ values (Figures~\ref{fig:hiv1all},~\ref{fig:riccardo.multirho.class},~\ref{fig:adult.multirho} and~\ref{fig:rcv1.multirho}).
For the smallest dataset (HIV1), we also compare with the primal-dual method proposed in~\citeN{nd16} (Figure~\ref{fig:rho0.1:HIV+nd16}).
As a representative example, for the Riccardo dataset, we provide results on the robust loss objective calculated over the given training dataset (Figure~\ref{fig:riccardo.multirho.robloss}).
We further include results for additional $\rho$ values beyond what is provided in the the main body of the paper for the RCV1 dataset (Figure~\ref{fig:rcv1.multirho}).
Each set of experimental results is presented in turn.\\

\noindent\textbf{HIV-1 Protease Cleavage.}    
This dataset helps develop effective protease
cleavage inhibitors by predicting whether the HIV-1 protease will
cleave a protein sequence
in its central position ($y=1$) or not ($y=-1$).
After preprocessing following~\cite{Rogn}, this dataset has $N=5830$ samples of
$d=160$ feature vectors using orthogonal binary
representation, of which $991$ are cleaved and $4839$ non-cleaved. 
Figure~\ref{fig:rho0.1:HIV+nd16} compares the fractional test misclassification loss of DSSG and FSG against the corresponding results for a single run of the primal-dual proximal algorithm of~\citeN{nd16}.
The latter algorithm is computationally prohibitive, running for over $2$ days, and the end results fall short of the optimal solution.
This is as anticipated based on the corresponding discussion in the introduction.

\reffig{fig:hiv1all} presents a comparison of the fractional misclassification
performance of DSSG, FSG and SGD over the testing data from $10$ experimental runs, where each run uses a
different random partition of the data into training and testing datasets.
The four plots solve~(\ref{absfmln}) with $\rho=0.01$ (first), $\rho=0.1$ (second), $\rho=0.5$ (third) and $\rho=1$ (fourth).
Consistent with our results in the main body of the paper, we observe that DSSG is significantly faster than FSG for each value of $\rho$ considered, and that the effect of bias on SGD is muted.\\

\noindent\textbf{Riccardo}. This dataset has $d=4296$ features and
a moderate count of $N=20000$ samples, of which $5000$ are labeled as class $1$.
The plots in~\reffig{fig:riccardo.multirho.class} compare the fractional misclassification performance of the three methods over $10$ experimental runs with $\rho=0.01$ (first), $\rho=0.1$ (second), $\rho=0.5$ (third) and $\rho=1$ (fourth).
Consistent with our results in the main paper,
the DSSG method is once again observed to be significantly faster than FSG, and it attains the optimal solution faster than the SGD method for each value of $\rho$ considered.
This dataset produces the best outcomes in terms of the smallest misclassification errors in the test datasets. 
Figure~\ref{fig:riccardo.multirho.robloss} presents analogous comparisons among the DSSG, SGD and FSG methods on this dataset over the optimization objective of the robust training loss obtained.
Note that the SGD estimate of the robust loss suffers a high noise factor that remains relatively constant as the iteration count grows, while the noise in the estimate of the DSSG shrinks.
The latter happens because of the increasing batch size in DSSG.
As $\rho$ increases, the noise suffered by SGD grows.
This can potentially lead to premature termination of the algorithm if the criterion were to monitor the training robust loss values, which in turn
may lead to the bias issue identified by~\refthm{thm:bias}.
On the other hand, the shrinking noise of the robust loss estimate by the DSSG method as the iterate count grows ensures that algorithm termination with the robust loss criterion will perform well.\\

%
\noindent\textbf{Adult Income.} This 
dataset comprises $N=48842$ observations of $14$ attributes 
used to predict whether the annual income of each adult is above $50$K ($y=1$) or not ($y=-1$).
Using binary encoding of the categorical attributes, the data is transformed into $(d=119)$-dimensional features. 
%
\reffig{fig:adult.multirho} compares the DSSG, SGD and FSG methods based on the Adult Income dataset over test misclassification loss for model formulations
with $\rho=0.01$ (left), $\rho=0.1$ (center) and $\rho=0.5$ (right).
Note first that the best models found by the DRO formulations suffer a high misclassification error of $17.1\%$ of all the test data,
which can indicate that the chosen model and loss function may be inappropriate for this dataset.
Nevertheless, consistent with our results in the main body of the paper,
our proposed DSSG algorithm once again significantly outperforms the FSG algorithm for each value of $\rho$ considered,
with the best results obtained for $\rho=0.1$. All three methods suffer significant increase in stochastic noise as $\rho$ increases.\\

\begin{figure}[htbp]
	\vskip -0.1in
	\begin{center}
		\includegraphics[width=0.6\columnwidth]{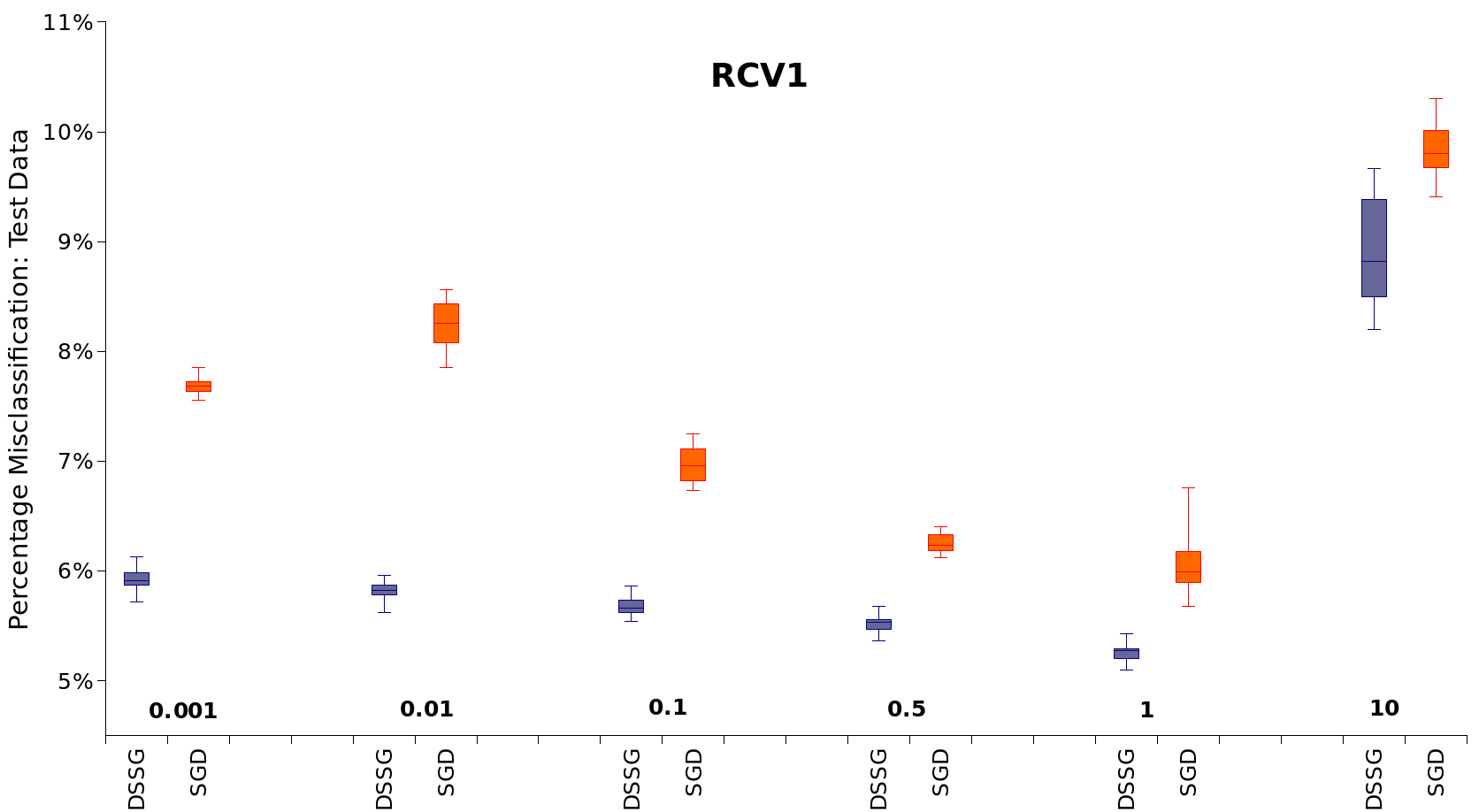}
		\caption{Comparisons of the misclassification performance of
			the SGD (red boxes) and the DSSG (blue boxes)
			algorithms over the RCV1 dataset, keeping algorithm parameters fixed and varying $\rho$; the $x$-axis labels contain the algorithm name and the $\rho$ value used. }
		\label{fig:moregen}
	\end{center}
	\vskip -0.2in
\end{figure}

\noindent\textbf{RCV1.}
Recall that this is the dataset considered in the main paper.
\reffig{fig:rcv1.multirho} compares the DSSG, SGD and FSG methods based on the RCV1 dataset over test misclassification loss for model formulations
with $\rho=0.01$ (left), $\rho=0.5$ (center) and $\rho=1$ (right).
Our proposed DSSG algorithm continues to considerably outperform the FSG algorithm for each of the additional $\rho$ values considered.
The SGD algorithm is hampered by the bias identified in~\refthm{thm:bias} and thus it is outperformed by DSSG for $\rho=0.01$.
This was also observed for $\rho=0.1$ in~\reffig{fig:rcv1} (left).
The impact of the bias diminishes when $\rho=0.5$ or $\rho=1$, holding all parameters of DSSG and SGD the same.

\reffig{fig:moregen} provides the test misclassification errors from the SGD and DSSG methods for various settings of $\rho$ (run with parameters set as described above).
The batch size of $10$ for SGD does produce bias, which in turn affects its performance in estimating solutions for the DRO formulation~\eqref{absfmln}.
However, a complex dependency exists between the bias and the parameter $\rho$ of the DRO problem, which further reiterates the message that DSSG saves on not having to
tune the batch size of the SGD for each instance of the DRO problem. \\


\begin{figure}[tbph]
	\centering
	\includegraphics[width=0.58\columnwidth]{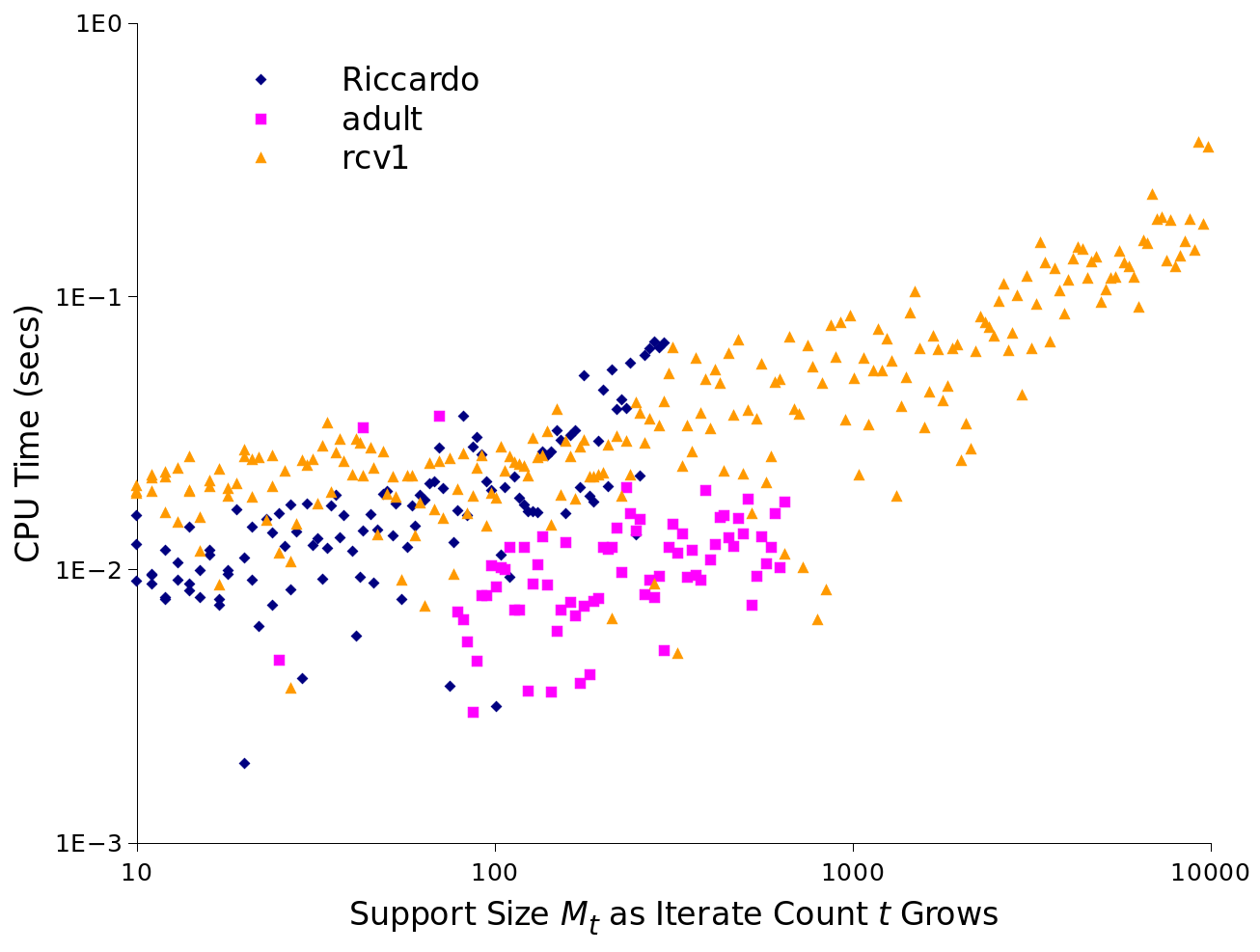}
	\caption{Comparison of CPU times (secs) taken to solve the inner formulation~(\ref{absfmln}) ($y$-axis) versus support size $M_t$ encountered over iterates when solving~(\ref{absfmln}) ($x$-axis) using DSSG for three datasets; $\rho=0.1$ in all cases.}
	\label{fig:cpuvsmbsize}
\end{figure}

\noindent\textbf{CPU Times as Computational Effort.}
Lastly, we turn to Figure~\ref{fig:cpuvsmbsize}, which plots the observed CPU computation times to solve the inner maximization problem in~(\ref{absfmln})
with respect to the size $M_t$ of the support that has been subsampled by the DSSG algorithm.
Observations from three separate optimization runs are provided for three datasets (Riccardo in dark blue, RCV1 in orange and Adult in pink), where the uncertainty radius was set to $\rho=0.1$;
both axes of the plot are in log-scale.
A broad near-linear trend is evident as anticipated from Proposition~\ref{prop:im_cmp_bnd}, with a small variability visible due to the computing platform and characteristics of each dataset, specifically the $z_m$ values in~(\ref{roblag}).
This supports our observation that there are no significant differences between the performance results presented in the paper as functions of cumulative samples and of CPU time,
and therefore the performance comparisons use cumulative samples as representative of the computational effort.\\ 	

{To summarize}, our experimental results above and in Section~\ref{sec:expt} support our theoretical results and show that DSSG provides the same quality of performance as FSG but with orders of magnitude less computational effort,
while also outperforming SGD and not requiring its hyper-parameter tuning.

\subsection{Generalization} \label{ssec:genexpt}
We now consider an extended set of experiments on model generalization to complement those presented in the main body of the paper.
In particular, Table~\ref{table:apdxgen} presents an extended set of results that complement those in~Table~\ref{table:gen},
also listing values for $d$, $N$ and $\sqrt{d/N}$ for each of the $14$ public-domain datasets we studied.
The broad guideline that $\rho = O(\sqrt{d/N})$ for binary classification with logistic models, provided by~\citeN{bwm16} and \citeN{nd17},
led us to pick the value of $\rho=0.1$ as being most suitable overall for the experiments in Section~\ref{ssec:droexpt} (and Section~\ref{sec:expt}).\\

\noindent\textbf{Varying $\rho$ in DRO~\eqref{absfmln}}.
The actual values in the third column of Table~\ref{table:apdxgen} suggest that, in a minority of datasets, a value of $\rho=0.01$ or $\rho=1.0$ might possibly be preferable.
To investigate this, we re-ran the DSSG algorithm with both these values of $\rho$ for all the datasets, and we present the corresponding results in columns four and six of Table~\ref{table:apdxgen}. 

Note first that the triplet of results for each dataset exhibits a monotonic or convex change in the performance outcomes.
The results for $\rho=0.01$ does not provide any additional benefit in terms of improving the generalization performance of DRO over that of the $10$-fold CV regularized ERM in any dataset
where DRO could not already achieve the best test performance with $\rho=0.1$.
However, the addition of the $\rho=1.0$ results does help the DRO formulation match the generalization performance of the regularized ERM method or further improve the DRO performance in a handful of cases.
For the cases of gina, gina\_prior and Bioresponse, ERM continues to provide better performance, but we note here that the best obtained test misclassification percentage is high,
indicating that the efficacy of linear models with the logistic loss function may be limited in these cases.
Each run of DSSG for $\rho=0.01$ and $\rho=1.0$ took the same order of computational effort as the runs for $\rho=0.1$.
Hence, the added benefit of a few extra DRO runs is obtained without significantly multiplying the computational time.
In practice, since the constants associated with the $\rho = O(\sqrt{d/N})$ guideline are hard to compute,
one may wish to try a couple of $\rho$ values relatively close by to determine if additional improvements in performance can be obtained.\\

\begin{figure*}[htbp]
	\begin{center}
		\includegraphics[width=0.47\linewidth]{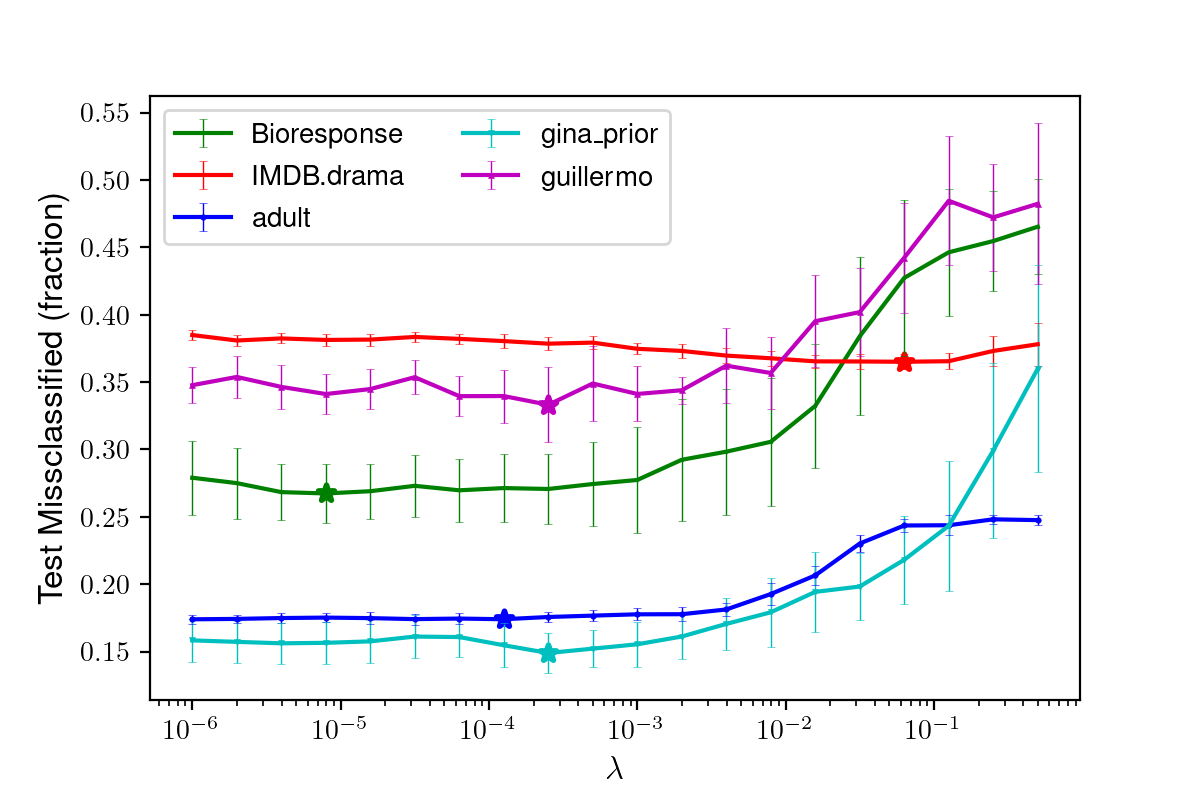}
		\includegraphics[width=0.47\linewidth]{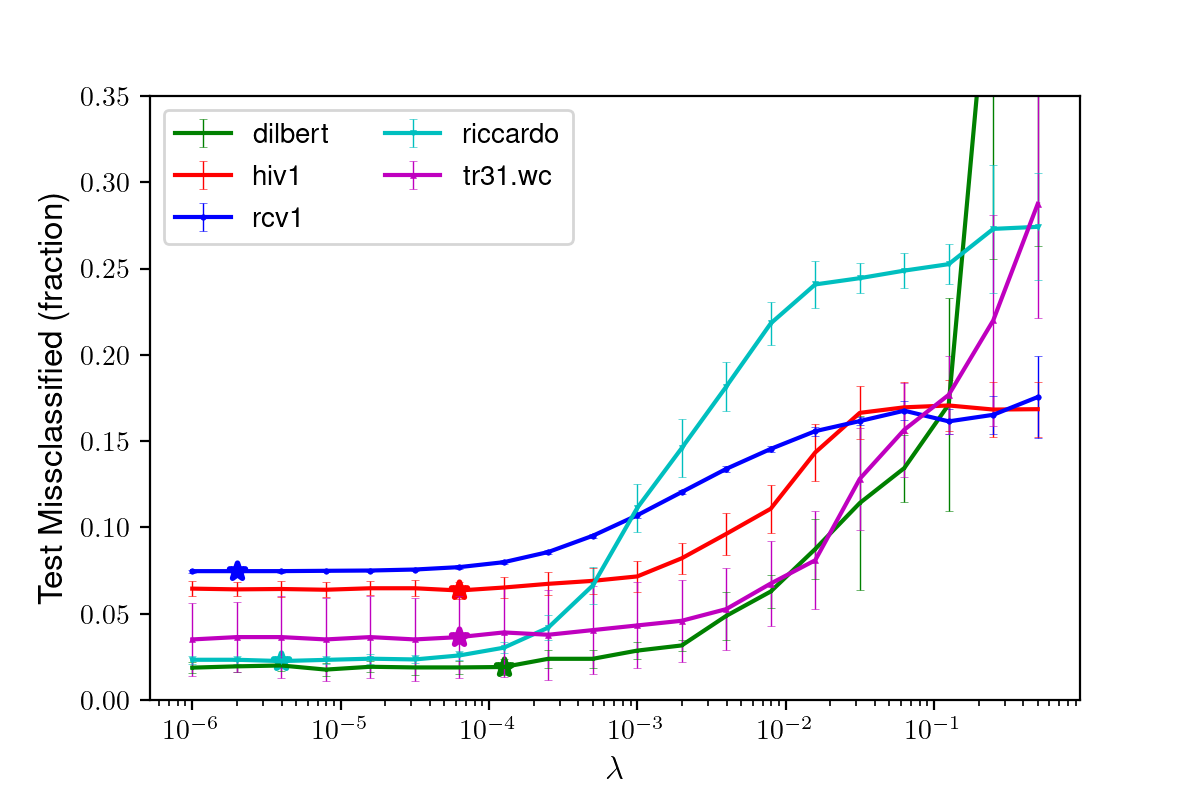}
		\caption{Illustration of optimizing the parameter $\lambda$ for the ERM regularized formulation by enumeration, with the average test misclassification (\%) over $10$ partitions of the data plotted on the $y$-axis versus $lambda$ on the $x$-axis (in log-scale). The left plot shows five datasets that achieve a best performance of $15\%$ or higher and the right five datasets that achieve a best performance of $9\%$ or lower. }
		\label{fig:multilambda}
	\end{center}
	\vskip -0.2in
\end{figure*}

\noindent \textbf{Regularization of ERM.} Figure~\ref{fig:multilambda} provides the outcome of the full set of enumerations on the 10 datasets to elicit the best value of the penalty parameter $\lambda$ in the regularized ERM objective of $L_{U_N}(\theta) + \lambda \|\theta\|^2$.
The SGD algorithm is used to solve each instance of the regularized formulation in each of the $10$ partitions of the dataset.
A mini-batch size of $10$ was used along with step size sequence $\gamma_t=0.25 *(5000./(5000.+t))$. 
The plots on the left provide the regularization enumeration for five datasets that achieved a best test misclassification error of 15\% or higher,
whereas the plots on the right provide the regularization enumeration for five datasets that achieved a best test misclassification error of 9\% or lower.
The $\lambda$ value chosen by the backtracking enumeration approach (described in Section~\ref{sec:expt}) is marked for each dataset.
It is clear from these plots that the hyperparameter tuning of $\lambda$ is non-trivial.
The shape of the curve of the mean outcomes of the regularization formulation varies significantly across the datasets, and moreover the variability exhibited is significantly impacted by the dataset characteristics.
While generalization performance seems to improve as $\lambda\tndo$ (except for imdb.drama), it is not clear that a single $\lambda$ value can be picked to perform well over all the datasets.
This is in sharp contrast to the DRO formulation, where the choice of $\rho=0.1$ (as the most likely order of the $\sqrt{d/N}$ values) provides good generalization performance.

Recall that DSSG provides this level of performance by solving a single instance of the DRO formulation~\eqref{absfmln},
thus avoiding the burdensome $10$-fold CV enumeration.
As observed in the discussion of the CPU times provided in Table~\ref{table:gen},
the time taken by DSSG to solve each DRO formulation is on average of the same order as that taken by SGD to solve a single instance of the ERM formulation for a single $\lambda$ value.
This indicates a significant computational savings in using DRO because of the elimination of the expensive hyper-parameter tuning step,
with the time required under DSSG being one to three orders of magnitude superior to the time taken by the ERM regularization (enumeration with backtracking) procedure.\\

To summarize, our experimental results above and in Section~\ref{sec:expt} support our theoretical results
and show that DSSG renders models of comparable or better quality as those from regularized ERM but with orders of magnitude less computational effort,
and thus provides a strong alternative ML approach to improve model generalization.
Once again, the main advantages of our DRO algorithm are that it does not need any further tuning, it efficiently provides a solution to~\eqref{absfmln}, and it naturally provides a strong generalization guarantee.

\input{apdx_results_table.tex}

%% file: apdx_results_table.tex
\begin{table*}[!htbp]
	\vskip -0.1in
	\setlength{\tabcolsep}{7.5pt}
	\centering
	\begin{tabular}{l||r|r|r||r||r||r||r}
Dataset	&$d$	&$N$	&$\sqrt{d/N}$	&{$\rho = 0.01$}	&{$\rho = 0.1$}		&{$\rho= 1.$} 	&{Reg. ERM } \\ 
\hline\hline
adult	&119	&45222	&0.05	&$\mathbf{16.6\pm 0.1}$	& $\mathbf{16.6\pm 0.1}$	& $17.7\pm 0.2$	& $\mathbf{16.7\pm 0.0}$ \\
imdb.drama	&1001	&120919	&0.09	& $36.6\pm 0.1$  &$\mathbf{36.2\pm 0.1}$  &$\mathbf{36.2\pm 0.1}$  &$37.1\pm 0.0$\\ 
hiv1	&160	&5830	&0.17	&$5.8\pm 0.1$	& $5.9\pm 0.1$	& $5.8\pm 0.1$	& $\mathbf{5.6\pm 0.0}$ \\
rcv1	&47236	&804414 &0.24	&$7.1\pm 0.0$  &$\mathbf{5.4\pm 0.0}$  &$\mathbf{5.4\pm 0.0}$  &$5.6\pm 0.0$ \\
fabert	&800	&8237	&0.31	& $10.0\pm 0.2$	& $9.8\pm 0.0$	& $\mathbf{9.1\pm 0.2}$	& $10.1\pm 0.0$ \\ 
dilbert	&2000	&10000	&0.45	& $\mathbf{1.2\pm 0.1}$  &$\mathbf{1.2\pm 0.1}$  &$1.7\pm 0.1$  &$\mathbf{1.2\pm 0.0}$ \\ 
guillermo	&4296	&20000	&0.46	&$31.1\pm 0.2$  &$\mathbf{30.2\pm 0.5}$  &$35.4\pm 0.5$  &$\mathbf{30.7\pm 0.1}$ \\ 
riccardo	&4296	&20000	&0.46	&$2.1\pm 0.1$  &$\mathbf{1.6\pm 0.0}$  &$\mathbf{1.5\pm 0.1}$  &$\mathbf{1.5\pm 0.0}$  \\ 
gina\_prior	&784	&3468	&0.48	&$12.7\pm 0.5$	&$ 13.0\pm 0.5$	&$ 14.0\pm 0.4$	& $\mathbf{11.8\pm 0.1}$ \\
gina\_agnostic	&970	&3468	&0.53	&$13.2\pm 0.3$	&$13.9\pm 0.3$  &$14.8\pm 0.6$  &$\mathbf{12.6\pm 0.1}$ \\ 
Bioresponse	&1776	&3751	&0.69	& $23.7\pm 0.3$  &$24.2\pm 0.4$  &$25.8\pm 0.5$  &$\mathbf{21.6\pm 0.2}$\\ 
la1s.wc	&13195	&3204	&2.01	&$8.9\pm 0.1$  &$\mathbf{8.3\pm 0.2}$  &$\mathbf{7.9\pm 0.3}$  &$8.5\pm 0.0$ \\
OVA\_Breast	&10935	&1545	&2.73	&$2.9\pm 0.1$  &$3.0\pm 0.1$  &$2.9\pm 0.1$  &$\mathbf{1.8\pm 0.1}$ \\
tr31.wc	&10128	&927	&3.32	&$\mathbf{2.8\pm 0.3}$  &$\mathbf{2.6\pm 0.3}$  &$\mathbf{2.9\pm 0.3}$  &$\mathbf{2.7\pm 0.1}$ \\ 
\hline
	\end{tabular}
\caption{Comparison of the DRO and regularized ERM formulations over $14$ publicly available machine learning (ML) datasets,
from UCI$^{\ast}$ \cite{UCI}, OpenML$^{\dag}$ \cite{openml} and SKLearn$^{\ddag}$ \cite{RCV1},
arranged in increasing value of the third column. 
The first pair of columns provides the $d$ and $N$ characteristics of each dataset. The third column provides $\sqrt{d/N}$, the characteristic that~\citeN{bwm16,nd17} provide as a guideline for setting $\rho$ in~(\ref{absfmln}). The fourth, fifth and sixth columns provide a 95\% confidence interval (CI) of the percentage misclassified over withheld test datasets for $\rho=0.01, 0.1 $ and $1$, respectively, while the final column provides the same for the $10$-fold CV regularized ERM method. The best-performing method (within its CI) is highlighted in bold. Note that the fifth and final columns are repeated from Table~\ref{table:gen}.}
\label{table:apdxgen}
\vskip -0.2in
\end{table*}